\documentclass{article}
\usepackage{arxiv}

\usepackage[utf8]{inputenc} 
\usepackage[T1]{fontenc}    
\usepackage{hyperref}       
\usepackage{url}            
\usepackage{booktabs}       
\usepackage{amsfonts}       
\usepackage{nicefrac}       
\usepackage{microtype}      
\usepackage{lipsum}

\usepackage[numbers]{natbib}
\usepackage{hyperref}
\usepackage{graphicx}
\usepackage{amsmath}
\usepackage{amssymb}
\usepackage{mathtools}
\usepackage{todonotes}
\usepackage{amsfonts}
\usepackage{verbatim}
\usepackage{float}
\usepackage{subfigure}
\usepackage{inc/slashbox}
\usepackage[english]{babel}
\usepackage{blindtext}
\usepackage{enumitem}
\usepackage{tikz}
\usepackage{siunitx}
\usepackage{xcolor}
\usepackage{multirow}
\usetikzlibrary{arrows,positioning,decorations.pathreplacing}

\xspaceaddexceptions{]}

\title{Transfer learning for Remaining Useful Life Prediction Based on Consensus Self-Organizing Models}

\author{
Yuantao Fan \\
CAISR, Halmstad University, Sweden \\
\texttt{yuantao.fan@hh.se} \\
 \And
S\l{}awomir Nowaczyk \\
CAISR, Halmstad University, Sweden \\
 \And
Thorsteinn R\"{o}gnvaldsson \\
CAISR, Halmstad University, Sweden \\
}

\begin{document}
\maketitle

\DeclareRobustCommand{\rchi}{{\mathpalette\irchi\relax}}
\newcommand{\irchi}[2]{\raisebox{\depth}{$#1\chi$}} 

\begin{abstract}

The traditional paradigm for developing machine prognostics usually relies on generalization from data acquired in experiments under controlled conditions prior to deployment of the equipment. Detecting or predicting failures and estimating machine health in this way assumes that future field data will have very similar distribution to the experiment data. However, many complex machines operate under dynamic environmental conditions and are used in many different ways. This makes collecting comprehensive data very challenging, and the assumption that pre-deployment data and post-deployment data follow very similar distributions is unlikely to hold.

Transfer Learning (TL) refers to methods for transferring knowledge learned in one setting (the source domain) to another setting (the target domain).
In this work we present a TL method for predicting Remaining Useful Life (RUL) of equipment, under the assumption that labels are available only for the source domain and not the target domain. This setting corresponds to generalizing from a limited number of run-to-failure experiments performed prior to deployment into making prognostics with data coming from deployed equipment that is being used under multiple new operating conditions and experiencing previously unseen faults. We employ a deviation detection method, Consensus Self-Organizing Models (COSMO), to create transferable features for building the RUL regression model. These features capture how different a particular equipment is in comparison to its peers. 

The efficiency of the proposed TL method is demonstrated using the NASA Turbofan Engine Degradation Simulation Data Set. Models using the COSMO transferable features show better performance than other methods on predicting RUL when the target domain is more complex than the source domain.

\end{abstract}


\section{Introduction}
\label{sec:introduction}

To ensure safety requirements of industrial systems in a reliable and cost-effective way, methods for predictive maintenance \cite{mobley2002introduction, zhang2019data, rieger2019fast} and condition-based maintenance \cite{bousdekis2018review, peng2010current} are increasingly demanded, as they allow flexible scheduling of the maintenance based on the condition of the equipment, e.g. based on how much remaining useful life (RUL) the equipment has.
Aiming at predicting the RUL of a system or a component, prognostics and health management (PHM) methods \cite{kothamasu2006system, tsui2015prognostics, schwabacher2007survey, khan2018review} have been researched for a long time and applied to various industrial applications such as aerospace, transportation, energy production, maritime equipment and manufacturing \cite{hanachi2018performance, rezvanizaniani2014review, kandukuri2016review, coble2012prognostics, ellefsen2019comprehensive, lee2014prognostics, vogl2019review}. Diagnostic and prognostic systems in the current industrial assets were developed based on controlled experiments with simulated operating conditions and pre-defined faults, assuming data from the simulation at the training time and the unseen future data are of the same population. However, this assumption might not hold for complex machines that can do different things and operate under varying conditions. Applying transfer learning (TL) techniques such as domain adaptation or feature representation transfer methods can help to improve the robustness of performing prognostics against scenarios where new conditions and new faults were present in the unseen future data. This corresponds to generalizing from a limited number of experiment cases into making prognostics for deployed equipment that operate under new conditions and experience new faults.

The RUL of a system is defined as the time interval from the particular time of operation until the end of the system's useful life, i.e. when it is incapable of performing its functions \cite{si2011remaining}. Predicting RUL is commonly considered as learning a functional mapping between an observation of sensor measurement and the health condition of the equipment and/or RUL. Prognostic methods are conventionally categorized into three types \cite{medjaher2012remaining}: physical model based \cite{oppenheimer2002physically, mathew2008failure}; data-driven; \cite{rigamonti2016echo}, and hybrid combinations of the former two \cite{garga2001hybrid, khorasgani2018framework}. 

The physical model-based approaches require a good theoretical understanding of the mechanism of the target system as well as the failure progression. Such approaches do not scale well with respect to the complexity of the system; the more complex the system, the more difficult it is to build a faithful physical model. In contrast, data-driven methods do not require extensive knowledge of the physical mechanisms \cite{heng2009rotating, eker2012major}, but they require data with comprehensive coverage of various usages, wear patterns and failure progression, e.g. run-to-failure cases, of the target system. 
Acquiring run-to-failure cases in industrial systems is expensive, and many systems are not allowed to run until failure, often for safety reasons. This means that priorities and trade-offs must be made on which failure cases to collect data for. Furthermore, deterioration of many wear failures progress very slowly and it might take months, perhaps even years, of continuous operation for the first failure cases to develop \cite{gebraeel2009residual}. 

The current industrial solution for developing data-driven prognostic methods relies heavily on data from simulations, stress tests (or accelerated degradation test) and experiments \cite{li2014experiments, nectoux2012pronostia, kleyner2017new, yin2008physics} with predefined faults under controlled conditions. This assumes that controlled experiments are representative of the operating conditions and failure progressions that occur in the field. If this holds, the prognostic model built based on the controlled experiment data will work fine in the real-world application. However, many complex machines, e.g. heavy-duty and construction vehicles, are deployed under many different conditions and sometimes deteriorate in unexpected ways. The traditional paradigm for designing diagnostic and prognostic method does not take this into account, i.e. that training and testing data come from different populations.

To address this issue, recently, TL \cite{pan2010survey, weiss2016survey} has been applied to machine prognostics, e.g. \cite{zhang2017transfer, wen2017new, zhang2018transfer, yang2018transfer, wang2019domain, lu2016deep, long2015learning, da2019remaining}. Transfer learning aims at acquiring knowledge from solving one problem, where labeled data are abundant, and modifying this knowledge to solve a different but related problem, where labeled data are difficult or expensive to collect. 
%
In the context of TL, the training and testing samples are referred to as the source samples $X_S$ and the target samples $X_T$. Correspondingly, they come from the source domain $\mathcal{D}_S$, where useful knowledge is obtained from solving the source task $T_S$, and the target domain $\mathcal{D}_T$, where knowledge acquired from the source domain $\mathcal{D}_S$ is adapted, transferred and applied to solve the target task $T_T$.

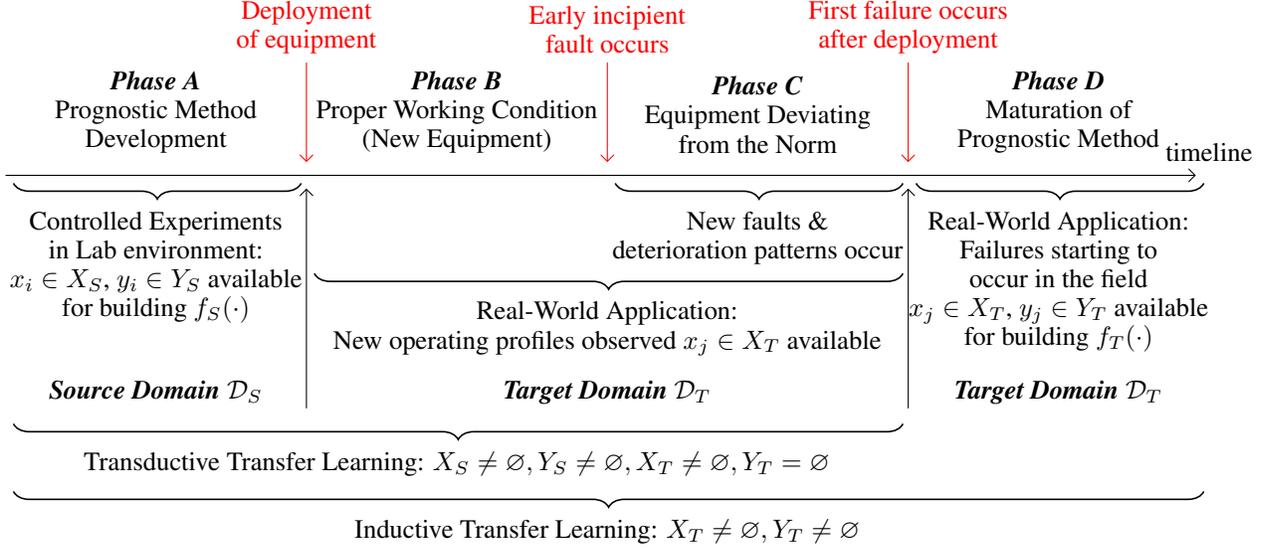
\begin{figure*}
\centering
\begin{tikzpicture}[
    every node/.style = {align=center},
          Line/.style = {-angle 90, shorten >=5pt},
    Brace/.style args = {#1}{semithick, decorate, decoration={brace,#1,amplitude=6pt,raise=3pt,
                             pre=moveto,pre length=3pt,post=moveto,post length=2pt,}},
            ys/.style = {yshift=#1}
                    ]
\linespread{1.}                    
\coordinate (a) at (0,0);
\coordinate[above=-32mm of a]    (a2);
\coordinate[above=-41mm of a]    (a3);
\coordinate[right=40mm of a]    (b);
\coordinate[below=12mm of b]    (b2);
\coordinate[right=40mm of b]    (c);
\coordinate[right=40mm of c]    (d);
\coordinate[below=12mm of d]    (d2);
\coordinate[above=.5mm of d2]   (d3);
\coordinate[above=-32mm of d]    (d4);
\coordinate[right=40mm of d]    (e);
\coordinate[above=12mm of e]    (e2);
\coordinate[above=-41mm of e]    (e4);
\coordinate[above=25mm of b]    (f);
\coordinate[right=15mm of f]    (f1);
\coordinate[above=35mm of c]    (g);
\coordinate[above=35mm of d]    (h);

\draw[Line] (a) -- node[above = 5pt] {\textbf{\emph{Phase A}} \\ Prognostic Method \\ Development} (b)
            (b) -- node[above = 5pt] {\textbf{\emph{Phase B}} \\ Proper Working Condition \\ (New Equipment)} (c)
            (c) -- node[above = 5pt] {\textbf{\emph{Phase C}} \\ Equipment Deviating \\ from the Norm}  (d)
            (d) -- node[above = 5pt] {\textbf{\emph{Phase D}} \\ Maturation of \\ Prognostic Method}
            (e) node[above=2pt] {timeline};
            
\draw[Line, color = red] ([ys=15mm] b) node[above] {Deployment \\ of equipment} -- (b);
\draw[Line, color = red] ([ys=15mm] c) node[above] {Early incipient \\ fault occurs} -- (c);
\draw[Line, color = red] ([ys=15mm] d) node[above] {First failure occurs \\ after deployment} -- (d);

\draw[Brace=mirror] (a) -- node[below=10pt] {Controlled Experiments \\ in Lab environment: \\ $x_i \in X_S$, $y_i \in Y_S$ available \\ for building $f_S(\cdot)$ } (b);            
\draw[Brace=mirror] (c) -- node[below=10pt] {New faults \& \\ deterioration patterns occur} (d);
\draw[Brace=mirror] (d) -- node[below=10pt] {Real-World Application: \\ Failures starting to \\ occur in the field \\ $x_j \in X_T$, $y_j \in Y_T$ available \\ for building $f_T(\cdot)$} (e);

\draw[Brace=mirror] (b2) -- node[below=10pt] {Real-World Application: \\ New operating profiles observed $x_j \in X_T$ available} (d2);


\draw[Line, color = black] ([ys=-31mm] b) node[] {} -- (b);
\draw[Line, color = black] ([ys=-31mm] d) node[] {} -- (d);

\node[below] at (20mm,-26mm) {\textbf{\emph{Source Domain $\mathcal{D}_S$}}};

\node[below] at (80mm,-26mm) {\textbf{\emph{Target Domain $\mathcal{D}_T$}}};

\node[below] at (140mm,-26mm) {\textbf{\emph{Target Domain $\mathcal{D}_T$}}};


\draw[Brace=mirror] (a2) -- node[below=10pt] {Transductive Transfer Learning: $X_S \neq \varnothing, Y_S \neq \varnothing, X_T \neq \varnothing, Y_T = \varnothing$} (d4);


\draw[Brace=mirror] (a3) -- node[below=10pt] {Inductive Transfer Learning: $X_T \neq \varnothing, Y_T \neq \varnothing$} (e4);



\end{tikzpicture}
\caption{Transfer Learning for Developing Prognostic Methods} \label{fig:TL_for_PHM}
\end{figure*}

Throughout this paper, we use the notation proposed by Pan and Weiss \cite{pan2010survey, weiss2016survey}. A domain $\mathcal{D}$ consists of two components: its feature space $\rchi$ and a marginal probability distribution $P(X)$, where $X$ denotes samples from this domain. For a given domain $D$, a task $T$ consists of two components: a set of labels $Y$ and a prediction function $f(\cdot)$. The domain and its correspondent task are denoted by $\mathcal{D}=\{\rchi, P(X)\}$ and $T=\{Y, f(\cdot)\}$. The prediction function $f(\cdot)$ can be learnt from sample pairs $\{\mathbf{x}_i, \mathbf{y}_i\}$, where $\mathbf{x}_i \in X$ and $\mathbf{y}_i \in Y$. 

We can exemplify what can happen in machine prognostics using the TL terminology. We refer to the situation before deployment as the source data, and the situation after deployment as the target data. If the operating conditions experienced in the field are different from those observed prior to deployment, then this means that the marginal distribution $P(X_T)$ is different from $P(X_S)$. If new (previously unseen) deterioration profiles (faults) are present in the target domain, the mapping function $f_S: x \mapsto y$ is different from $f_T$. 
Based on whether the labels $Y_S$ and $Y_T$ are available at the training time, and whether the tasks $T_S$ and $T_T$ are equivalent, TL can be categorized into three different settings: inductive, transductive and unsupervised TL \cite{pan2010survey}.

A timeline of the development of prognostic methods is illustrated in Figure \ref{fig:TL_for_PHM}. Prognostic methods are initially built during \emph{Phase A}, based on controlled experiments. In this phase are labeled data $\{\mathbf{x}_s, \mathbf{y}_s\}$ available. In \emph{Phase B}, the equipment is deployed to the application. It may encounter operating profiles that were not previously observed during \emph{Phase A}. New (unseen) faults might occur at some point in time (\emph{Phase C}) after the deployment, and the equipment might deteriorate with a pattern that is different from the ones observed during \emph{Phase A}. Observations $\mathbf{x}_i$ are available during \emph{Phase B} and \emph{Phase C}. The first (batch) occurrence of failures marks the starting of \emph{Phase D}, i.e. maturation of prognostic methods. During this phase, prognostic methods can be improved with deterioration patterns that actually occur in the target real-world application. 

The maturation (\emph{Phase D}) of prognostic methods can be conducted under inductive TL setting, or with multitask learning, if new deterioration pattern(s) are present in the real-world application (\emph{Phase B} to \emph{Phase D} as the target domain $D_T$) but not in the controlled experiment (\emph{Phase A} as the source domain $D_S$). In this case, some amount of labeled data are required in $D_T$ to induce a prediction model $f_T(\cdot)$ for solving $T_T$. The objective of the inductive transfer learning is to utilize labeled or unlabeled data $X_S$ from $D_S$ to improve the prediction performance of $f_T(\cdot)$ in solving $T_T$. 

Transductive TL can be conducted when labels $\mathbf{y}_T$ of testing samples are unavailable. Transductive TL aims at utilizing unlabelled testing data $X_T$ for improving the learning of the target prediction function $f_T(\cdot)$ in $D_T$, using the knowledge in $D_S$ and $T_S$. It makes sense to perform transductive TL when $D_S \neq D_T$. This implies that the marginal distributions of the source and target data are different, i.e. $P(X_S) \neq P(X_T)$, or the source and target data reside in different feature spaces, i.e. $\rchi_S \neq \rchi_T$. The objective, in this case, is very similar to feature representation transfer or domain adaptation: finding a latent feature space that has predictive quality in solving $T_T$ while the discrepancy between the marginal distributions of samples from the two domain is reduced. 

Most TL approaches to machine prognostics are examples of inductive TL. Several parameter transfer techniques \cite{zhang2017transfer, wen2017new, zhang2018transfer, yang2018transfer} based on deep neural networks (DNN) have been applied for machine prognostics; DNNs are first trained with the source data $\{\mathbf{x}_S, \mathbf{y}_S\}$ and then fine-tuned with (usually a relatively small amount of) labeled target data $\{\mathbf{x}_T, \mathbf{y}_T\}$ to solve task $T_T$. This approach requires labeled samples from both domains and cannot be conducted before \emph{Phase D}. Supervised and unsupervised fault detection technique \cite{Venkat-03a, Venkat-03b, Venkat-03c} have been applied to monitor equipment after deployment, finding abnormal behavior that is deviating from a reference model. 

A suitable technique for transductive TL is domain adaptation. It aims at discovering meaningful common structures between the source and the target domain, finding transformations $\Phi(\cdot)$ that project $X_S$ and $X_T$ into a common latent feature space $\rchi_\Phi$, which has predictive qualities for solving $T_T$. At the same time is the difference in the marginal distribution between the source and the target domain in the latent feature space $\rchi_\Phi$ reduced.
Maximum mean discrepancy (MMD) \cite{gretton2012kernel} is a popular metric for estimating the discrepancy between distributions in domain adaptation methods. Transfer component analysis (TCA) \cite{pan2011domain} finds components across domains based on MMD such that, in the subspace found, data distributions of the two domain are closer and data properties still persevered. Correlation alignment (CORAL), proposed by Sun et al. \cite{sun2016return}, aligns the second-order statistics of source and target distributions to minimize the domain shift. Structural correspondence learning (SCL) \cite{blitzer2006domain} learns a common feature representation that is meaningful across the source and the target domains. Another emerging approach is based on using domain adversarial neural networks (DANN) \cite{ajakan2014domain, ganin2014unsupervised} for domain adaptation. Augmented with a gradient reversal layer that backpropagates gradient from a domain classifier, the DANN is designed to train (deep) neural networks to extract domain-invariant features that also contain predictive quality for the learning task on the source domain.
Domain adaptation has a strong similarity to feature representation based TL \cite{pan2008transfer, pan2011domain, blitzer2006domain, fernando2013unsupervised, sun2016return, zhang2017joint}, which is one of the four general types of TL summarized in \cite{pan2010survey}. The other three types are instance based \cite{dai2007boosting, huang2007correcting, jiang2007instance, tan2015transitive, tan2017distant}, parameter based \cite{nater2011transferring, zhao2011cross, zhang2018transfer}, and relational knowledge based \cite{mihalkova2007mapping, mihalkova2008transfer} TL.

In this study, we perform feature representation transfer for RUL prediction under the TL scenario where labeled source data ($X_S$ and $Y_S$) and unlabeled target data ($X_T$) are available throughout, but target labels ($Y_T$) are not available (before \emph{Phase D}). 
The proposed transferable feature for predicting RUL is computed based on the ideas in the consensus self-organizing models (COSMO) method \cite{ByttnerRS-11, rognvaldsson2018self}. 
The COSMO method computes deviation levels, based on p-values, that reflect how likely it is that an individual system is deviating from a reference group (a peer group), ideally composed by nominal samples. The COSMO method was developed to be a generic method for detecting anomalies.
In previous works \cite{FanSR-15, FanSCAI-15, FanPHME-16}, the COSMO method was applied to detect deviations and faults in a fleet of city buses with streaming onboard data. The reference (or peer) group for computing deviation levels was drawn from active vehicles during the same time period (seven days) across the whole fleet. Concept drift problem such as seasonality changes were handled with a dynamically updated reference group.

We propose to use \emph{distance} to the peers, instead of the probability for deviation (which is bounded), as a transferable feature with predictive quality for RUL prediction. The hypothesis is that both the source and the target data are projected into a latent space where distances of each sample to a reference group is preserved, i.e. the feature is a transferable feature. The proposed approach is tested and verified on the \href{https://ti.arc.nasa.gov/tech/dash/groups/pcoe/prognostic-data-repository/}{Turbofan Engine Degradation Simulation Data Set}, which is generated by C-MAPSS (Commercial Modular Aero-Propulsion System Simulation) \cite{saxena2008damage}. The data set contains four subsets with different operating conditions and faults, which resemble a good case for transfer learning in the context of machine prognostics. 

The contribution of this work is the COSMO TL approach for predicting RUL of equipment under the scenario that labeled data are only available for the source domain but not for the target domain. 
We propose to employ COSMO, a group based deviation detection method, for generating transferable features that capture differences between a particular equipment sensor readings and its peers. This feature is computed sensor-wise and can be considered as an indicator of how different each sub-system of the equipment performs compared to its peers.
The hypothesis is that the COSMO method, as a feature representation transfer technique, transforms the source and the target data into a latent feature space where distances of testing samples to its peers are preserved.
%
A mapping function using random forest (RF) regression model is learned between the COSMO features and RUL for the prediction task.
Through experimental results on the Turbofan Engine Degradation Simulation Data Set, we demonstrate that the proposed approach with RF regression can predict RUL. 
Although only trained using labeled data from a simpler scenario, the proposed approach, with RF regressor, is capable of generalizing from run-to-failure trajectories in a simpler scenario to predicting RUL on data coming from more complex scenarios, where new operating conditions and novel faults are encountered. 
The proposed TL approach is compared to a traditional approach and well-known domain adaptation methods TCA, CORAL, and SCL. 
\section{RUL Prediction for Run-to-Failure Turbofan Engines}\label{sec:background}

\subsection{C-MAPSS Dataset}\label{subsec:C-MAPSS Dataset}

The Turbofan Engine Degradation Simulation Data Set \cite{saxena2008damage} contains simulated run-to-failure trajectories of one type of turbofan engine. The data were generated using C-MAPSS (Commercial Modular Aero-Propulsion System Simulation). The data set has four subsets, each with a different number of operating conditions and fault modes, see Table~\ref{tab:Dataset}. Each subset is further split into a training and a test set. In the training sets $X_\alpha$, the faults grow until the system ultimately fails, i.e. the training sets contain full run-to-failure trajectories. In the test sets $X_\beta$, the trajectories end before the system failure, i.e. the test sets consist of truncated trajectories (right censored data). Sometimes the end of the trajectory is a long time before the end of life for the system. 

We denote each of the four subsets with $X^i$, where $i \in \{ 1, 2, 3, 4\}$ indicates the number of the subset. The training and testing part of each subset are denoted by $X^i_\alpha$ and $X^i_\beta$, respectively. Trajectories in subsets $X^1$ and $X^2$ were operated under a single operating condition while trajectories in subsets $X^2$ and $X^4$ were operated under six different operating conditions. Trajectories in subsets $X^1$ and $X^2$ fail due to high-pressure compressor (HPC) degradation, while trajectories in subsets $X^3$ and $X^4$ can suffer from HPC degradation and/or fan degradation (i.e. two possible fault modes).

Each subset contains several multivariate time series (trajectories) of $24$ features; $21$ are sensor measurements and three features correspond to the operating (flight) conditions of the engine during each cycle. We denote the multivariate data of each trajectory $u$ by:
\begin{align*}
\mathbf{x}_u = \{\ x_{u,t}^{i} \ |\ t = 1, 2, ..., l(u), \ i = 1, 2, ..., 24\}
\end{align*}
Where $x_{u,t}^{i}$ is the value of the $i^{th}$ feature of trajectory $u$ at cycle $t$, and $l(u)$ is the length of the trajectory (i.e. the number of cycles in that trajectory). The sample $\mathbf{x}_{u,t}\in \mathbb{R}^{24}$ is the feature vector of trajectory $u$ at cycle $t$. 

Each engine trajectory starts with different degrees of initial wear and manufacturing variation; detailed information about this is not available. The engine operates normally at the start of each time series. At a random time during each trajectory, a fault begins to develop and grow until the system fails.

Examples of sensor readings (features 2 and 7) from different subsets are shown in Table~\ref{tab:Dataset}. They illustrate the differences between trajectories with a single operating condition and six operating conditions. During an engine trajectory, the engine may change operating condition between cycles but each cycle corresponds to only one operating condition.

\begin{figure*}
\centering
\subfigure[]{%
\includegraphics[width=0.30\textwidth]{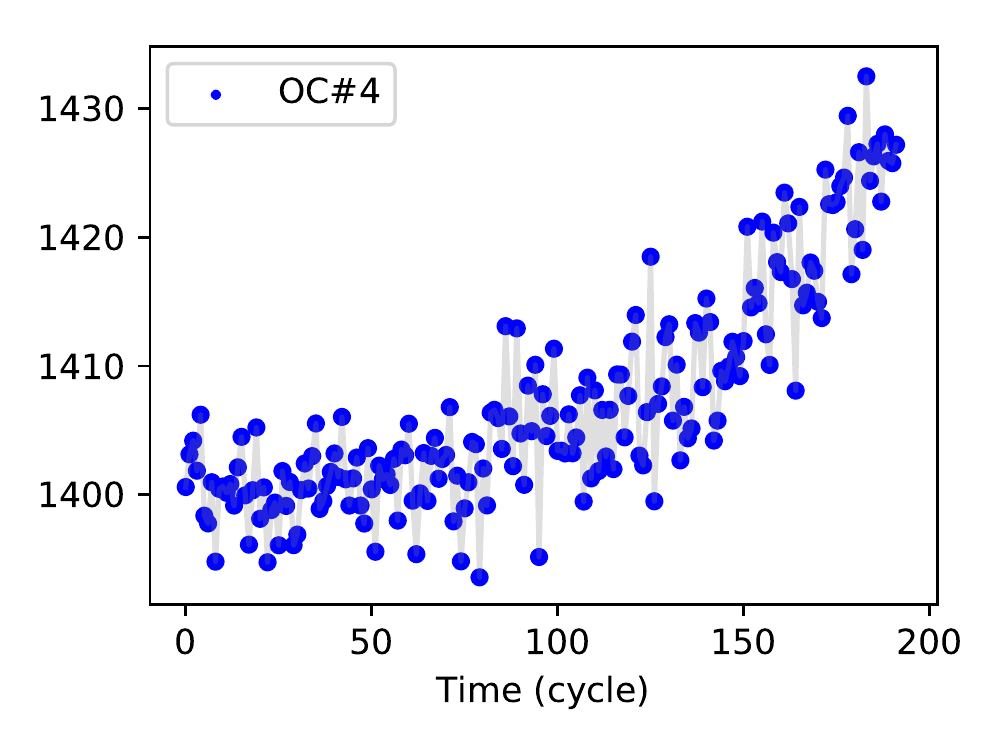}%
\label{fig:sig1}%
}
\subfigure[]{%
\includegraphics[width=0.30\textwidth]{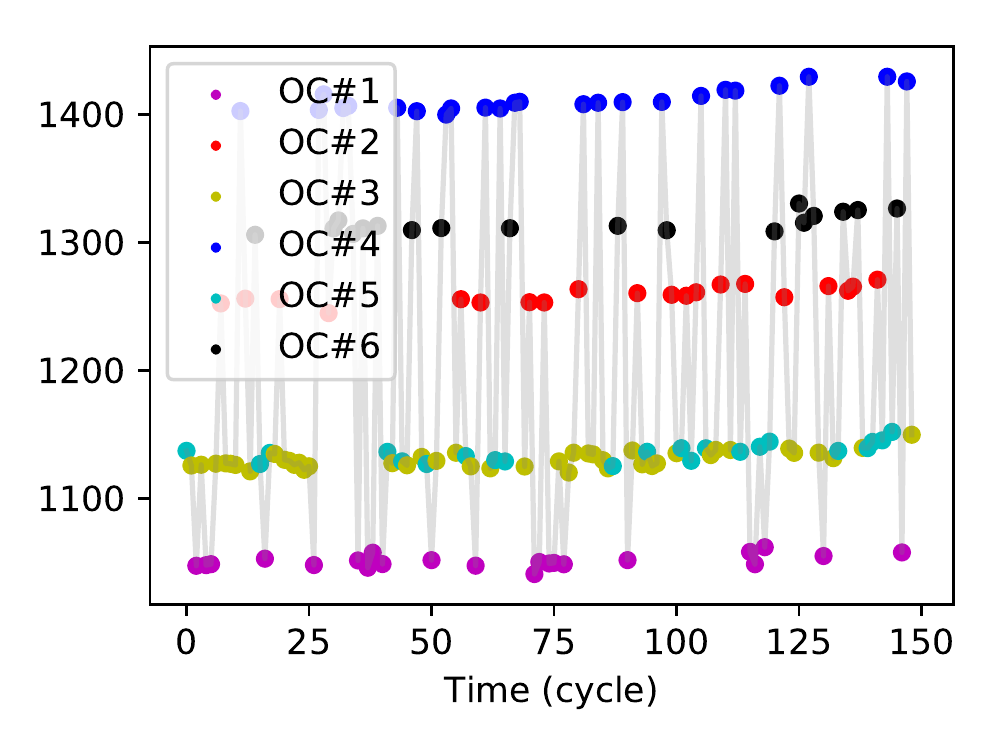}%
\label{fig:sig2}%
}
\subfigure[]{%
\includegraphics[width=0.30\textwidth]{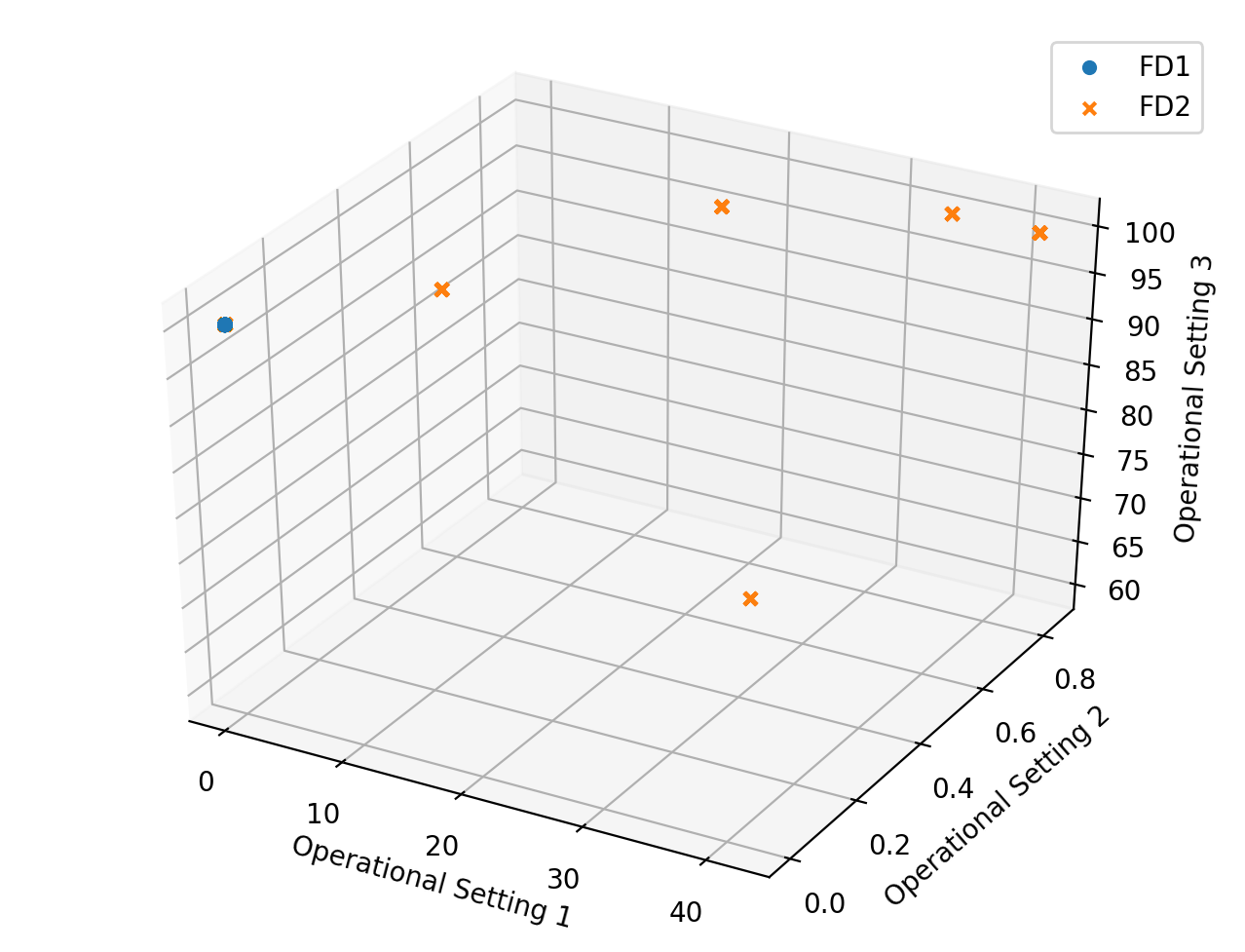}%
\label{fig:operating_condition}%
}
\caption{\ref{fig:sig1} and \ref{fig:sig2}: Difference in time series (the $8$-th feature) between example trajectory from $X^1$ (single operating condition) and $X^2$ (multiple operating conditions); \ref{fig:operating_condition} shows all six operating conditions with the first three features.}
\label{fig:signals_diff} 
\end{figure*}

\newcommand{\tableSubplotSizeA}{0.18}
\begin{table*}
\begin{center}
\begin{tabular}{l*{4}{c}r}
\toprule
Dataset            & FD001 $X^1$ & FD002 $X^2$ & FD003 $X^3$ & FD004 $X^4$ \\
\midrule
Train trajectories $X_\alpha$ & 100 & 260 & 100 & 248 \\
Test trajectories $X_\beta$ & 100 & 259 & 100 & 249 \\ \midrule
$X_\alpha$ Life Span &
\begin{minipage}{\tableSubplotSizeA\textwidth}
      \includegraphics[width=\textwidth]{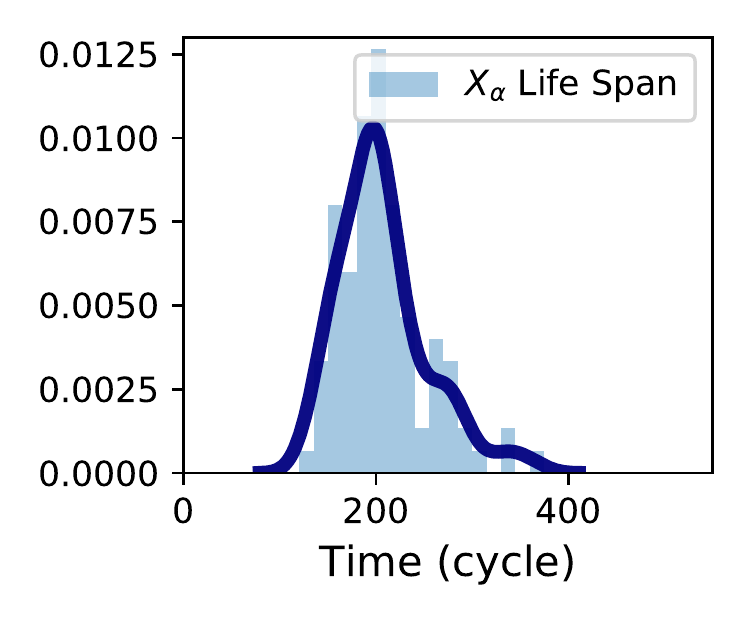}
    \end{minipage}& 
\begin{minipage}{\tableSubplotSizeA\textwidth}
      \includegraphics[width=\linewidth]{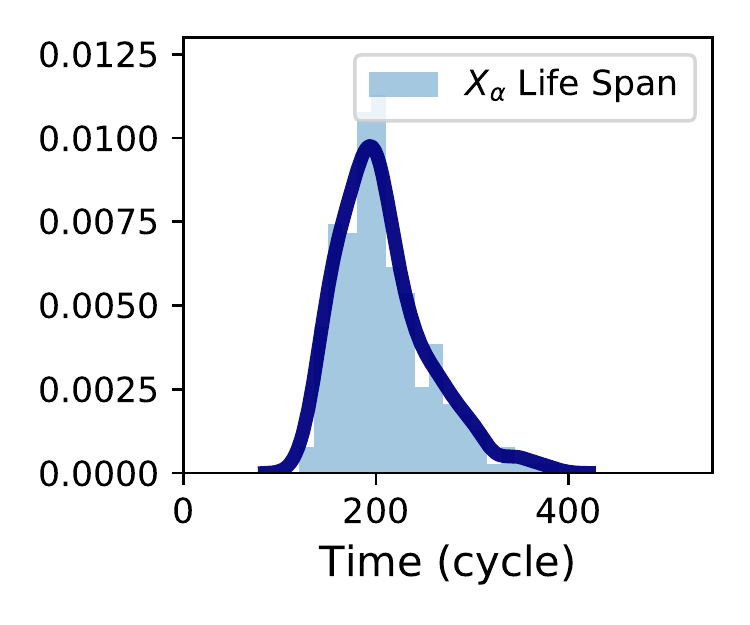}
    \end{minipage}&
\begin{minipage}{\tableSubplotSizeA\textwidth}
      \includegraphics[width=\linewidth]{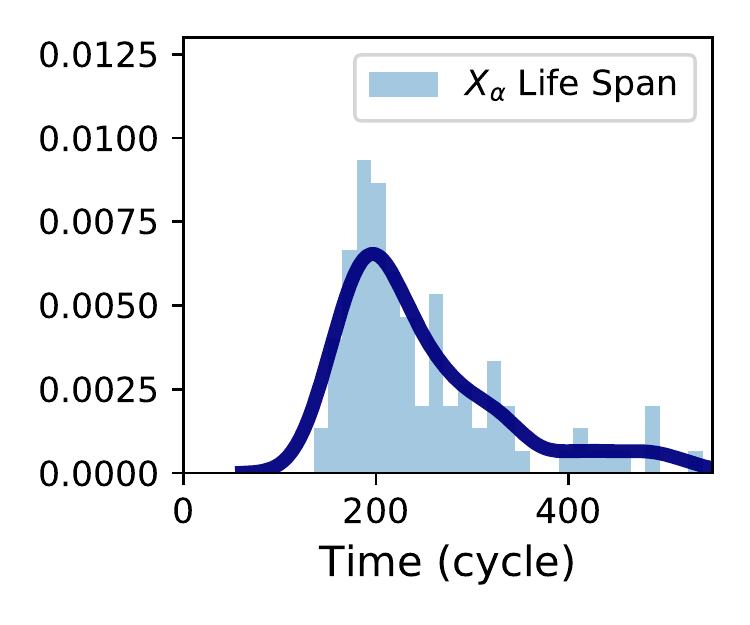}
    \end{minipage}&
\begin{minipage}{\tableSubplotSizeA\textwidth}
      \includegraphics[width=\linewidth]{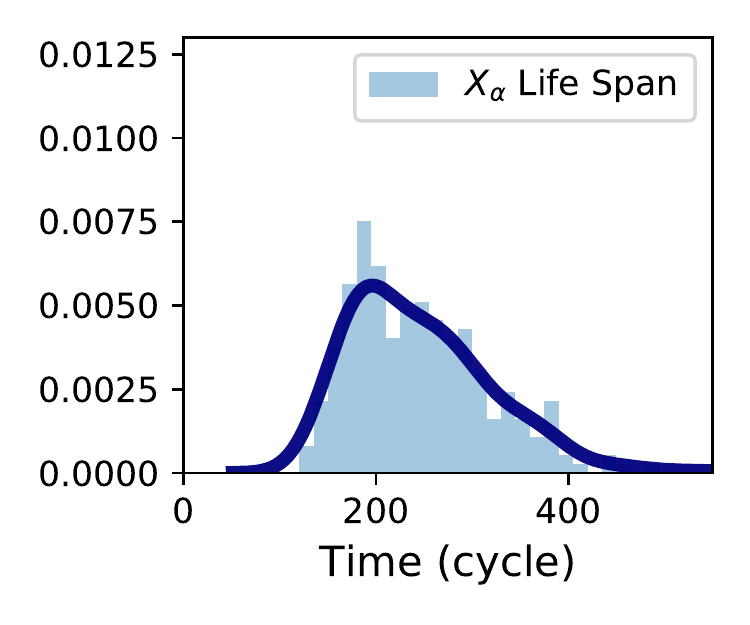}
    \end{minipage} \\
$X_\beta$ Life Span &
\begin{minipage}{\tableSubplotSizeA\textwidth}
      \includegraphics[width=\linewidth]{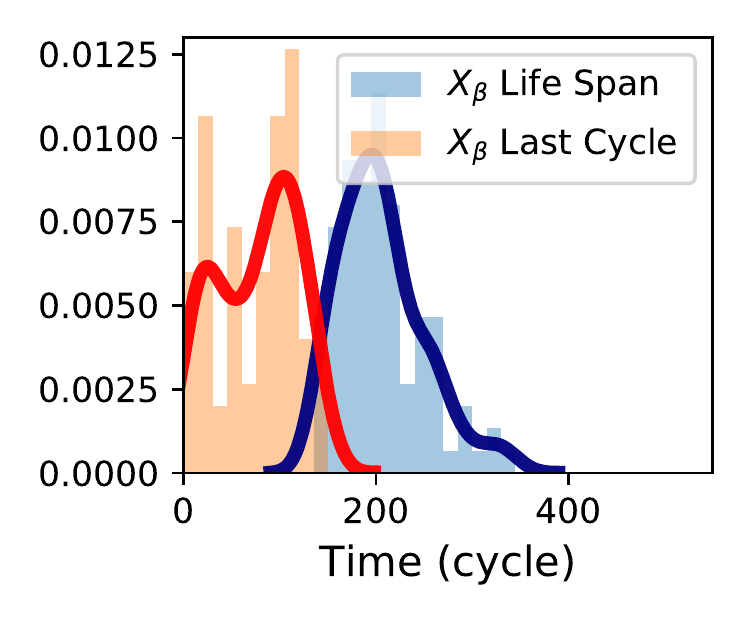}
    \end{minipage}& 
\begin{minipage}{\tableSubplotSizeA\textwidth}
      \includegraphics[width=\linewidth]{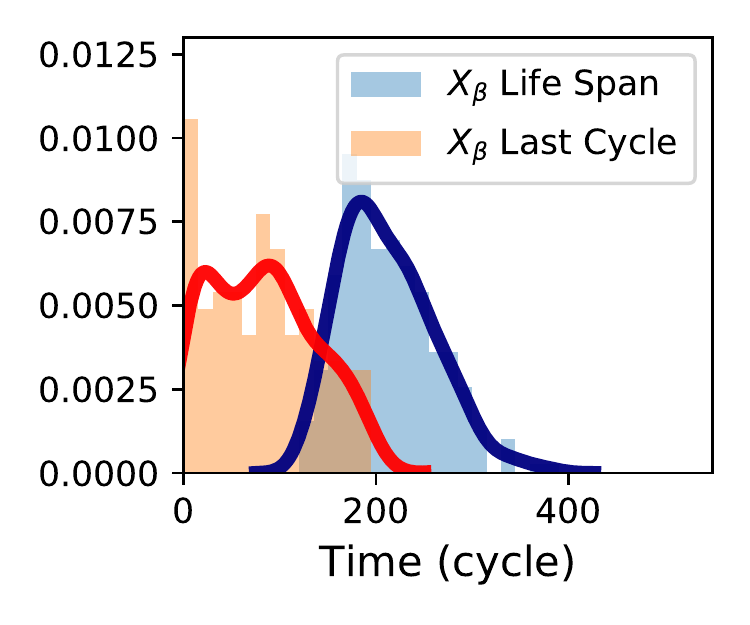}
    \end{minipage}&
\begin{minipage}{\tableSubplotSizeA\textwidth}
      \includegraphics[width=\linewidth]{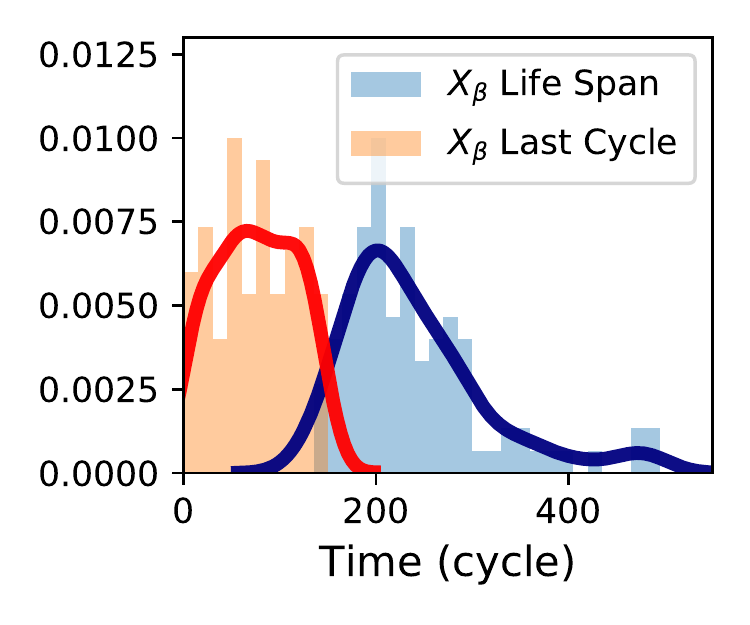}
    \end{minipage}&
\begin{minipage}{\tableSubplotSizeA\textwidth}
      \includegraphics[width=\linewidth]{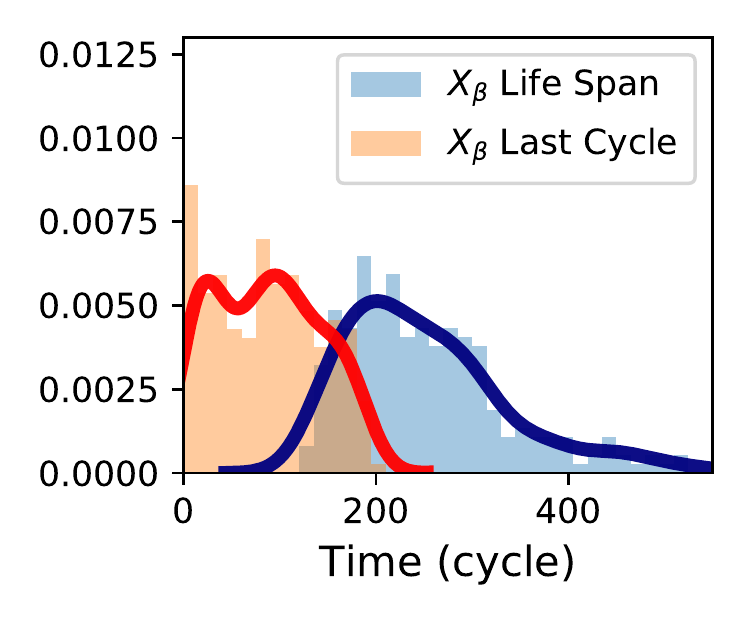}
    \end{minipage} \\ \midrule
Operating Conditions & 1 & 6 & 1 & 6 \\
Fault Modes        & 1 & 1 & 2 & 2 \\ \midrule
\shortstack[l]{Feature 2 of\\trajectory $x_{20}$ from $X^1_\alpha$} &
\begin{minipage}{\tableSubplotSizeA\textwidth}
      \includegraphics[width=\textwidth]{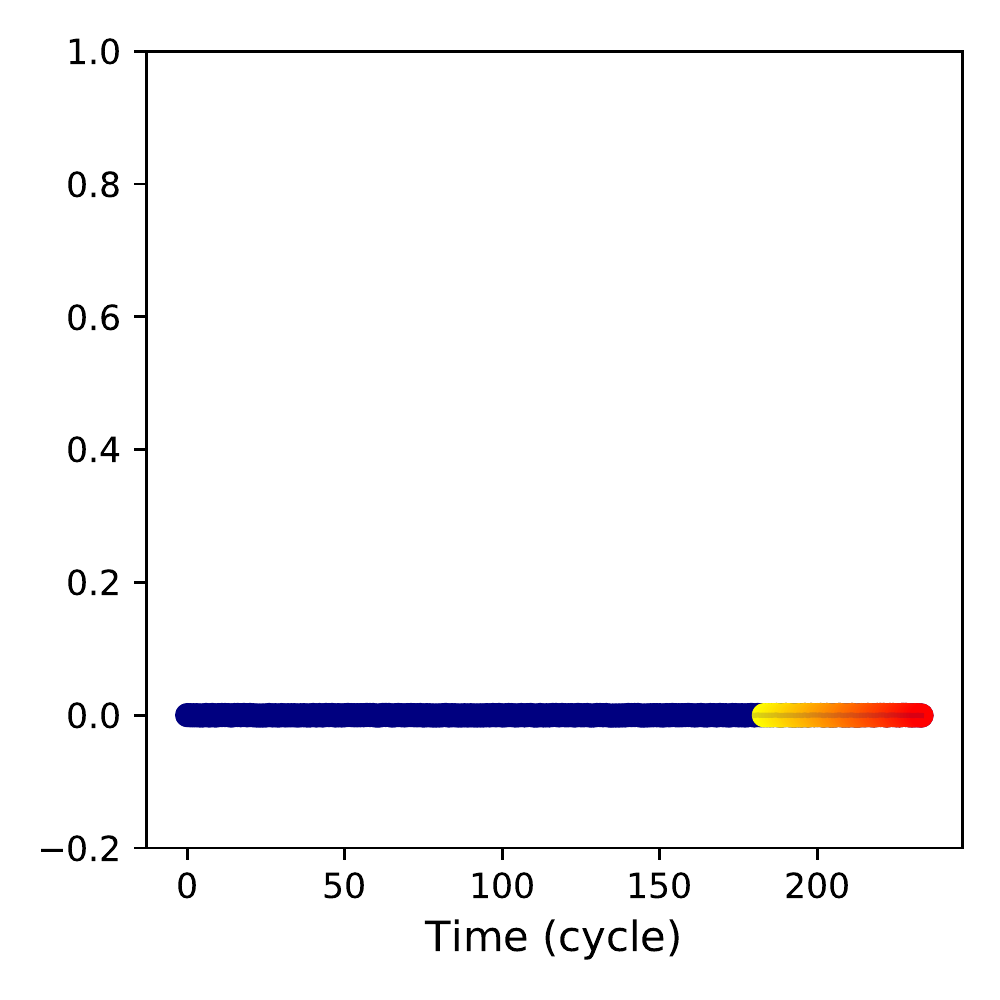}
    \end{minipage}& 
\begin{minipage}{\tableSubplotSizeA\textwidth}
      \includegraphics[width=\linewidth]{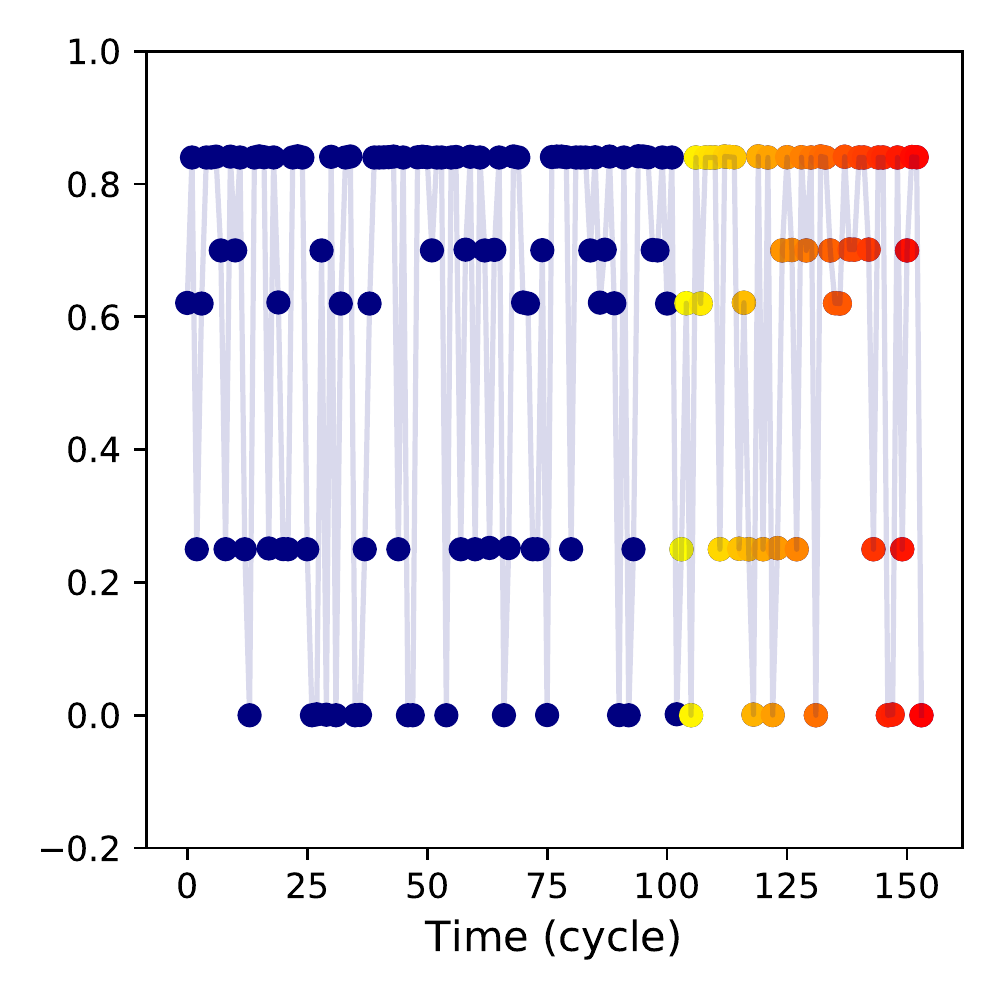}
    \end{minipage}&
\begin{minipage}{\tableSubplotSizeA\textwidth}
      \includegraphics[width=\linewidth]{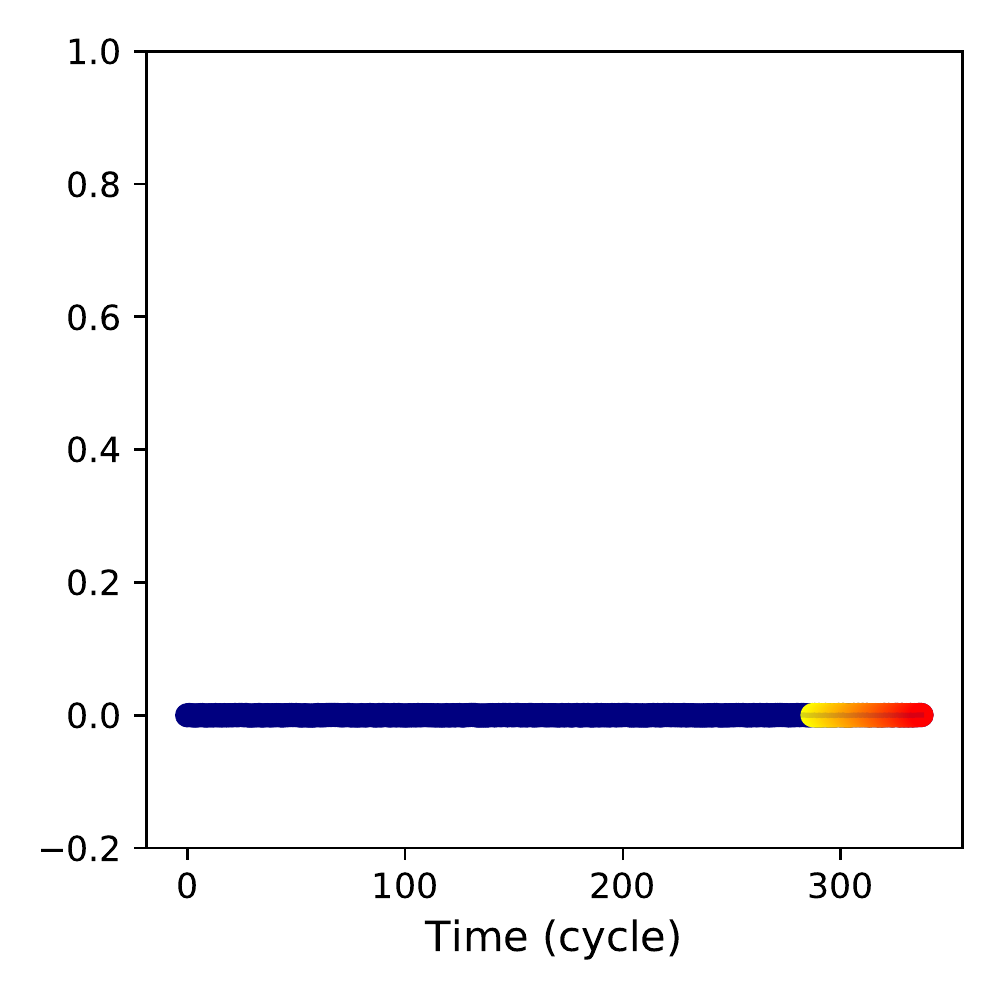}
    \end{minipage}&
\begin{minipage}{\tableSubplotSizeA\textwidth}
      \includegraphics[width=\linewidth]{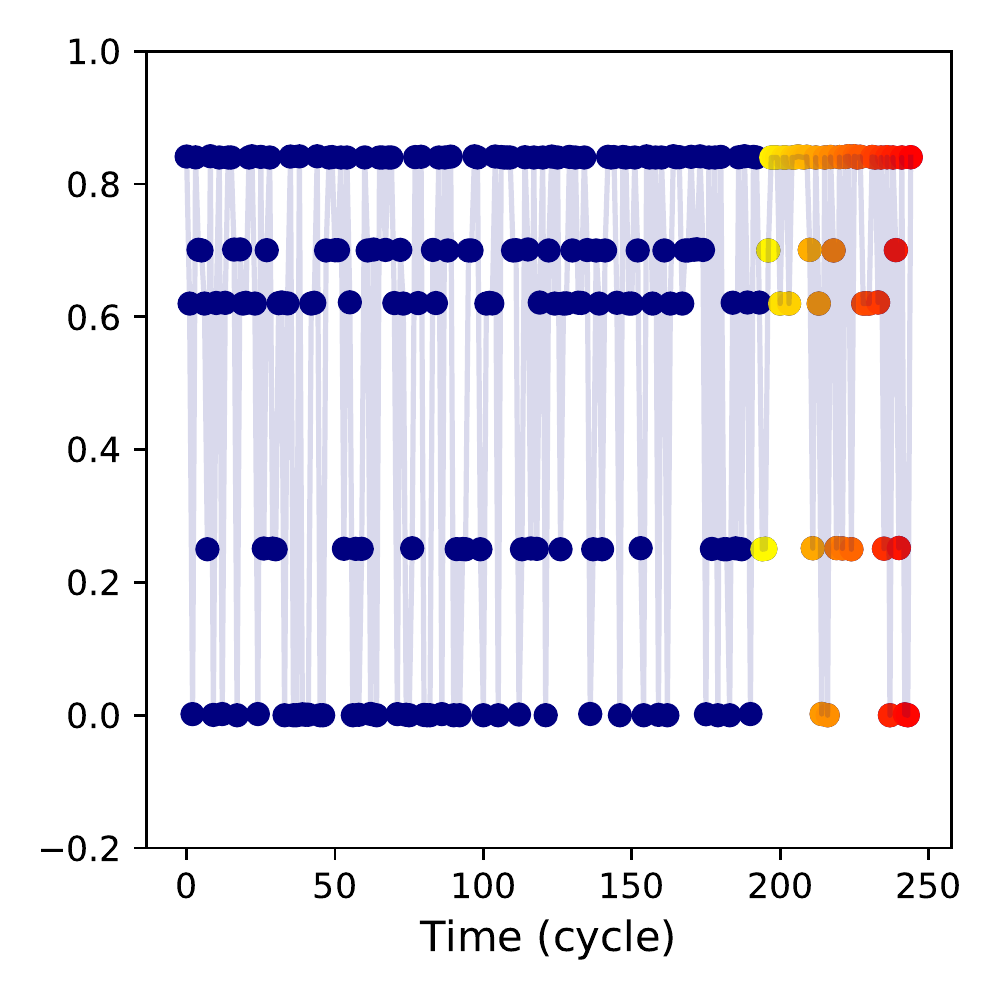}
    \end{minipage} \\
\shortstack[l]{Feature 7 of\\trajectory $x_{20}$ from $X_\alpha^4$} &
\begin{minipage}{\tableSubplotSizeA\textwidth}
      \includegraphics[width=\textwidth]{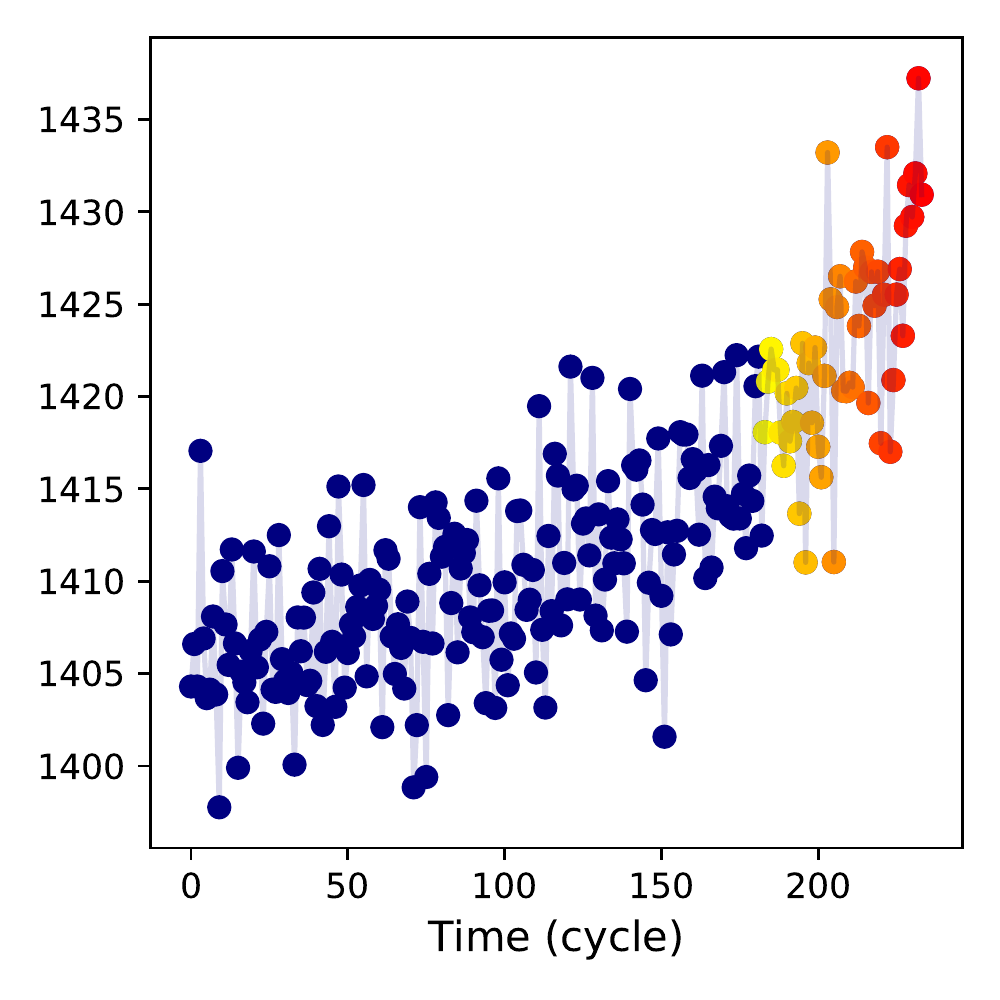}
    \end{minipage}& 
\begin{minipage}{\tableSubplotSizeA\textwidth}
      \includegraphics[width=\linewidth]{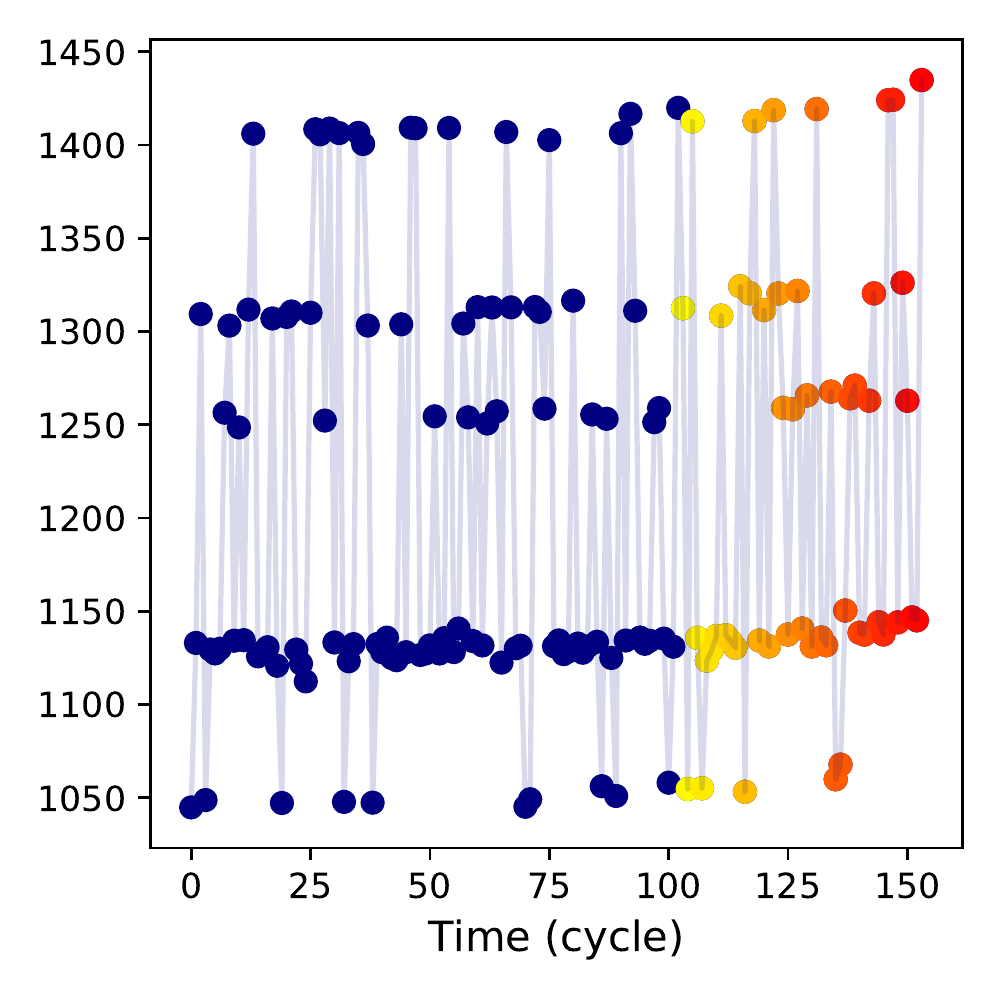}
    \end{minipage}&
\begin{minipage}{\tableSubplotSizeA\textwidth}
      \includegraphics[width=\linewidth]{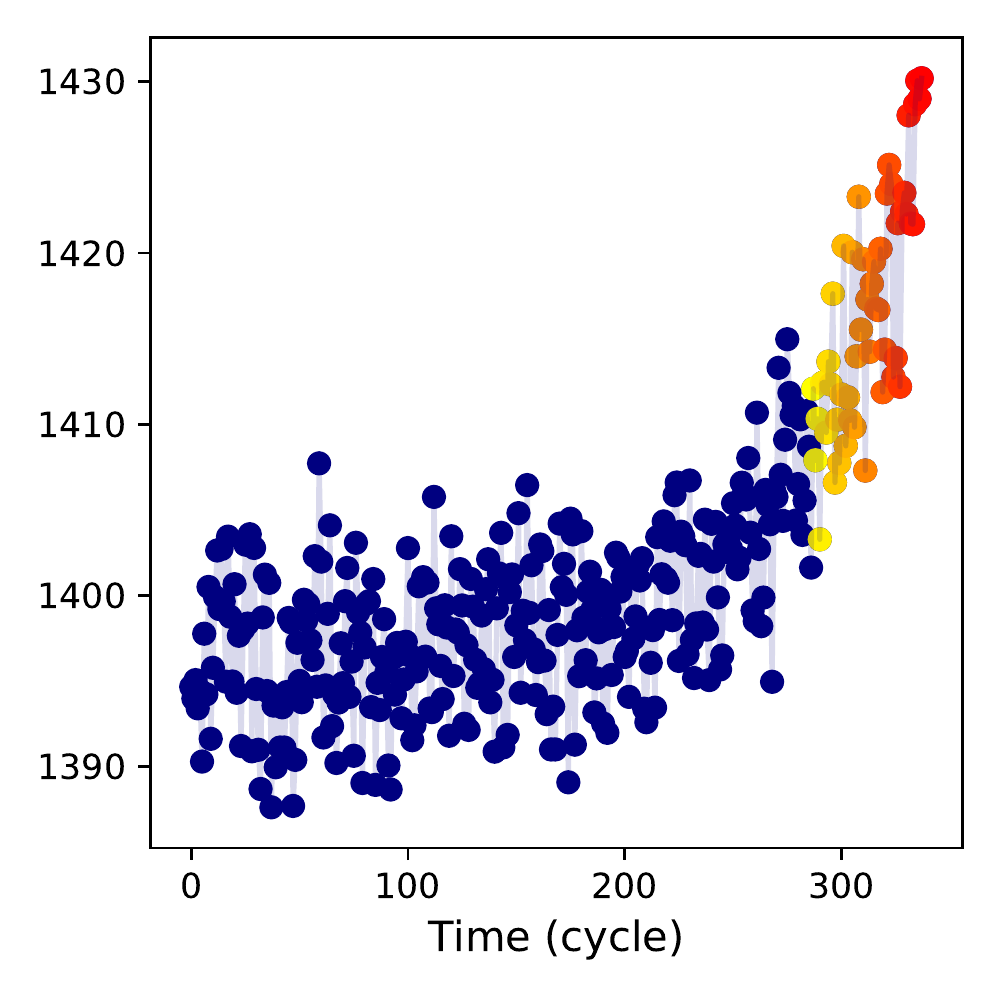}
    \end{minipage}&
\begin{minipage}{\tableSubplotSizeA\textwidth}
      \includegraphics[width=\linewidth]{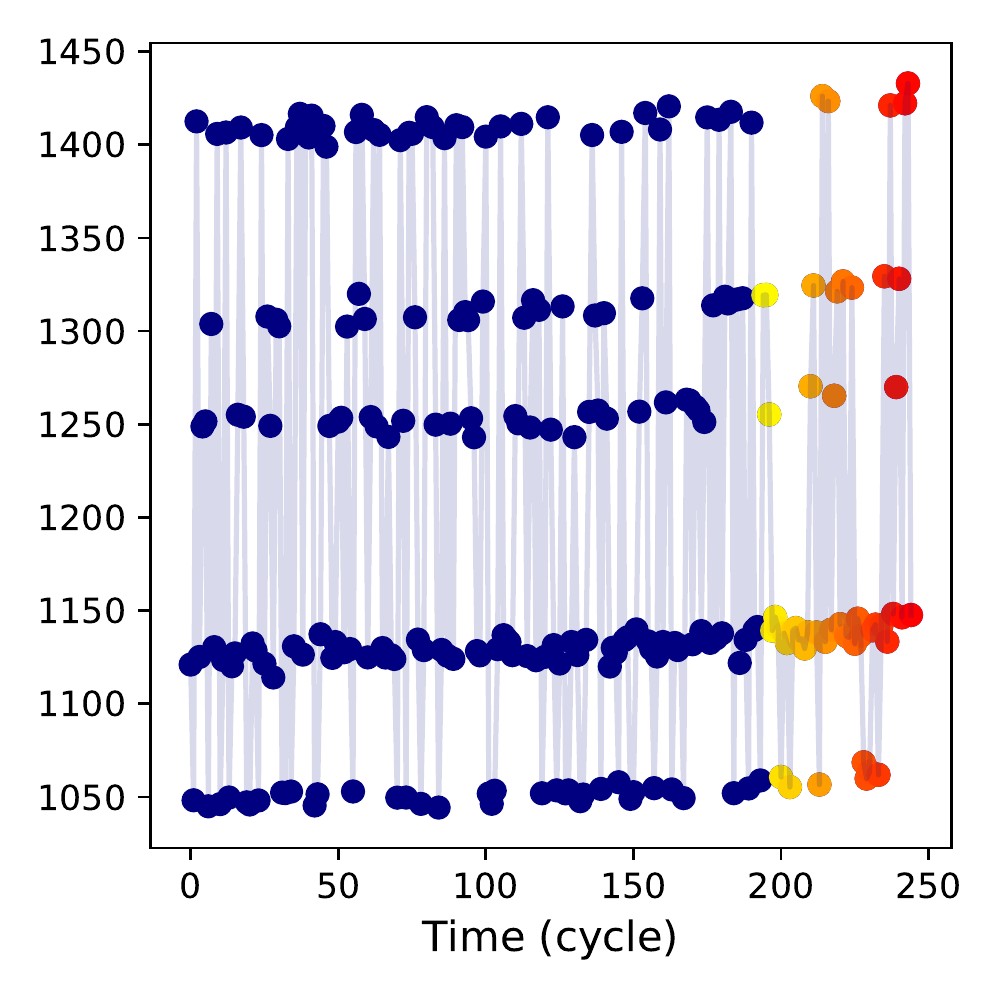}
    \end{minipage} \\
\shortstack[l]{T-SNE  \\ $10$ trajectories} & 
\begin{minipage}{\tableSubplotSizeA\textwidth}
      \includegraphics[width=\linewidth]{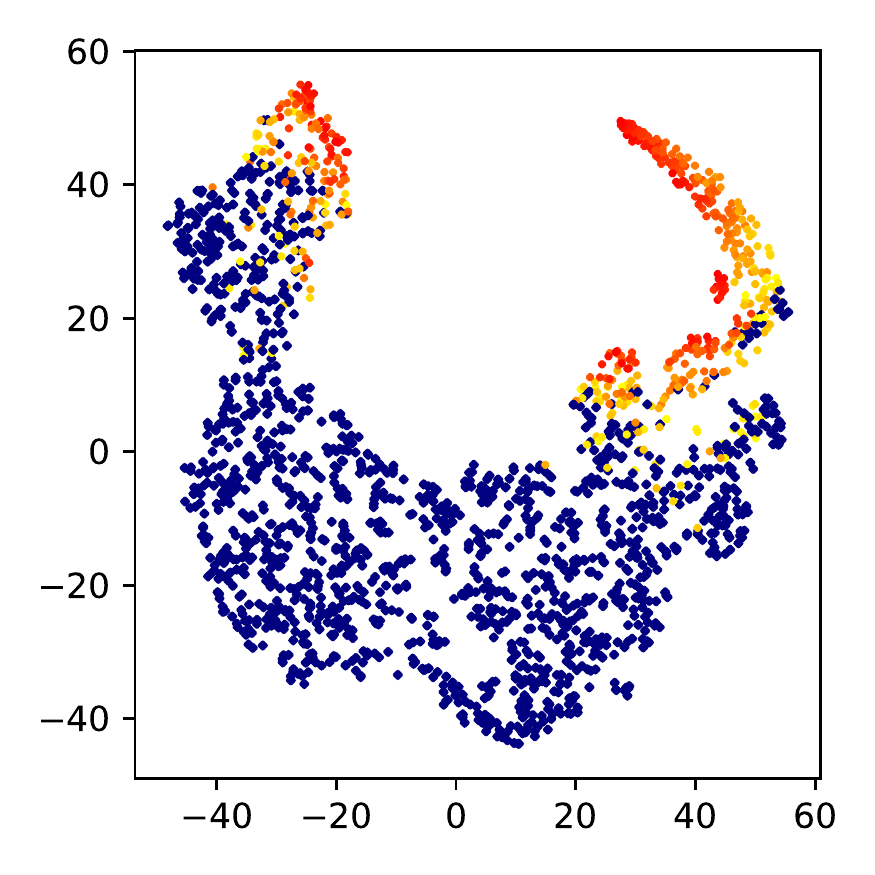}
    \end{minipage}& 
\begin{minipage}{\tableSubplotSizeA\textwidth}
      \includegraphics[width=\linewidth]{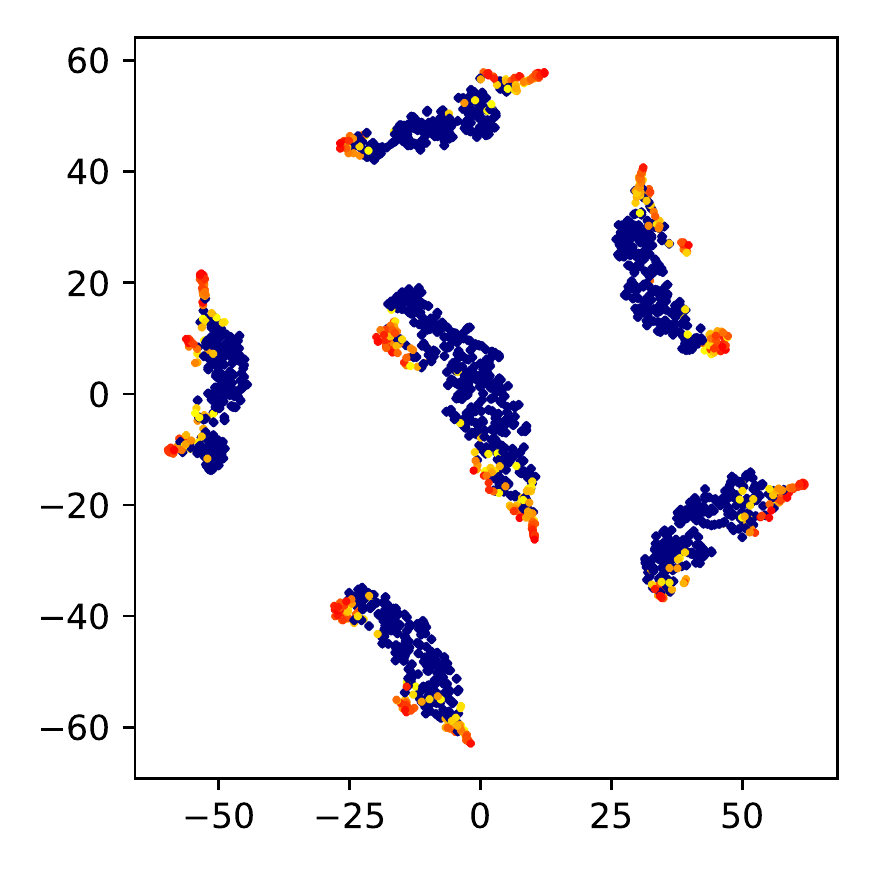}
    \end{minipage}&
\begin{minipage}{\tableSubplotSizeA\textwidth}
      \includegraphics[width=\linewidth]{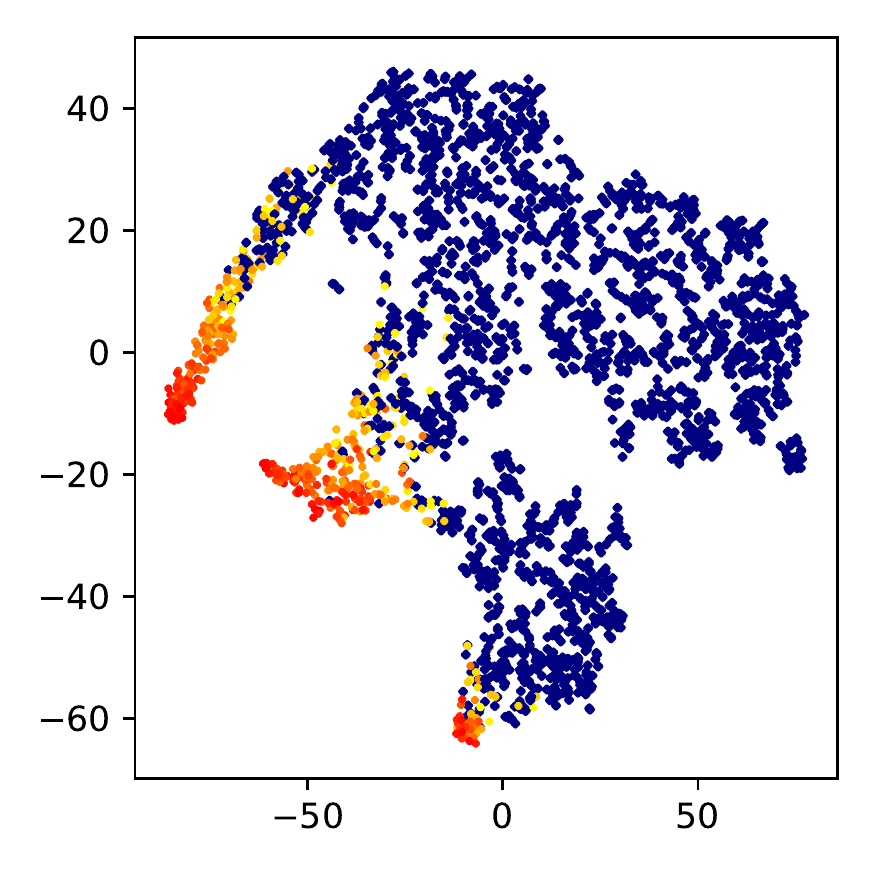}
    \end{minipage}&
\begin{minipage}{\tableSubplotSizeA\textwidth}
      \includegraphics[width=\linewidth]{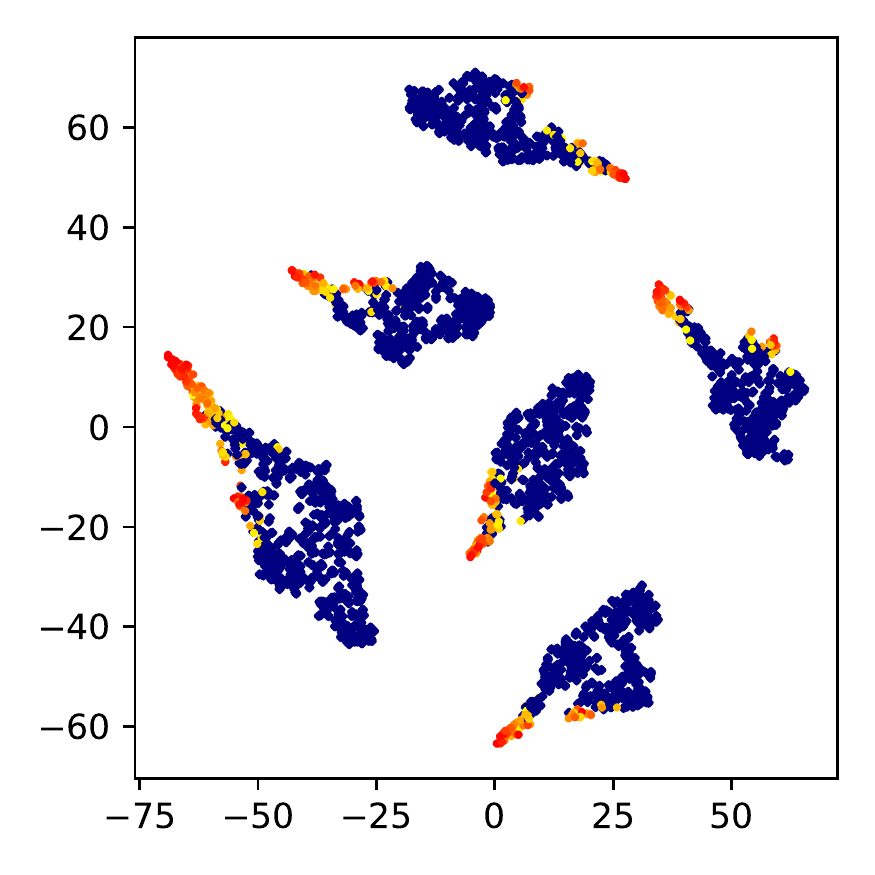}
    \end{minipage} \\
\bottomrule
\end{tabular}\caption[l]{C-MAPSS dataset: Life Span population of the training and the testing subset is illustrated; two sample features of one unit from the four subsets were shown; T-SNE visualizations of $10$ degradation trajectories from FD001 and from FD004 were shown on the last row, where blue dots mark out observations that have RUL larger than 50 cycles and the rest (yellow to red) are from observations from RUL 50 cycles to 0.}
\label{tab:Dataset}
\end{center}
\end{table*} 
Figure~\ref{fig:operating_condition} shows the first three features, i.e. Altitude, Mach Number, and Throttle Resolver Angle \cite{saxena2008damage}, of all samples within subsets $X^1$ and $X^2$. There are six distinct groups of samples in this space -- each corresponding to one operating condition. Furthermore, values of feature $7$ from trajectory $\mathbf{x}_{20}$ in $X^1_\alpha$ and $X^2_\alpha$ are shown in Fig.~\ref{fig:sig1} and Fig.~\ref{fig:sig2}. They illustrate that sensor readings under a single operating condition reside in a sub-part of the sensor readings under six operating conditions. 

We used t-distributed stochastic neighbor embedding (t-SNE) \cite{maaten2008visualizing} on each data subset to visualize how samples close to end of life (EOL) were distributed with respect to other samples. Each figure in the bottom row of Table~\ref{tab:Dataset} is the result of applying t-SNE on ten run-to-failure trajectories from the training sets $X_\alpha$. The blue dots mark observations that have RUL $> 50$ cycles, i.e. ``healthy'' engines, and the rest (yellow to red) mark observations where RUL $\leq 50$ cycles. Red marks observations at or very near EOL, which tend to be on the edge of the clusters.

The turbofan data sets (including an extra set for the PHM 2008 Challenge) generated by C-MAPSS have been widely used for testing and analyzing prognostics methods; a benchmark report by Saxena et al. \cite{ramasso2014performance} is available. Prognostic modeling efforts in most of the research studies \cite{yongxiang2016data, peng2012modified, medjaher2012remaining, rigamonti2016echo, heimes2008recurrent, wang2008similarity, ellefsen2019remaining, babu2016deep, li2018remaining, zheng2017long} have been dedicated to predicting RUL when training and testing data correspond to the same conditions, i.e. prognostic models were built using $X_\alpha^i$ and predictions done on $X_\beta^i$ from the same subset $X^i$.
In general, the prognostic modeling approaches taken on this problem include \cite{ramasso2014performance}: $i$) mapping between a set of sensor input and RUL; $ii$) mapping between an approximated health index (one-dimensional variable) and RUL; $iii$) Similarity-based matching.
Over the years, the prediction performances have improved; a summary is shown in Table~\ref{tab:SOA_RMSE_tab}. 
Many prognostic methods have been based on neural networks. One of the two winner methods from the PHM 2008 challenge, by Hemies et al. \cite{heimes2008recurrent}, used recurrent neural networks (RNNs) to capture temporal information from the multivariate sensor readings and learn the complex system dynamics for predicting RUL. Echo state networks (ESNs), with similar characteristics as RNNs, have been applied to perform the prognostic modeling as well \cite{rigamonti2016echo, peng2012modified}. Zheng et al. \cite{zheng2017long} applied RNNs with long short-term memory (LSTM) to predict RUL. 
Another approach to model system degradation, besides learning a direct mapping function between the sensor input and RUL, is to construct an intermediate scalar feature, such as a health index (HI) or degradation index. This index should capture the degradation pattern of the equipment for RUL prediction. The objective is to learn two mapping functions: the first one maps sensor data to the index and the second one maps this index to RUL. Various techniques, e.g. stochastic modeling, neural networks, and distance-based approaches have been employed. 
Le Son et al. \cite{le2012remaining} proposed, to use a Gamma process with Gaussian noise to model the degradation indicator of the equipment for RUL prediction. Liu et al. \cite{liu2013data} proposed a data-level fusion approach for generating health indices with exponential models. Le Son et al. \cite{le2013remaining} proposed to estimate RUL by simulating a Wiener process based on a degradation path that is generated using distance to the center of failure (EOL) sample in the PCA space. Zhao et al. \cite{zhao2017remaining} proposed to learn the degradation pattern with adjacent difference neural networks.

The third type of approach is to weigh trajectories differently based on the similarity in the degradation pattern for training the mapping function between sensor data and RUL. Wang et al. \cite{wang2008similarity} used this idea to create a library of degradation patterns based on a health index fused using multivariate sensor data. For each testing sample, the predictions were cast using a model that was trained only with trajectories that had similar degradation patterns to the one observed. 

The purpose of our study is to introduce and demonstrate a method that translates well when predicting between different data subsets, not to produce the best predictions on the same data set. Hence, we have used a robust yet effective model, the RF regressor, to map from data $\mathbf{x}$ to RUL. We used the scikit-learn library for this \cite{scikit-learn}. Our RF regression model, trained with raw features without any data pre-processing, achieves very similar performance to the RF model reported by Zhang et al. \cite{zhang2017multiobjective} (see Table~\ref{tab:SOA_RMSE_tab}).

There have been some previous work done on learning to predict between data subsets (i.e. to transfer the knowledge). The most straightforward has been to incorporate the three features with information about the operating conditions (Altitude, Mach Number, and TRA). Zhang and Zhao \cite{zhang2018transfer, zhao2017remaining} used a one-hot encoding of the six operating conditions. Le Son et al. \cite{le2013remaining} computed a degradation index based on failure samples in the same operating condition as the sample they predicted for.

Ellefsen et al. \cite{ellefsen2019remaining} pointed out that high-quality labeled training data is hard to acquire and performed a study assuming that labeled data are available for a limited amount of trajectories (both in training and testing sets). They used a semi-supervised approach to first train a recurrent model with unlabeled data and afterwards fine tune it with a limited amount of labeled trajectories. They have achieved the best performance in two out of four subsets, see Table~\ref{tab:SOA_RMSE_tab}. In another recent work, Zhang et al. \cite{zhang2018transfer} conducted a study based on the assumptions that $i$) the population of the training set and test set are different, and $ii$) labeled data in both domains may be limited. Parameter transfer was performed based on an LSTM network under the setting of inductive transfer learning. The trained parameters of an LSTM network based on trajectories from the source domain are transferred and fine-tuned with trajectories from the target domain. The work demonstrated that labeled data coming from a similar domain is useful for achieving better performance in some transfer learning scenarios. Both studies (Ellefsen et al. and Zhang et al.) have addressed the problem that labeled data is limited and their approaches can be used when some (a few) labeled trajectories are available in the target domain (i.e. when in \emph{Phase D} in Fig.~\ref{fig:TL_for_PHM}). 
Recently, Da Costa et al. \cite{da2019remaining} applied an LSTM network to capture temporal information and used DANN to learn domain-invariant features for predicting RUL.
According to their examples, the proposed LSTM-DANN approach is unable to reflect the deteriorating condition of engines under new operating conditions.
Nevertheless, the proposed LSTM-DANN approach has provided more reliable predictions in other TL scenarios.

In summary, there are many studies that have developed and tested prognostic methods on the C-MAPSS subsets. They are almost always based on the assumption that training and testing samples are of the same population. Few have addressed the need for generalizing outside of the data; that, in practice, test samples and training samples may come from different populations. Of those who do generalize across data sets, few have considered the case that labeled test samples may not be available (ever), which is the challenge we meet with our approach. The four CMAPSS subsets resemble a very good case study for TL. Table~\ref{tab:transfer_learning_scenarios} presents in total $16$ learning scenarios, $12$ of them are TL scenarios where the source and the target domain is different.

\begin{table*}
\begin{center} 
\begin{tabular}{l*{4}{c}r}
\toprule
& FD001  & FD002  & FD003  & FD004 \\ 
& 1 OC 1 Fault  & 6 OCs 1 Fault  & 1 OC 2 Faults  & 6 OCs 2 Faults \\  \midrule
FD001  & Same population & \textbf{New Operating Conditions (OCs)} & \textbf{New faults} & \textbf{New faults \& New OCs} \\
FD002  & Fewer OCs & Same population & \textbf{New faults} \& Fewer OCs & \textbf{New faults} \\
FD003  & Fewer faults & Fewer faults \& \textbf{New OCs} & Same population & \textbf{New OCs} \\
FD004  & Fewer faults \& Fewer OCs & Fewer faults & Fewer OCs & Same population \\
\bottomrule
\end{tabular}
\end{center}
\caption{Learning Scenarios of CMAPSS dataset, with row subset selected as the source (training) data and column subset selected as the target (testing) data.}
\label{tab:transfer_learning_scenarios} 
\end{table*}

\subsection{Training Sequences for RUL Prediction}

The training sets $X_\alpha$ are run-to-failure trajectories, i.e. the last cycle of each trajectory is the EOL. Ideally, the constructed RUL teaching sequences should reflect the degradation pattern of the trajectories. This presents some challenge, since the degradation begins when a fault starts to develop. It is not a good idea to assume that the degradation rate is constant throughout the trajectory.
Based on the analysis by Heimes et al. \cite{heimes2008recurrent}, we use a piecewise linear target signal: 
\begin{equation}
y_{u,t} = 
\begin{cases}
l(u) - t &  \mbox{if } l(u) - \tau_{max} \leqslant t \leqslant \ l(u) \\
\tau_{max}  & \mbox{otherwise}
\end{cases} \label{equ:equ_RUL_model}
\end{equation}
where $\tau_{max}$ is set to $130$. This form for the target signal is used in many studies \cite{zhang2017multiobjective, zheng2017long, li2018remaining, babu2016deep, ellefsen2019remaining, zhang2018transfer}, with minor differences in the value for $\tau_{max}$.

\section{Transfer Learning with Consensus Self-Organizing Models}
\label{sec:method}

The consensus self-organizing models (COSMO) approach for detecting deviations on equipment builds on first creating a representation of the equipment, then comparing distances between the representations and look for consistent outliers \cite{ByttnerRS-11, rognvaldsson2018self}. The representations can e.g. be individual samples, distributions of samples, relationships between samples, et cetera.

\begin{figure*}
\centering
\includegraphics[width=0.90\textwidth]{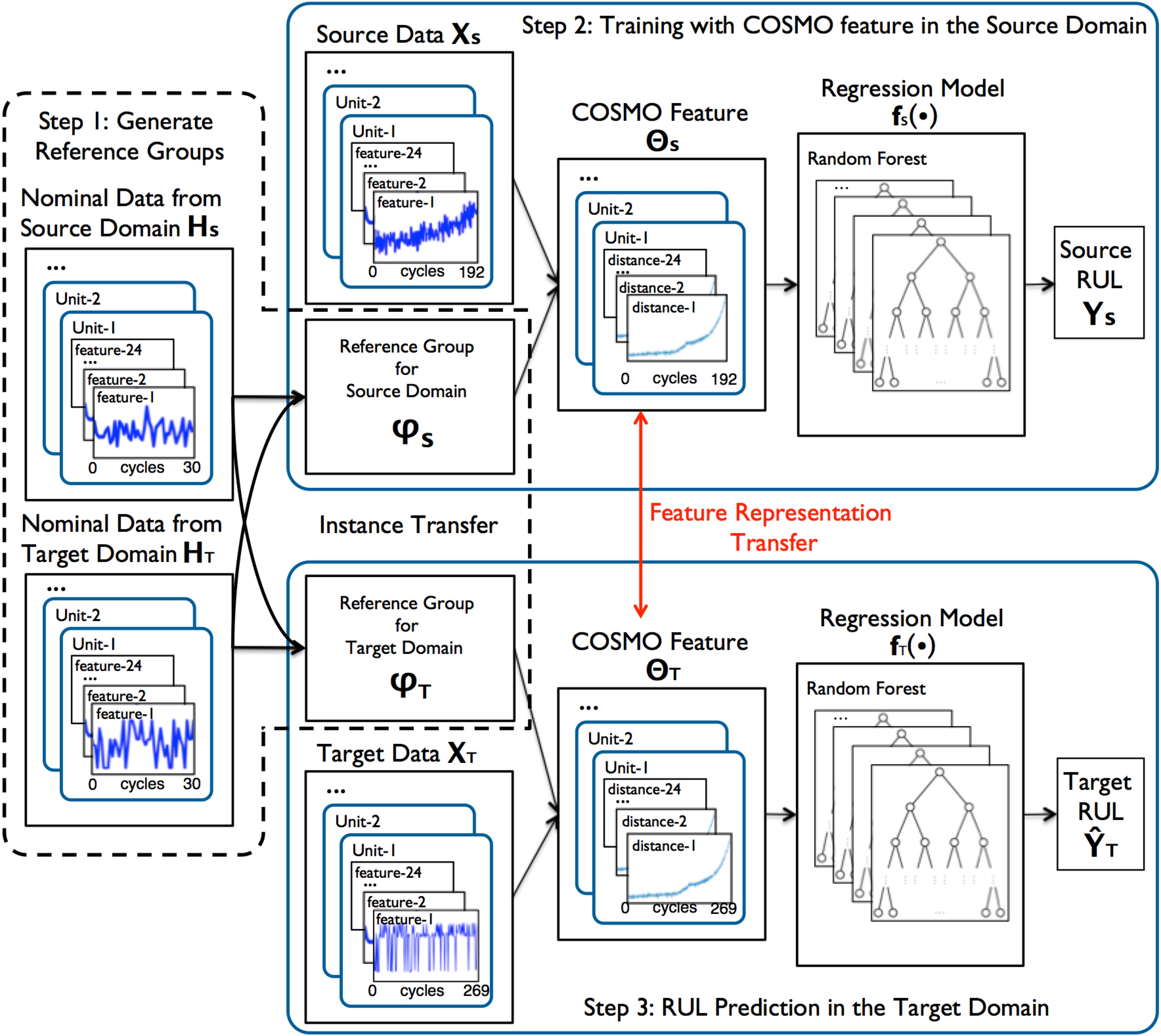}
\caption{Transfer Learning with COSMO Method}
\label{fig:COSMO_TL}
\end{figure*}

Applying TL on prognostics with the COSMO method consists of three steps (illustrated in Figure~\ref{fig:COSMO_TL}): \emph{i}) Generating reference groups, $\Phi_S$ and $\Phi_T$, from the source data $X_S$ and the target data $X_T$; \emph{ii}) computing COSMO features, $\Theta_S$ and $\Theta_T$, for samples from the two domains and use $(\Theta_S, Y_S)$ to train a regression model $f_S(\cdot)$; \emph{iii}) performing parameter transfer, $f_S(\cdot) \rightarrow f_T(\cdot)$ and predict remaining useful life $\hat{Y_T}$ for samples in the target domain with COSMO features $\Theta_T$, instead of $X_T$, as the input data to the regression model.

\subsection{Reference Group Generation}\label{subsec:rgg}

A COSMO feature is generated based on the distance between an individual sample and its peers, i.e. samples in the reference group. The reference group provides a normal variation of nominal units (or at least mostly normal units). Ideally, the reference group should contain samples from all operating conditions. This can be achieved by incorporating explicit knowledge of what each operating condition looks like or if there is any indicator signal (but such knowledge might be difficult to acquire in a real-world application). 

According to the description of the C-MAPSS data, all trajectories start with some random initial wear and then deteriorate over time. We assume that a small number of cycles $\tau$ at the beginning of these trajectories correspond to fairly ``healthy'' systems and use these samples as the nominal data samples $H$:
\begin{eqnarray}
H_S &= \bigl\{\, \mathbf{x}_{u, t} \in X_S  \bigm|  t \leqslant \tau \,\bigr\} \\
H_T &= \bigl\{\, \mathbf{x}_{u, t} \in X_T  \bigm|  t \leqslant \tau \,\bigr\}
\end{eqnarray}
where, nominal data samples of the source data $X_S$ are denoted as $H_S$ and similarly for the target domain. We use $\tau = 30$ (i.e. the first 30 cycles are considered ``healthy''). 

A reference group ($\varphi$) for the source and the target domain can then be generated with four possibilities. As an example, $\mbox{Mode }(S,T)$ and $\mbox{Mode }(S,ST)$ are shown as follows:
\begin{eqnarray}\label{equ:equ_RUL_teaching}
\mbox{Mode }(S,T) &:&
\begin{cases}
\varphi_S &= H_S  \\
\varphi_T &= H_T
\end{cases} \\
\mbox{Mode }(S,ST) &:&
\begin{cases}
\varphi_S &= H_S \\
\varphi_T &= H_S \cup H_T
\end{cases}
\end{eqnarray}
The first letter in the ``mode'' denotes the reference group used when building the RUL model (before deployment), the second letter denotes the reference group used when predicting RUL (after deployment); $S$ denotes source data and $T$ denotes target data. Note that the reference group data are always unlabeled.
Among all possibilities for selecting a reference group $\mbox{Mode }(S,T)$ and $\mbox{Mode }(S,ST)$, are the most practical. In this paper, reference group $\mbox{Mode }(ST,ST)$ and $\mbox{Mode }(ST,T)$ are also investigated. They only require source domain data during model construction. The other two require that (unlabeled) target domain data is also available when the model is built.

\subsection{Computing COSMO Features}\label{subsec:cosmo_features}

The COSMO approach builds on having a representation of the current operation of a system, e.g. distributions of its sensor values or models of how they relate to each other, measuring the distance between the observed system and peer systems, and repeatedly estimating the probability for the system to be inside the peer group distribution. Key ingredients in COSMO are how the system is represented, and how the distance between these representations is measured. Part of the COSMO approach builds on the concept of nonconformity introduced by Vovk et al. \cite{Vovk-02, VovkGS-05, shafer2008tutorial}. 


In this work, each sensor value was treated as a representation and the distances between sensor readings from different engines were measured with the $L_1$ norm. 
Three different ways for measuring the distance to the peers were tried: average distance to the $k$ nearest neighbors in the reference group ($k$NN), median distance to the $k$ nearest neighbors in the reference group (m-{\it k}NN), and distance to the most central pattern (MCP) in the reference group. R\"ognvaldsson et al. \cite{RognEtAl-14} showed that the MCP method works fine for unimodal distributions, but that complex distributions (e.g. multimodal) require something like a $k$NN distance. The turbofan engine data is multimodal when there are many operation conditions, and the m-$k$NN approach is therefore used in the main paper, but results for $k$NN and MCP are shown in the complementary material. The $k$NN results are very similar to the m-$k$NN results.

The proposed COSMO feature $\Theta_{u, t}$ for sample $x_{u, t}$ is a vector with the same dimensionality:
\begin{align*}
\Theta_{u, t} &= \{\  \theta_{u,t}^j \ |\ j = 1, 2, ..., |\rchi| \}
\end{align*}
Where $\theta_{u,t}^j$ captures the difference in the $j$-th feature between sample $x_{u, t}$ and samples in the reference group $\varphi$, and $|\rchi|$ is the dimensionality of the feature space $\rchi$, in which sample $x_{u, t}$ resides.

The first step in computing $\theta_{u,t}^j$ is to compute the absolute-value norms between sample $x_{u, t}$ and all samples in the reference group $\varphi$:
\begin{eqnarray}
\Delta^j({x_{u,t}} , \varphi) = \ \bigcup_{i=1}^{|\varphi|}\ \left| x_{u,t}^j - \varphi_i^j \right|
\end{eqnarray}
where $\varphi_i^j$ is the $j$-th feature of the $i$-th sample in the reference group $\varphi$ and $|\varphi|$ is the number of samples in $\varphi$. 
The second step is to select the $k$ smallest values within $\Delta^j({x_{u,t}} , \varphi)$:
\begin{eqnarray}
\Delta_{(-k)}^j(\cdot) = \bigl \{ \, \delta \mid \left|\Delta^j(\cdot) \cap [0,  \delta]\right| \leq k, \forall \delta \in \Delta^j(\cdot) \bigr \} 
\end{eqnarray}
Where $\left|\cdot\right|$ is the set cardinality and $\Delta_{-k}^j(\cdot)$ contains absolute-value norms of the $j^{th}$ feature between sample $x_{u,t}$ to its $k$-nearest neighbors in the reference group $\varphi$.
Afterwards, the median to the $k$-nearest neighbors distance (m-{\it k}NN), which corresponds to the $j$-th feature of $x_{u, t}$, can be computed and selected as $\theta_{u,t}^j$:
\begin{equation}
\theta_{u,t}^j(x_{u,t}, \varphi, k) \: = \:
\mbox{median}(\Delta_{(-k)}^j(x_{u,t}, \varphi))
\label{equ:theta_cosmo_j}
\end{equation}
%
%
where $k$ is the number of nearest neighbors selected for computing $\theta_{u,t}^j$. 

To ensure that the $k$NN distances are computed based mostly on samples that come from the same operating conditions, the following condition needs to be satisfied:
\begin{equation}\label{eq:knn-condition}
k \leq \frac{|\varphi|}{|\Omega_{oc}|}
\end{equation}
where $|\varphi|$ is the number of samples within the reference group and $|\Omega_{oc}|$ is an estimate of the number of different operating conditions in the data. From the data point of view, $|\Omega_{oc}|$ is equivalent and related to the number of distinct clusters in $X$. 

There are various methods for estimating the number of clusters. One approach is to compute eigenvalues of the Laplacian matrix of $X$ and use Eigengap heuristic \cite{von2007tutorial} to estimate the optimal number of clusters, which is usually given by the value that maximizes the difference between consecutive eigenvalues, i.e. eigengap.

The COSMO features $\Theta_S$ and $\Theta_T$, of the source and the target data (respectively), are inputs to the regression model for training the model as well as for predicting RUL. The RF regression model in the source domain $f_S$ is trained with labeled outputs $Y_S = f_S(\Theta_S)$. The trained regression model $f_S(\cdot)$ is transferred to the target domain for the prediction task, i.e. $f_S(\cdot) \rightarrow f_T(\cdot)$.

\section{Evaluation Method}

Common evaluation metrics used when predicting RUL for the turbofan data are root mean square error (RMSE), the PHM score function, and mean absolute percentage error (MAPE). Other widely used prognostic metrics are mentioned and explained in the survey by Saxena et al. \cite{saxena2008metrics}.

The proposed TL approach was designed to handle various new operating conditions and new deterioration progressions affected by unseen faults. Our approach shall provide a robust prediction on RUL that reflects the health condition of the equipment (ideally) during the deterioration period.
%
%
Therefore, in this study, the evaluation shall follow three criteria: \emph{i}) the evaluation shall include multiple consecutive cycles of each testing trajectory, rather than only focus on the last cycle, so that a more comprehensive coverage on samples with various operating conditions is ensured; \emph{ii}) the evaluation metric shall weight samples closer to EOL with greater importance: industrial systems are often not allowed to fail and therefore prediction near failure samples is of greater interest; \emph{iii}) the evaluation shall not overwhelm errors occur closer to EOL with errors from cycles during which the equipment is in healthy condition.
We use MAPE \cite{de2016mean}, which was applied for evaluation by Rigamonti et al. \cite{rigamonti2016echo}. The \mbox{MAPE} of an engine unit $u$ is computed in the following way (with floating-point representation):
\begin{eqnarray}\label{eqn:mape}
\mbox{MAPE}(u) & = & \frac{1}{l(u)} \ \sum _{t=1}^{l(u)} \ \left| \frac{y(u, t) - \hat{y}(u, t)}{ y(u, t) } \right|
\end{eqnarray}
Here $l(u)$ is the time interval the average is computed over for engine unit $u$, $y(u, t)$ is the true remaining useful life at time $t$ and $\hat{y}(u, t)$ is the estimated RUL at time $t$ of $u$. The $\mbox{MAPE}$ for $N$ engines is computed with: $\frac{1}{N}\sum _{u=1}^N \mbox{MAPE}(u)$.
%
The MAPE evaluates RUL predictions on a given trajectory and penalizes errors more when they occur close to EOL. 
In contrast, the RMSE weights all samples equally, regardless of how much RUL there is left. When computing MAPE and RMSE, we only include the degradation period, i.e. the RUL plateau period is ignored (the constant section in the target value). 


\section{Results}\label{sec:results}


The proposed COSMO feature TL technique was evaluated with several experiments, using scenarios where new faults and/or operating conditions were introduced in the target domain, see Table~\ref{tab:transfer_learning_scenarios_detailed} in the supplementary material. 

As already mentioned, each of the four C-MAPSS subsets is split into predefined training and testing sets. The training sets $X_\alpha$ contain complete run-to-failure trajectories, whereas the testing sets $X_\beta$ are right censored. 
Full run-to-failure trajectories are more interesting to study since they have more samples that are closer to EOL. We performed two types of experiments in this study: $i$) Where both the source and the target domain samples come from the run-to-failure trajectories in $X_\alpha$ (but not from the same data subset); $ii$) Where the source domain data were drawn from run-to-failure trajectories in $X_\alpha$ but the target domain data were drawn from the censored test set $X_\beta$.
We refer to $i$) as mode $\alpha$ and $ii$) as mode $\beta$. All experiments were done using four fold cross-validation and the uncertainty measure was computed accordingly.

The regression method chosen was the RF regressor \cite{breiman2001random, geurts2006extremely} and the python implementation in scikit-learn \cite{scikit-learn}. We did not fine-tune the learning parameters for the different data subsets and the performance of our RF regressor was essentially the same as what is reported by Zhang et al. using RF \cite{zhang2017multiobjective}. The reference group $\varphi$ in this work was a subset of the nominal data samples $H$, The size of $\varphi_S$ and $\varphi_T$ was set to $80$ and all samples were drawn randomly. The number of nearest neighbors ($k$) for computing the COSMO features was set to $8$. Based on the Eigengap heuristic, the maximum number of clusters in the C-MAPSS dataset was set to $6$ and the condition 
eq.~(\ref{eq:knn-condition}) was fulfilled.

The result section is organized as follows: $1$) we first illustrate the performance of applying a traditional approach, based on RF regressor, on various scenarios listed in table~\ref{tab:transfer_learning_scenarios_detailed}; 
$2$) performance comparison between using the sensor data, four variations of COSMO features, and three domain adaptation techniques (SCL, CORAL and TCA); $3$) performance comparison of COSMO features and sensor data on samples when the RUL limit for evaluation is varied.


\begin{figure*}
\centering
\subfigure{%
\includegraphics[width=0.48\textwidth]{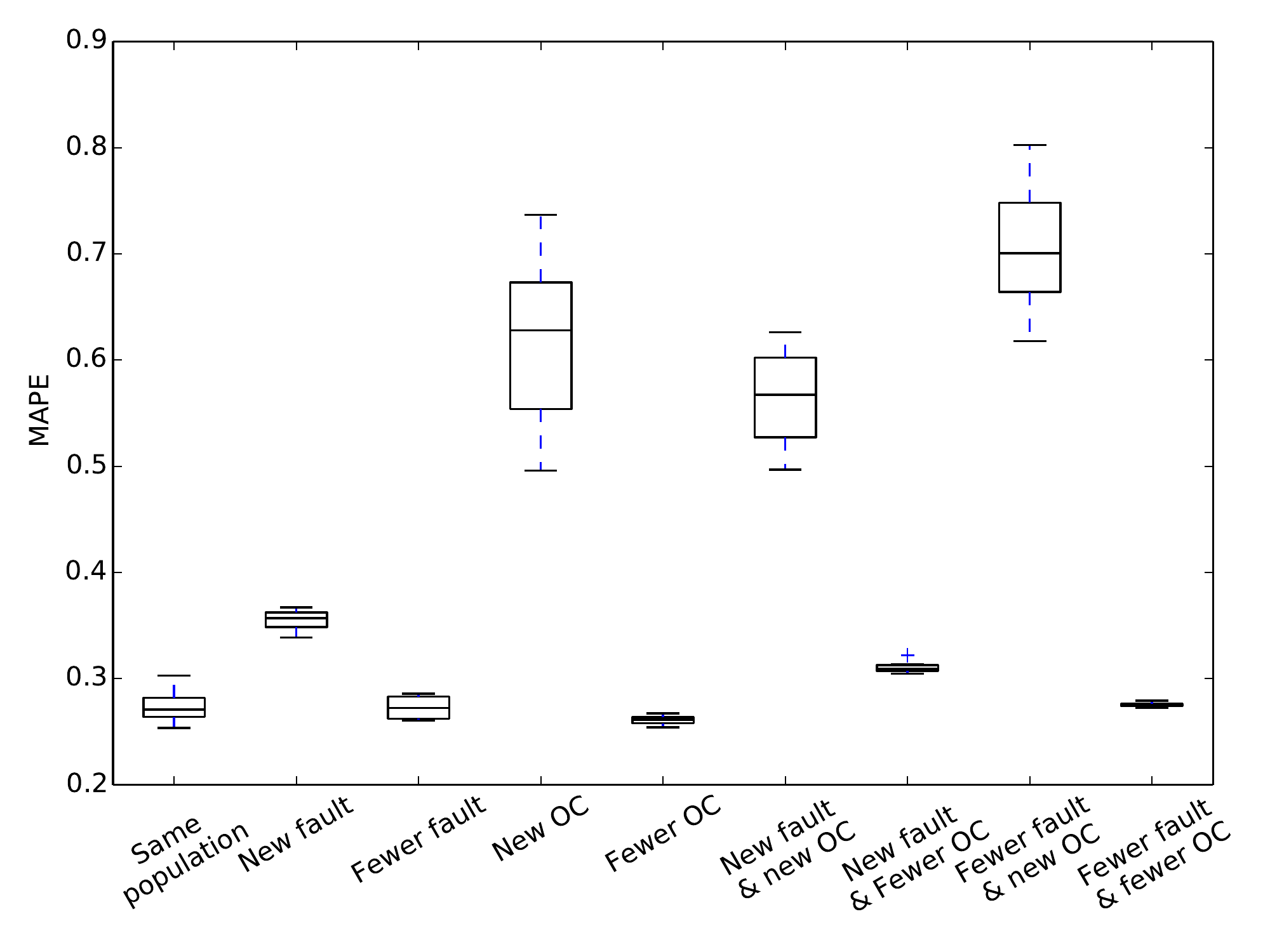}
\label{fig:rf_tf_ra}%
}
\subfigure{%
\includegraphics[width=0.48\textwidth]{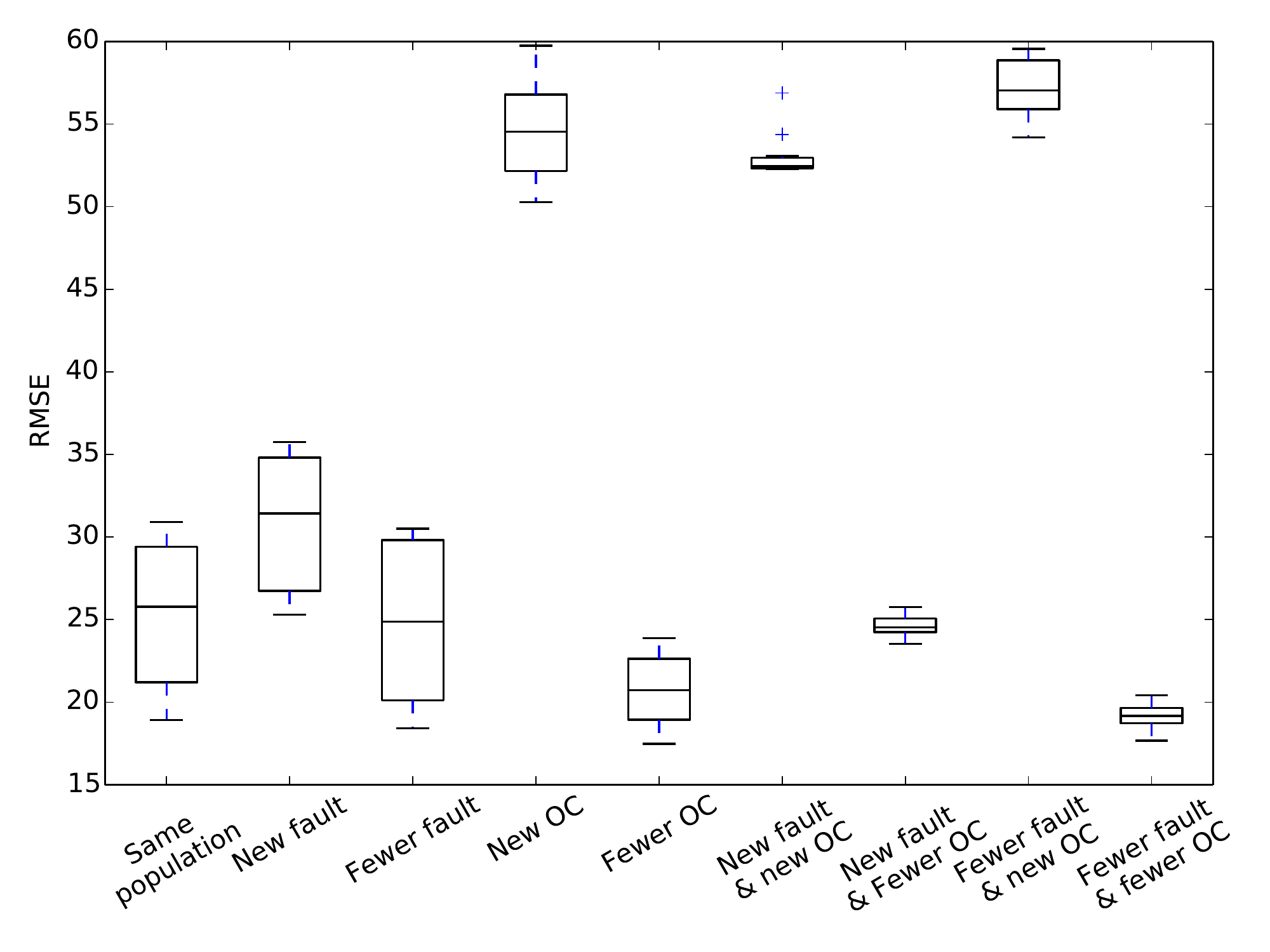}
\label{fig:rf_tf_rmse}%
}
\caption{MAPE (left) and RMSE (right) of using Random Forest Regressor on different scenarios: performances are significantly worse when encountering new operating conditions or new fault are presented in the target domain.}
\label{fig:performance_rf_tl}
\end{figure*}

The results shown in Figure~\ref{fig:performance_rf_tl} illustrate the performance of RF regression on various learning scenarios (mode $\beta$) averaged over ten experiments each, where MAPE was computed based on samples with RUL $<$ 130 cycles, and RMSE was computed based on only the last cycle. The label \emph{Same population} covers scenarios A1 to A4 (i.e. where no TL is needed); \emph{New fault} covers B1 and B2; \emph{Fewer fault} covers E1 and E2; \emph{New OCs} covers C1 and C2; \emph{Fewer OCs} covers F1 and F2; \emph{New fault \& new OCs} covers D; \emph{New fault \& Fewer OCs} covers G1; \emph{Fewer fault \& new OCs} covers G2 and \emph{Fewer fault \& fewer OCs} covers H.
The RF performance without any TL technique is significantly worse in scenarios where new faults or new operating conditions occur in the target domain (cf. the source domain). Therefore, experiments on scenarios ``\emph{same population}", ``\emph{New fault}", ``\emph{New operating conditions}", and ``\emph{New fault and operating conditions}" are the focus of this study and will be evaluated.
%
%
The ``\emph{same population}" experiments were not done to demonstrate TL since they do not require any transfer; they were done to compare the RF performance against the typical applications on this data.

\begin{figure*}
\centering
\subfigure[$\alpha$: scenario A1-A4]{%
\includegraphics[width=0.24\textwidth]{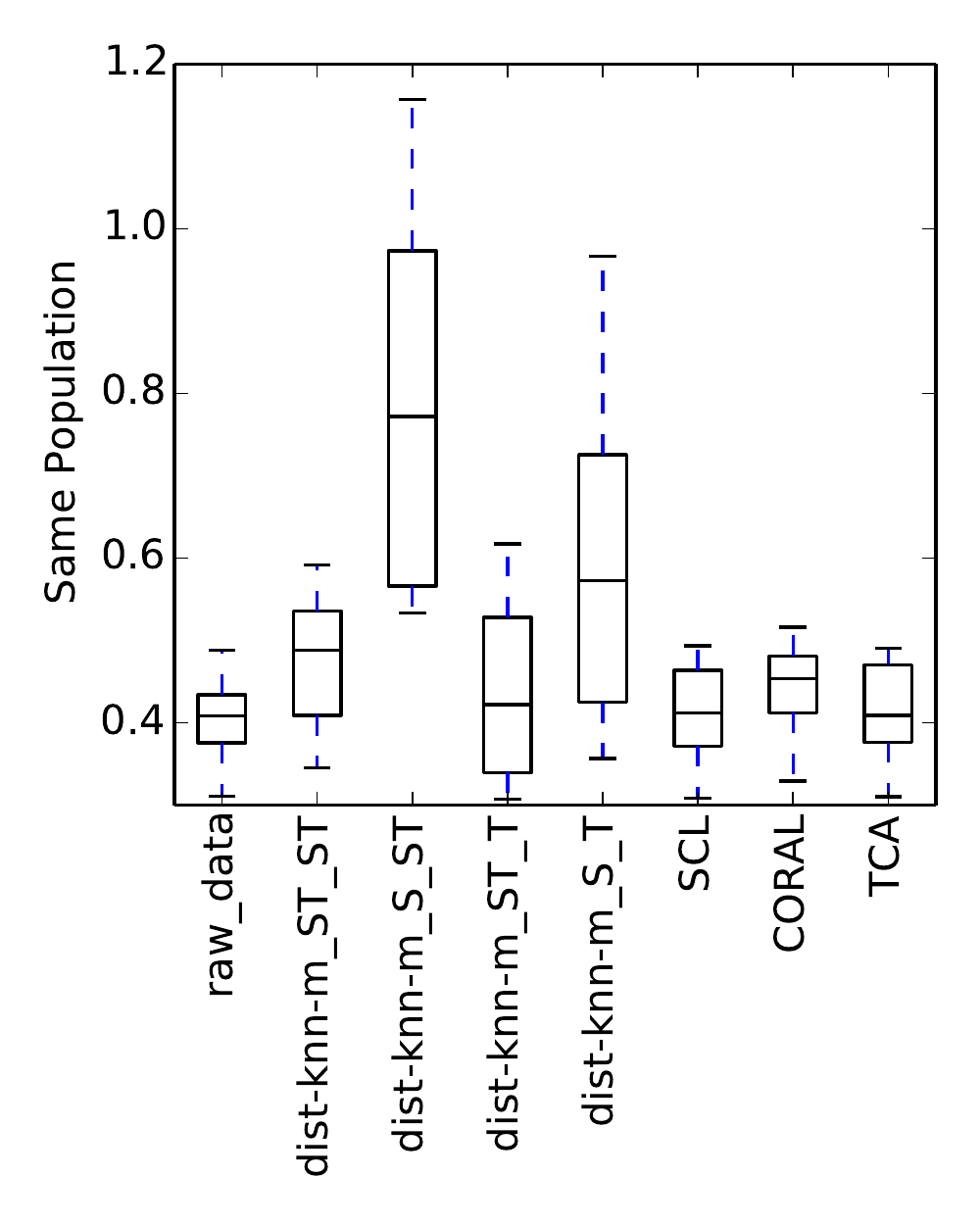}%
\label{fig:box_de_r}%
}
\subfigure[$\alpha$: scenario D]{%
\includegraphics[width=0.24\textwidth]{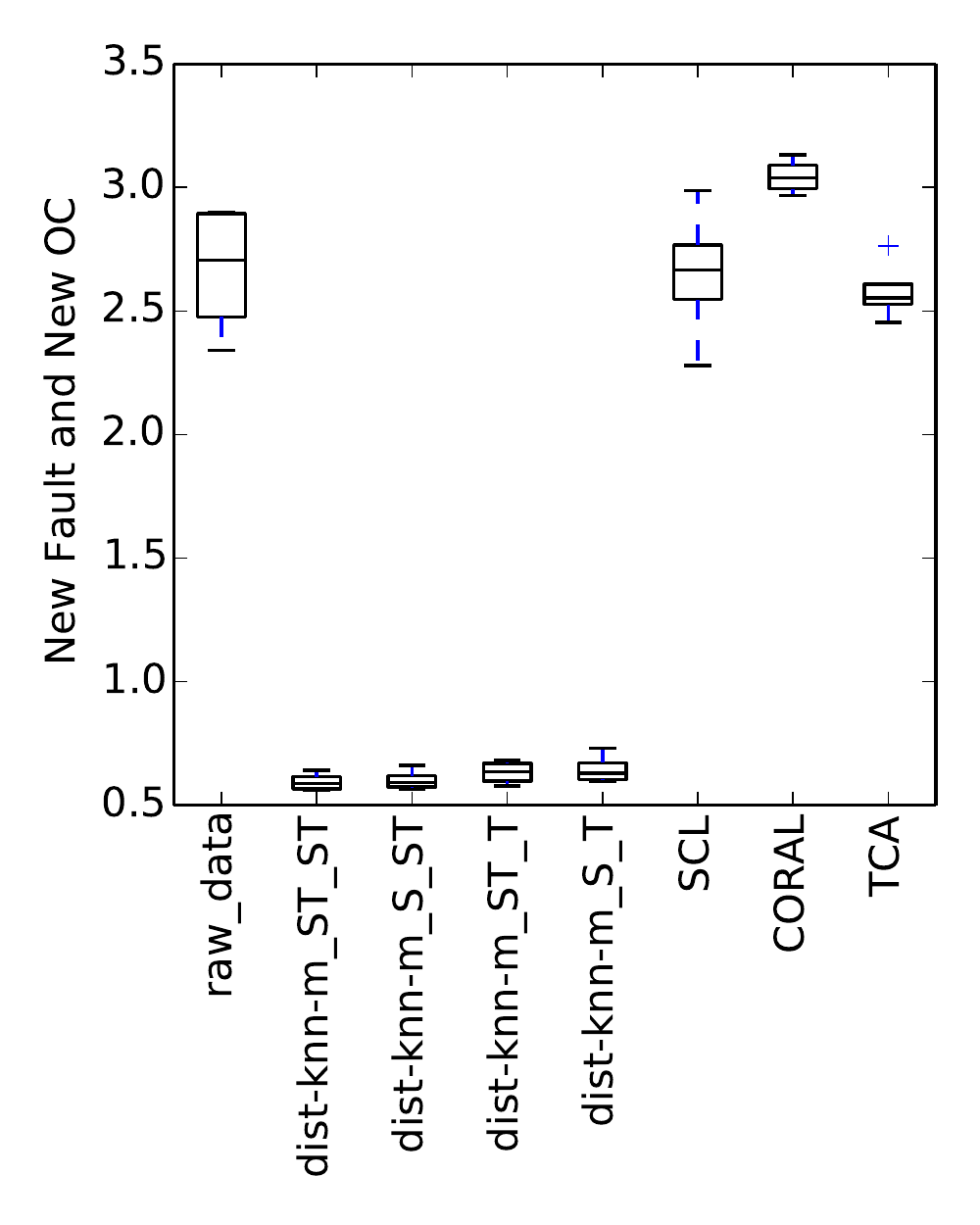}%
\label{fig:box_bo_r}%
}
\subfigure[$\alpha$: scenario B1]{%
\includegraphics[width=0.24\textwidth]{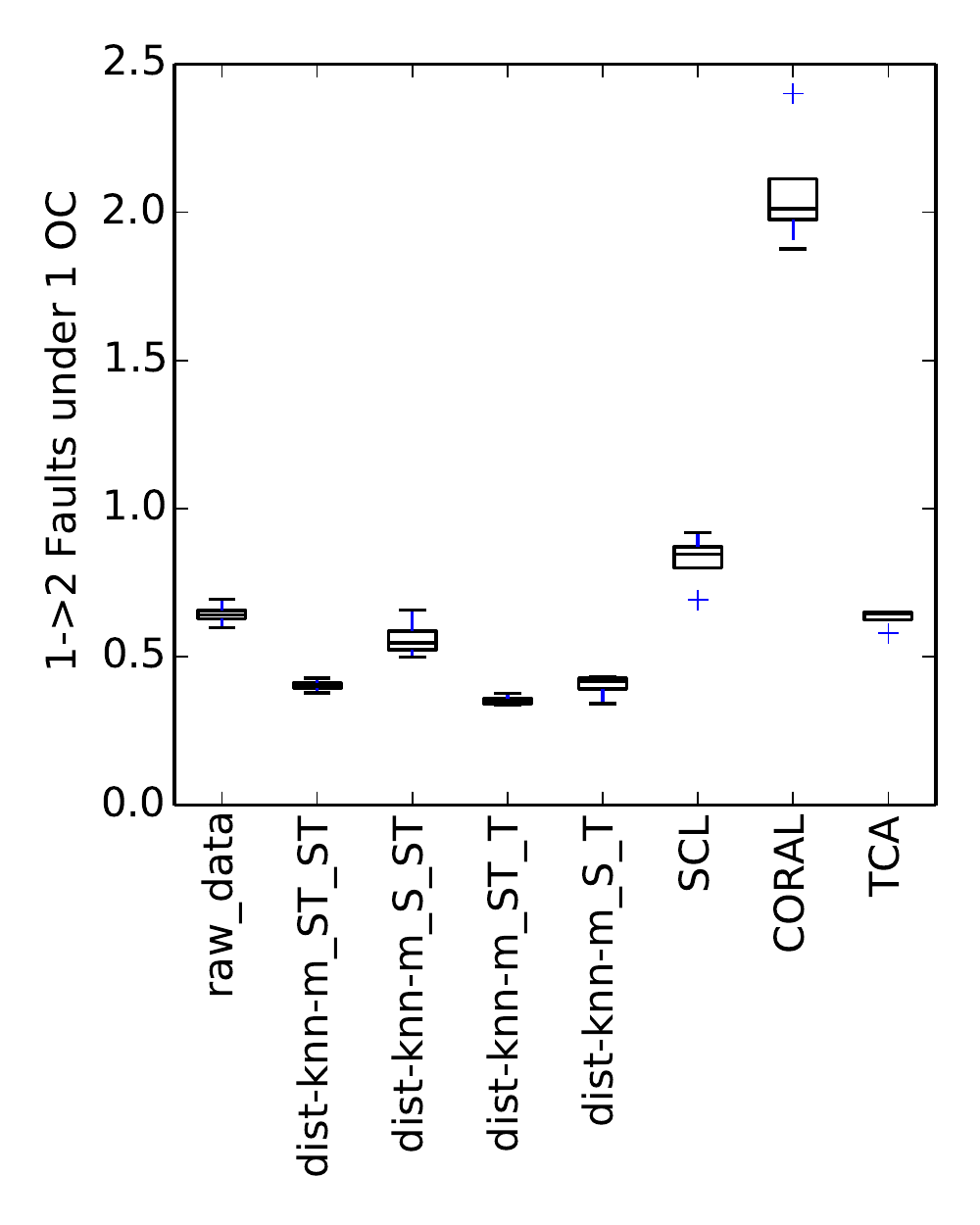}%
\label{fig:box_b1_r}%
}
\subfigure[$\alpha$: scenario C1]{%
\includegraphics[width=0.24\textwidth]{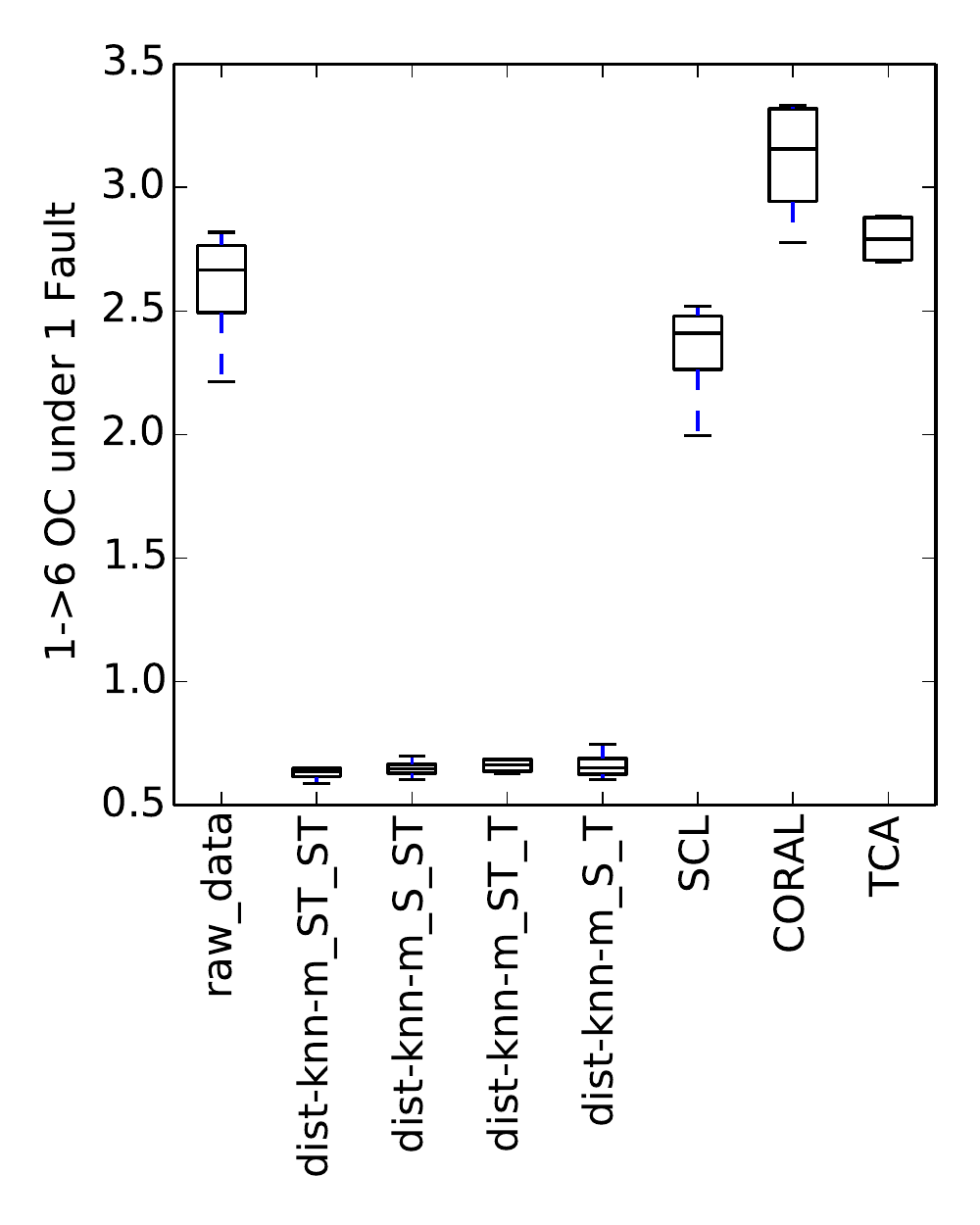}%
\label{fig:box_c1_r}%
}\vspace{-2mm}
\caption{Performance Comparison on scenario (a) ``\emph{same population}"; (b) ``\emph{New fault}"; (c) ``\emph{New operating conditions}", and (d) ``\emph{New fault and operating conditions}" under mode $\alpha$.
}
\label{fig:box_result}
\end{figure*}

The results shown in Figure~\ref{fig:box_result} are comparisons between using (raw) sensor data, COSMO features ($\mbox{m-\it{k}NN}$ distance) and other feature representation transfer (or domain adaptation) techniques (i.e. SCL, CORAL and TSA) for predicting RUL under mode $\alpha$. The performance is evaluated on samples with RUL less than 130 cycles since only the deterioration period is of interest. Figure~\ref{fig:box_de_r} shows that using COSMO features, except $(S, ST)$, yields results that are very similar to using the raw sensor data in predicting RUL for scenarios A1 to A4, i.e. ``\emph{same population}''. However, when dealing with scenarios where the source and the target domain are different (D, B1, and C1) all variations of COSMO features yield results that are significantly better than all other methods, as is shown in Figure~\ref{fig:box_bo_r}, \ref{fig:box_b1_r}, and \ref{fig:box_c1_r}. Results in other scenarios listed in Table~\ref{tab:transfer_learning_scenarios_detailed}, including mode $\beta$, are provided in the supplementary material. Results for mode $\beta$ have lower statistical significance compared to mode $\alpha$. This is because the test data are right censored.

In summary, COSMO features outperform other methods when new fault and/or operating conditions are present in the target domain, i.e. scenarios D, B1, and C1. Using COSMO features with $mode(ST, ST)$ achieves overall lowest MAPE compared to other methods.
An illustration of COSMO features is available in the supplementary material. With the proposed approach, the difference in features generated from the source and the target data is less than the raw sensor readings, while it still preserves some characteristics of the underlying degradation process.

\begin{figure*}
\centering
\subfigure[$\alpha$: scenario A1-A4]{%
\includegraphics[width=0.24\textwidth]{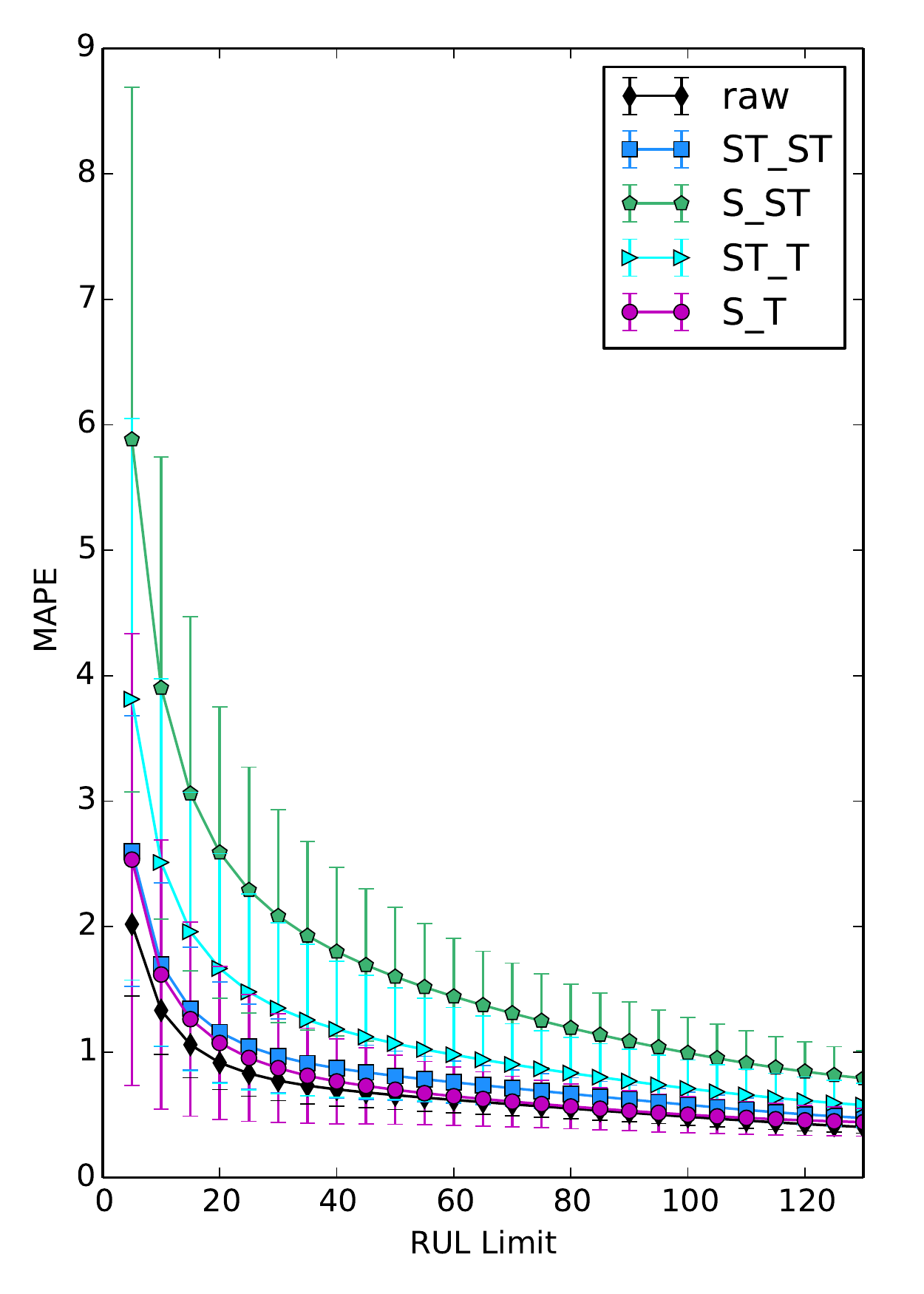}%
\label{fig:err_de_r}%
}
\subfigure[$\alpha$: scenario D]{%
\includegraphics[width=0.24\textwidth]{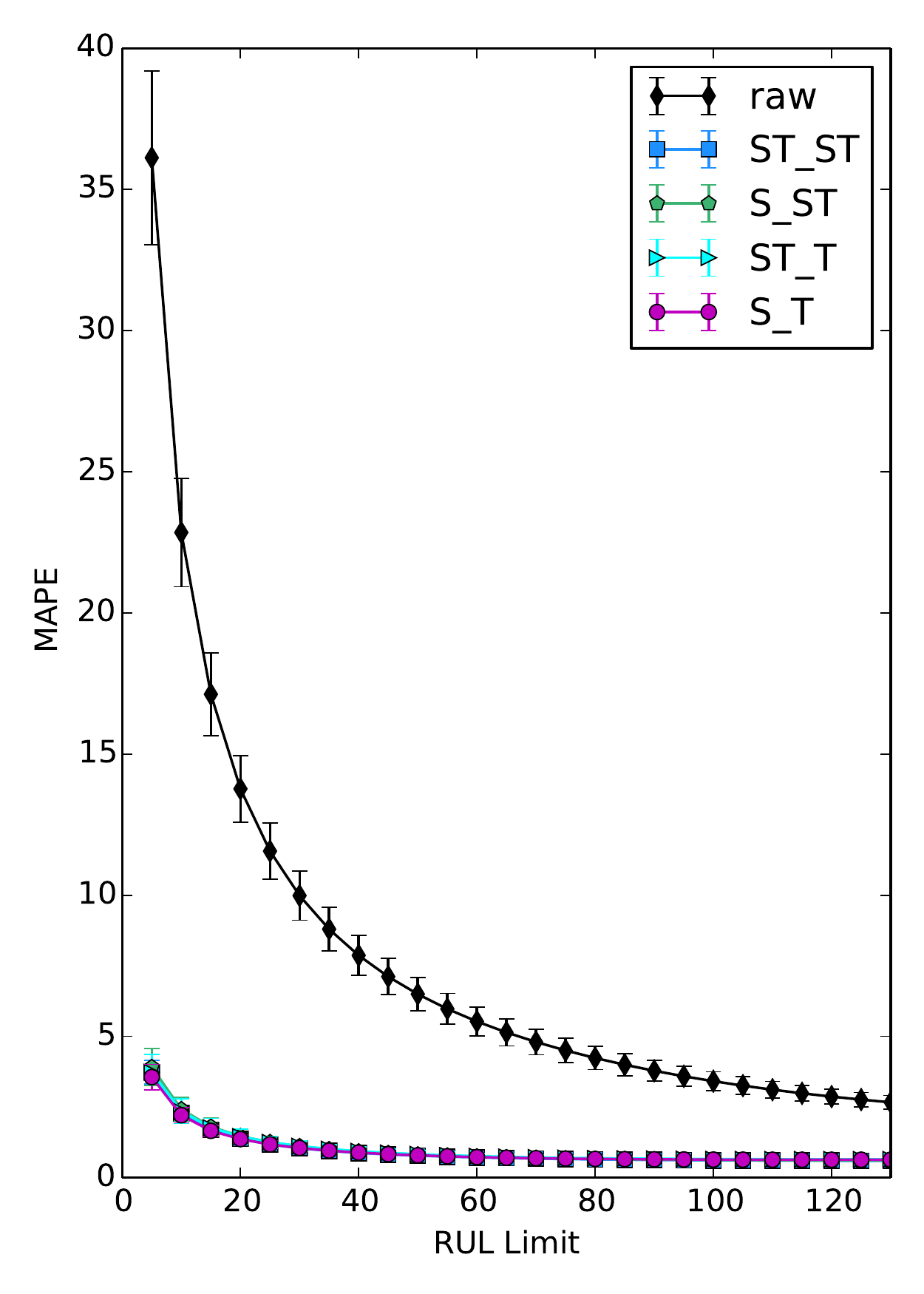}%
\label{fig:err_bo_r}%
}
\subfigure[$\alpha$: scenario B1]{%
\includegraphics[width=0.24\textwidth]{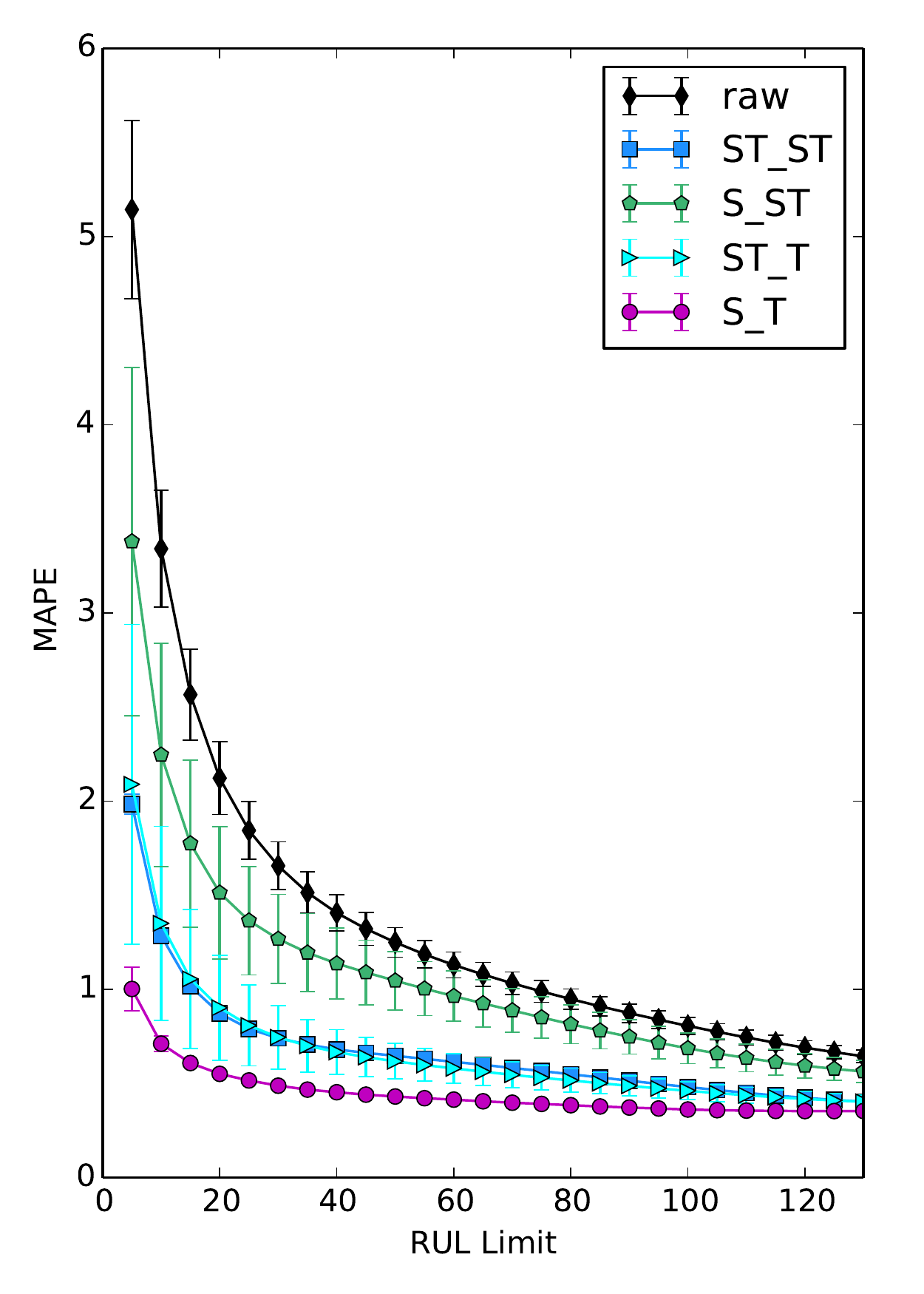}%
\label{fig:err_b1_r}%
}
\subfigure[$\alpha$: scenario C1]{%
\includegraphics[width=0.24\textwidth]{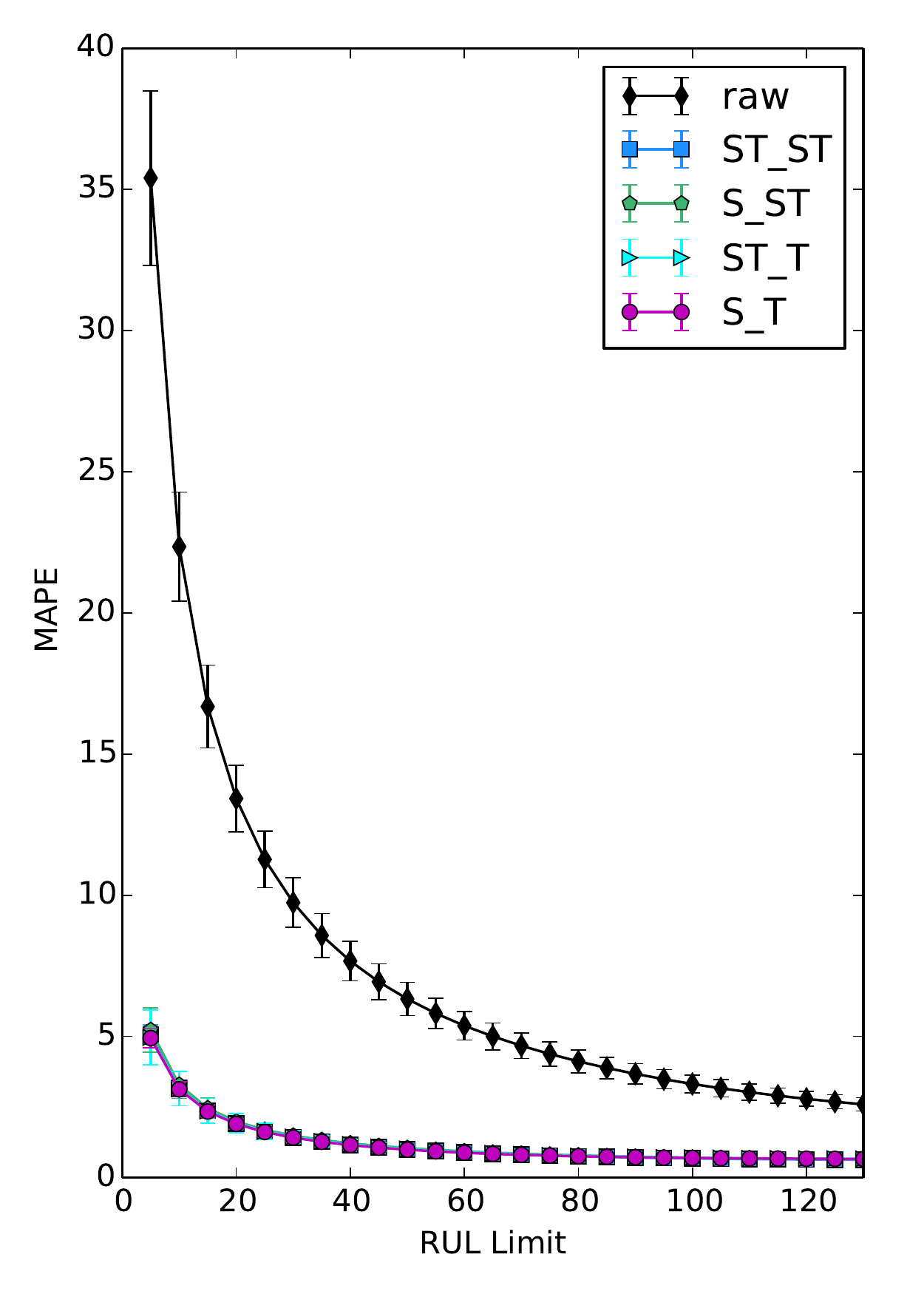}
\label{fig:err_c1_r}%
}
\caption{Performance comparison: MAPE w.r.t. samples with varied RUL limits on scenario (a) ``\emph{same population}"; (b) ``\emph{New fault}"; (c) ``\emph{New operating conditions}", and (d) ``\emph{New faults and operating condition}" under mode $\alpha.$}
\label{fig:err_result} 
\end{figure*}

\begin{figure*}
\centering
\subfigure[$\alpha$: scenario A1-A4]{%
\includegraphics[width=0.24\textwidth]{figure/result/bar_ph/__A_DE__rul_var_case_MAPE_A_z.pdf}%
\label{fig:err_de_a}%
}
\subfigure[$\beta$: scenario A1-A4]{%
\includegraphics[width=0.24\textwidth]{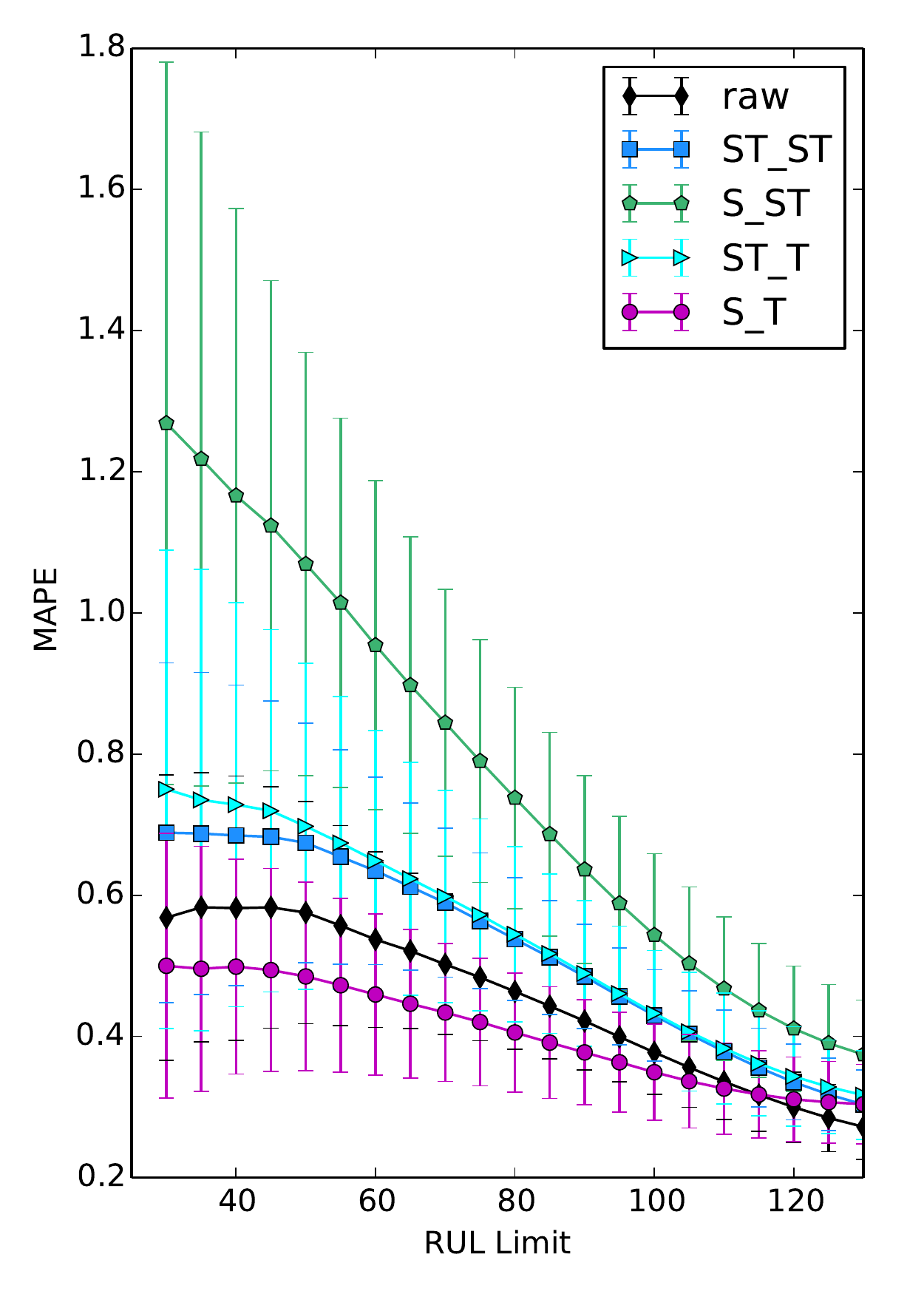}%
\label{fig:err_de_b}%
}
\subfigure[$\alpha$: scenario D]{%
\includegraphics[width=0.24\textwidth]{figure/result/bar_ph/__A_BO__rul_var_case_MAPE_D_z.pdf}%
\label{fig:err_bo_a}%
}
\subfigure[$\beta$: scenario D]{%
\includegraphics[width=0.24\textwidth]{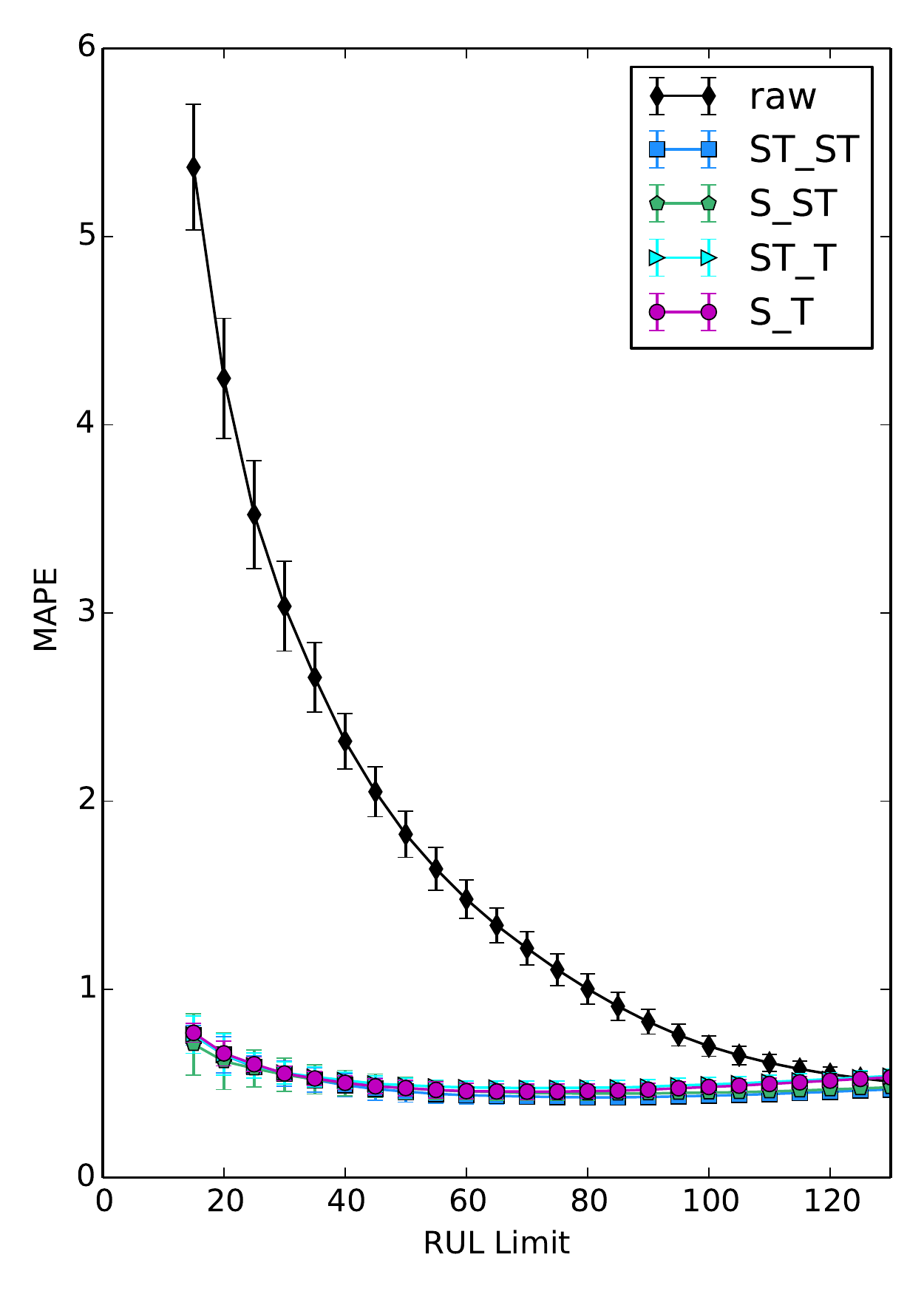}
\label{fig:err_bo_b}%
}
\subfigure[$\alpha$: scenario B1]{%
\includegraphics[width=0.24\textwidth]{figure/result/bar_ph/__A_NF__rul_var_case_MAPE_B1_z.pdf}%
\label{fig:err_nf_b1a}%
}
\subfigure[$\alpha$: scenario B2]{%
\includegraphics[width=0.24\textwidth]{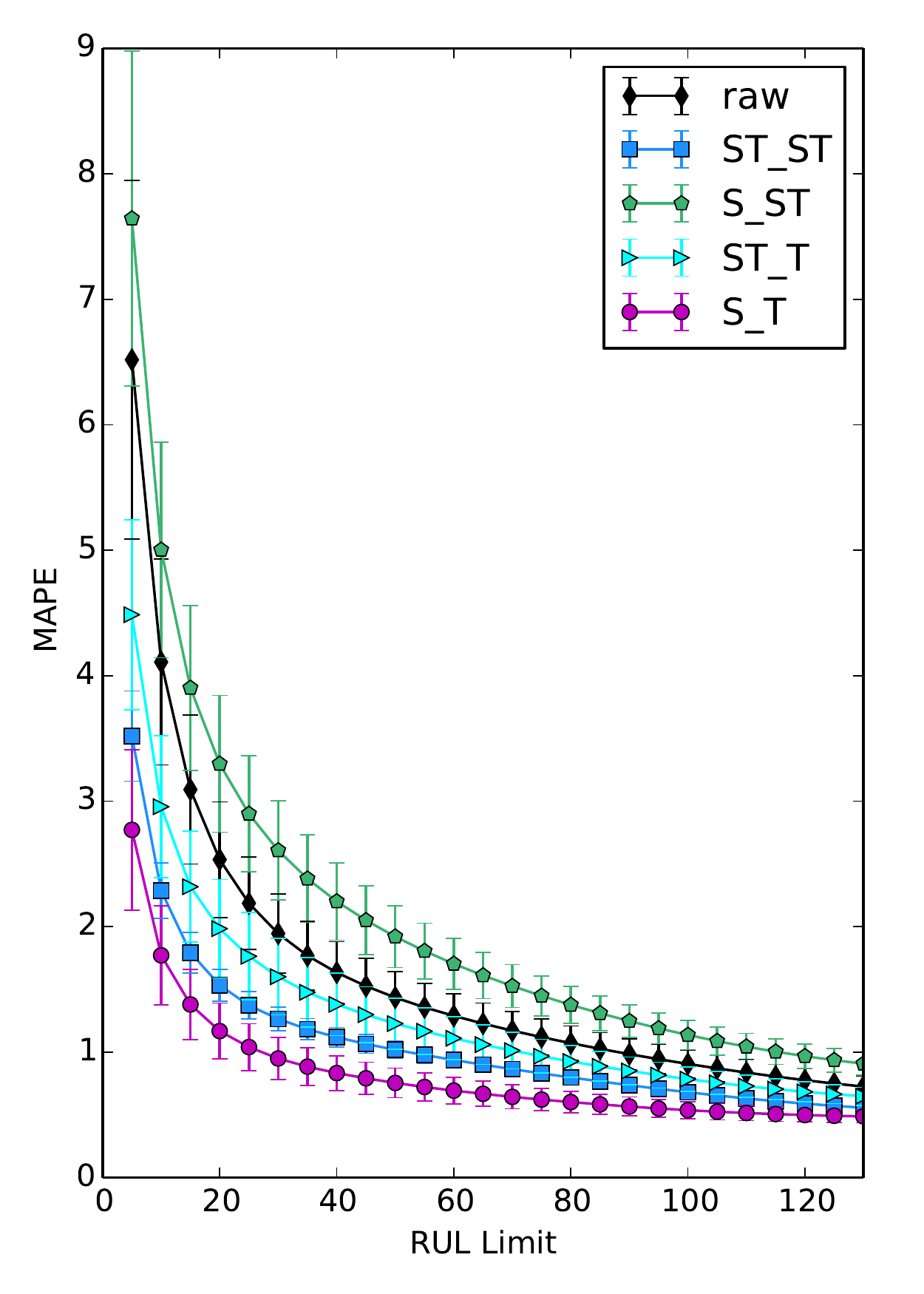}%
\label{fig:err_nf_b2a}%
}
\subfigure[$\beta$: scenario B1]{%
\includegraphics[width=0.24\textwidth]{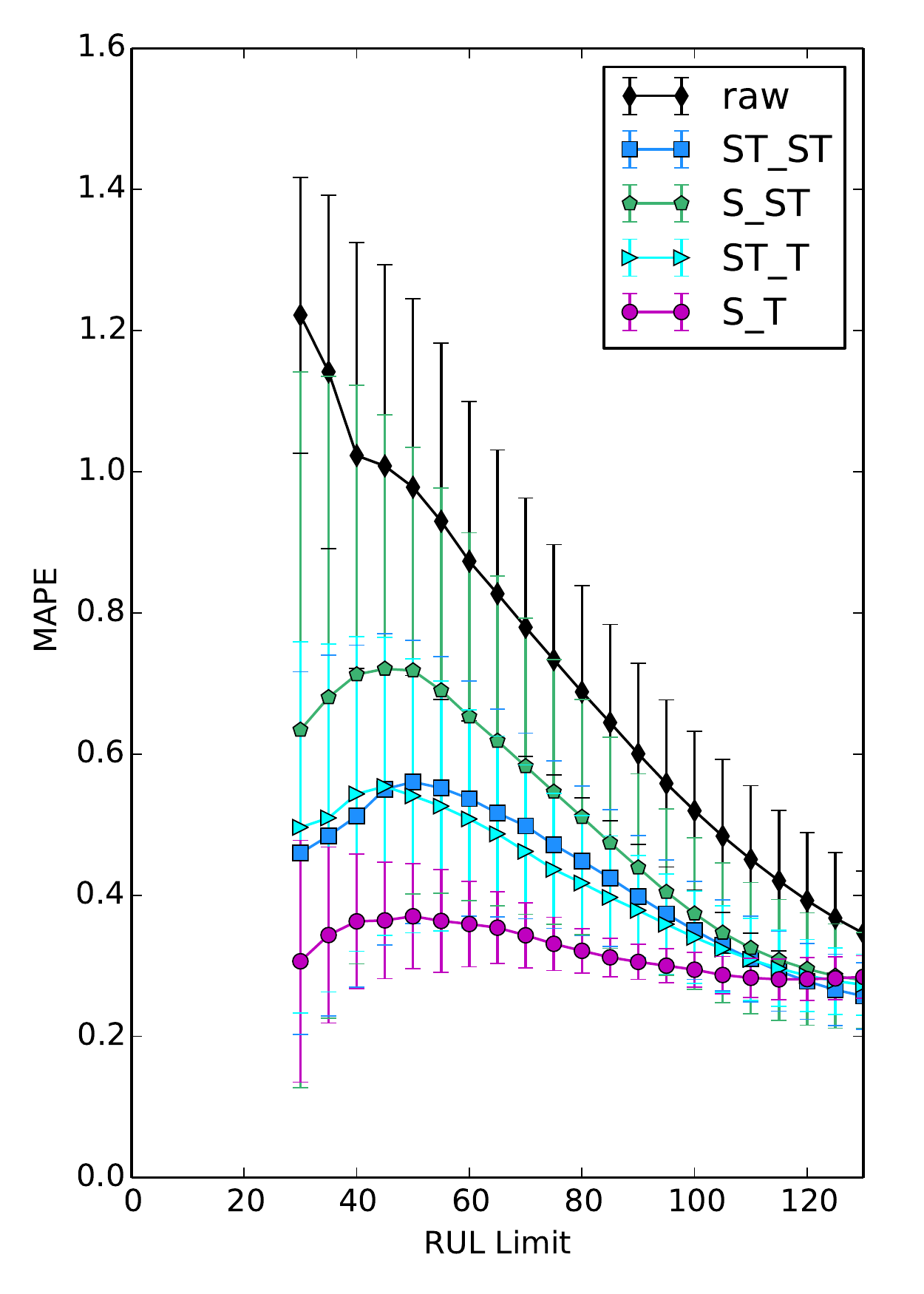}%
\label{fig:err_nf_b1b}%
}
\subfigure[$\beta$: scenario B2]{%
\includegraphics[width=0.24\textwidth]{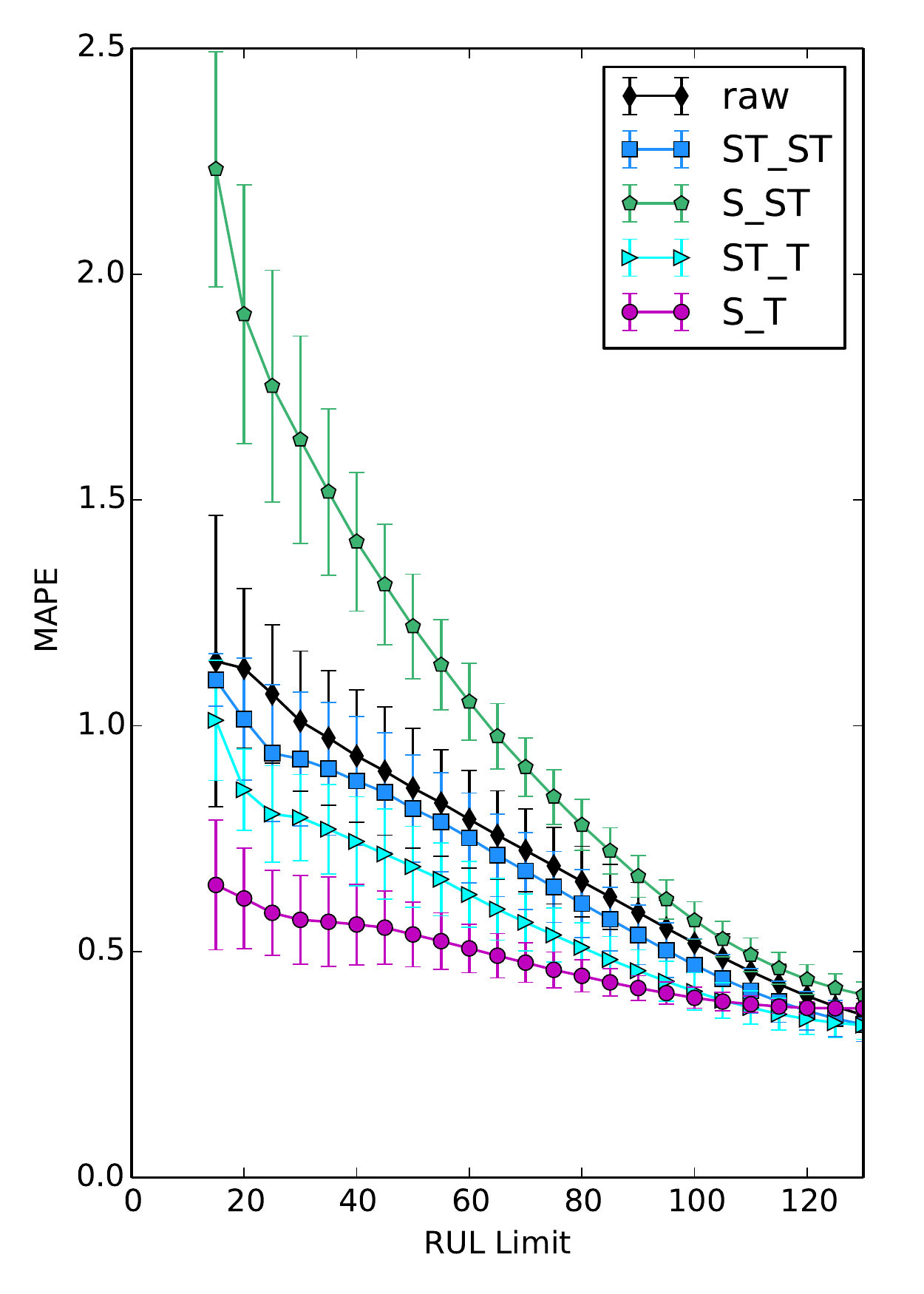}
\label{fig:err_nf_b2b}%
}
\subfigure[$\alpha$: scenario C1]{%
\includegraphics[width=0.24\textwidth]{figure/result/bar_ph/__A_NO__rul_var_case_MAPE_C1_z.pdf}%
\label{fig:err_no_c1a}%
}
\subfigure[$\alpha$: scenario C2]{%
\includegraphics[width=0.24\textwidth]{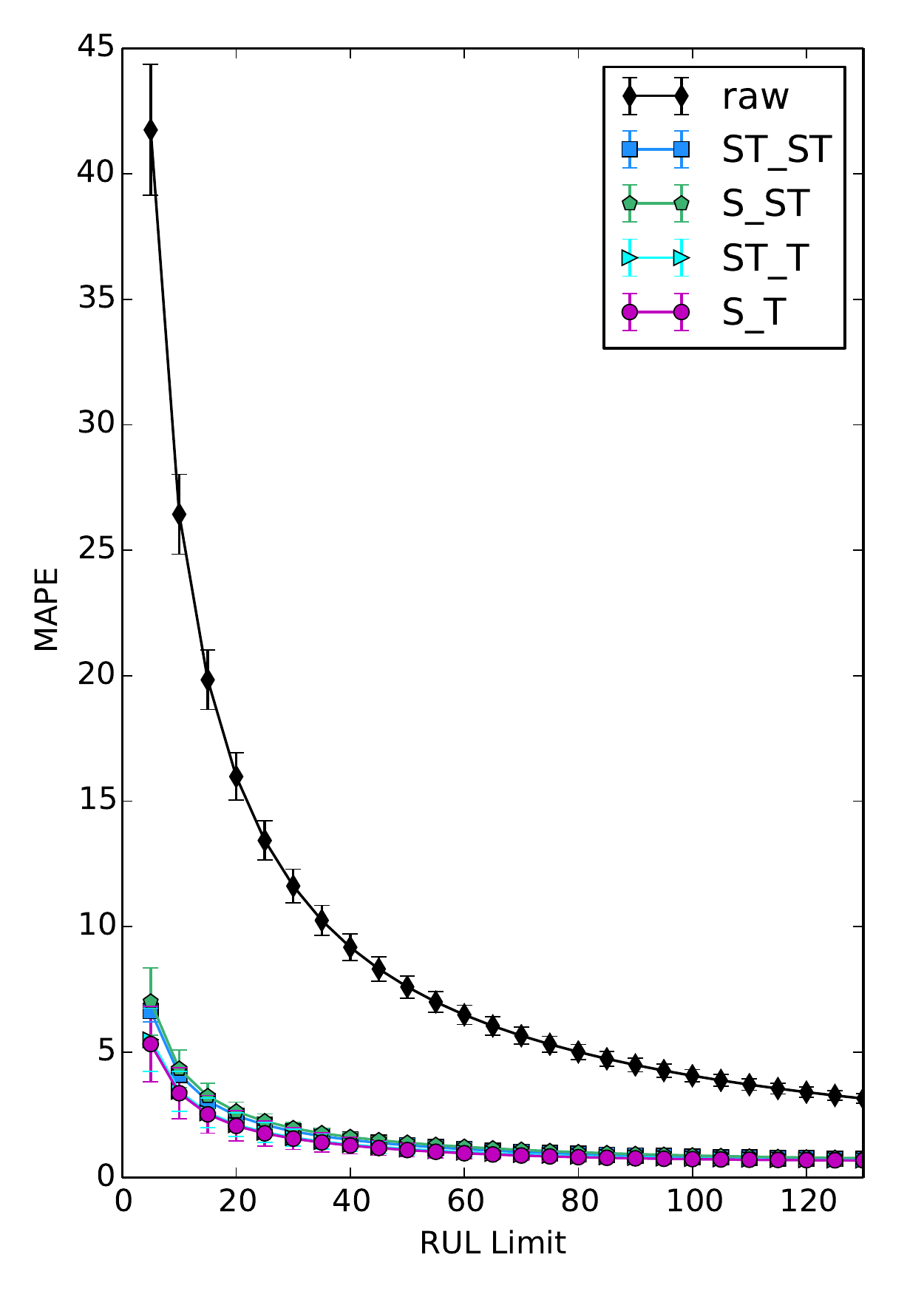}%
\label{fig:err_no_c2a}%
}
\subfigure[$\beta$: scenario C1]{%
\includegraphics[width=0.24\textwidth]{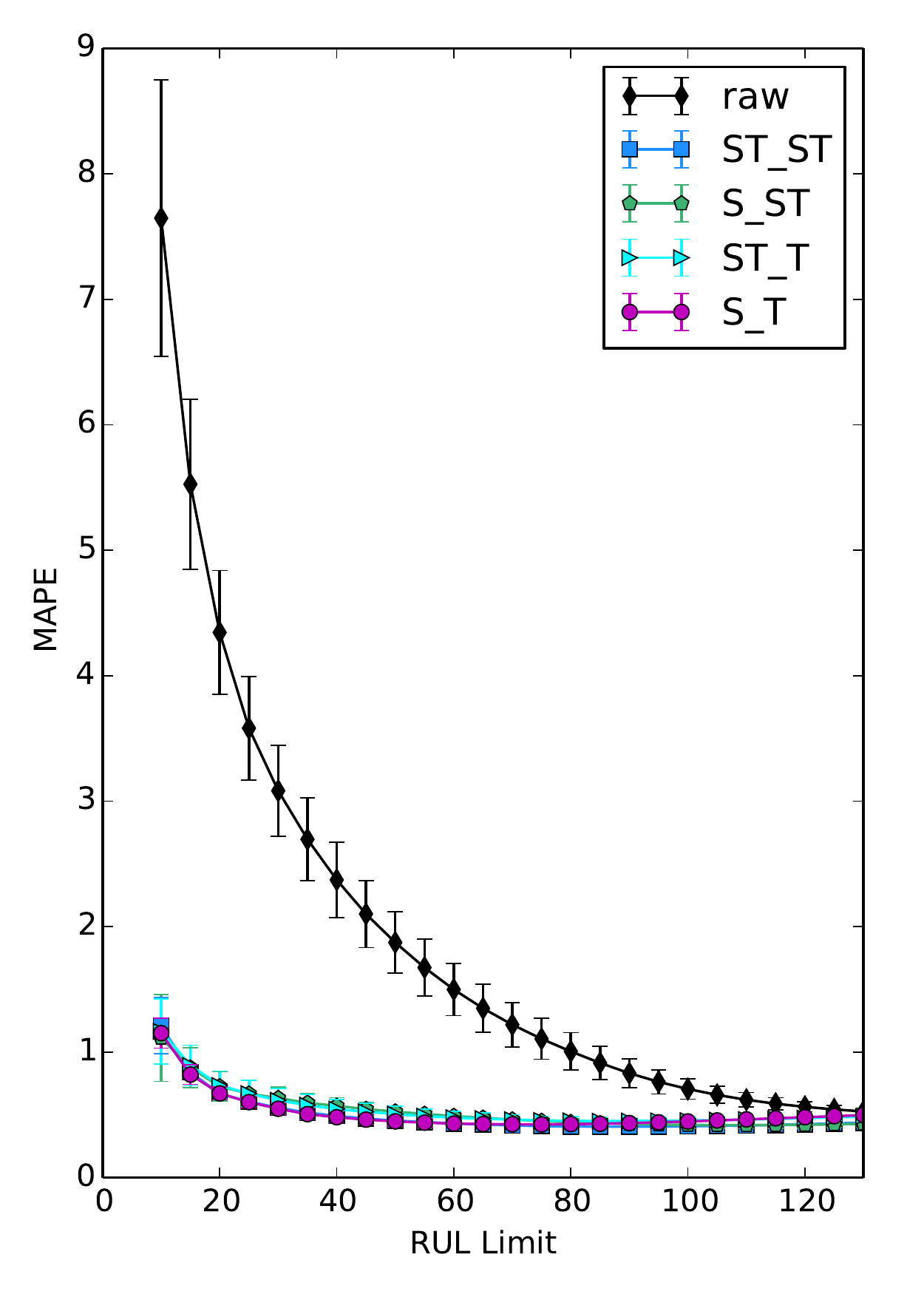}%
\label{fig:err_no_c1b}%
}
\subfigure[$\beta$: scenario C2]{%
\includegraphics[width=0.24\textwidth]{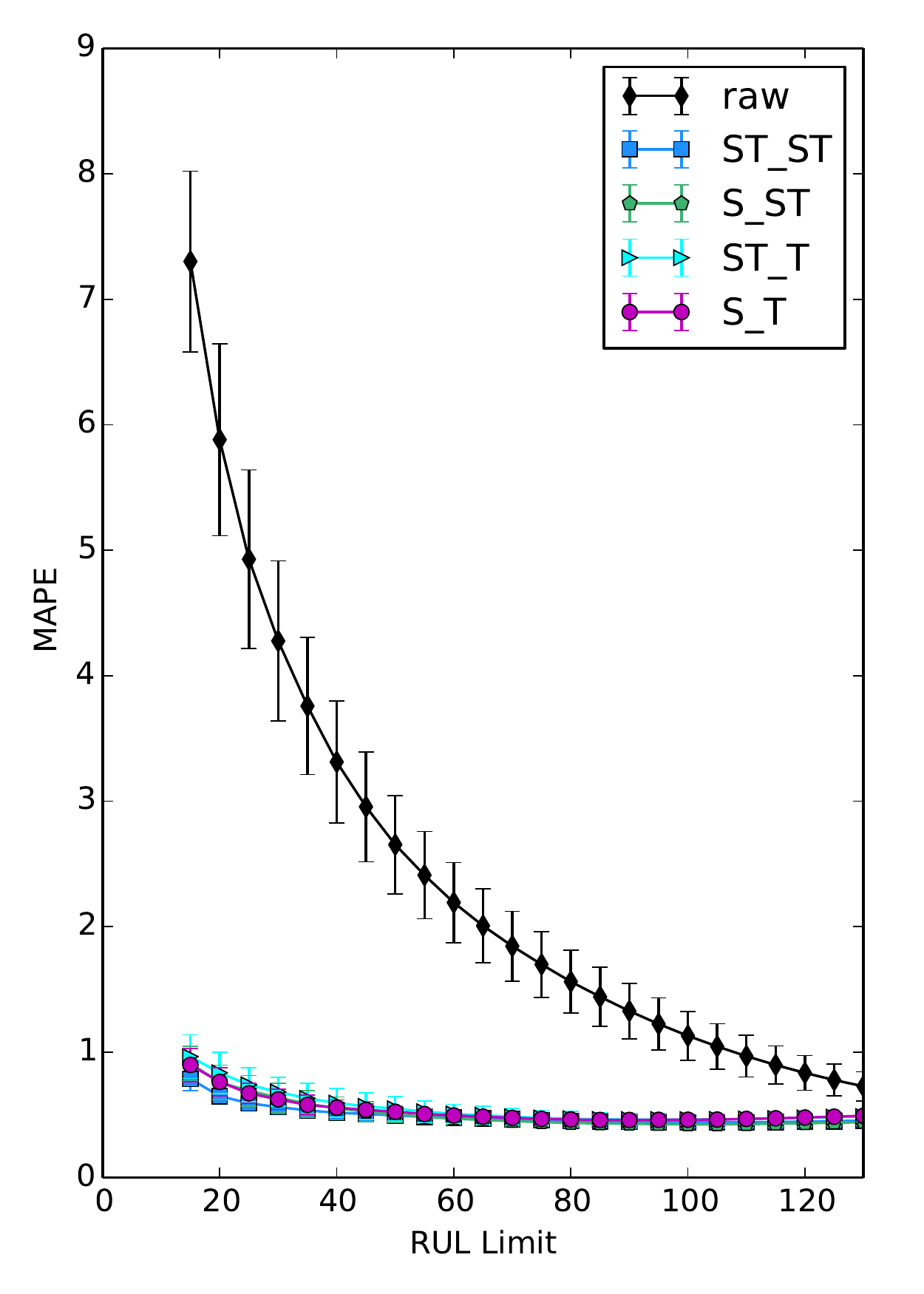}
\label{fig:err_no_c2b}%
}
\caption{Performance comparison: MAPE w.r.t. samples with varied RUL limits on scenario A to D under mode $\alpha$ and $\beta$.}
\label{fig:err_all} 
\end{figure*}

Figure~\ref{fig:err_result} illustrates how MAPE changes over samples with RUL from $1$ to $130$. 
Four variations of COSMO feature (m-$k$NN distance) are compared with sensor data in predicting RUL. The MAPE converges when more samples with larger RUL are included for evaluation. Figure~\ref{fig:err_de_r} shows that COSMO features, except $(S, ST)$, have no significant difference compared to the traditional approach on scenario A1 to A4 (``\emph{same population}"). For scenarios where new fault and/or operating conditions are present in the target domain, all four variations of COSMO features outperform the traditional approach, illustrated in Figure~\ref{fig:err_bo_r}, \ref{fig:err_b1_r} and \ref{fig:err_c1_r}. COSMO features with $mode(S, T)$ achieve the best performance, especially on scenario B1 (``\emph{New fault}"). Results on some other scenarios are illustrated in Figure~\ref{fig:err_all}.




\section{Conclusion}
\label{sec:conclusions}

This work addressed the need of performing TL in adapting prognostic methods to handle future data samples that may come from unseen distribution, undertaking new faults and deterioration progressions, i.e. a target domain that is different from the source domain. The proposed feature-representation based TL approach utilizes a transferable feature that captures the distance to the peer for each individual sample. 
If the selected peer group is representative of nominal conditions across all operating profiles, the proposed COSMO feature (m-{\it k}NN) will generalize samples from different domains to a common latent feature space where discrepancy of marginal distributions between domains is reduced and deviating (or near EOL) samples are projected onto the edge of the majority.
The experimental result shows that with a comprehensive peer group including nominal samples from both domains, COSMO feature with $\mbox{mode(ST, ST)}$ achieves the best performance in two out of four experiment scenarios. The error in dealing with new operating conditions (scenario C1 and D) is significantly lower than TCA, CORAL and SCL, approximately four times lower compared to the traditional approach.

\renewcommand*{\bibfont}{\footnotesize}
\bibliographystyle{abbrv}
\bibliography{r2s}

\section{Supplementary Material}

\begin{table*}
\vspace{1cm}
\begin{center}
\begin{tabular}{ccccc}
\toprule
\multirow{2}{*}{Approach} & Dataset 1  & Dataset 2  & Dataset 3 & Dataset 4 \\ 
& 1 OC 1 Fault  & 6 OCs 1 Fault  & 1 OC 2 Faults  & 6 OCs 2 Faults \\  \midrule
LSTM-DANN \cite{da2019remaining} & 13.64 & {\bf 17.76} & 12.49 & {\bf 21.30} \\ 
GA-LSTM (Semi-Deep) \cite{ellefsen2019remaining} & {\bf 12.56} & 22.73 & {\bf 12.10} & 22.66\\ 
CNN-FFNN \cite{li2018remaining} & 12.61 & 22.36 & 12.64 & 23.31\\
MODBNE \cite{zhang2017multiobjective}  & 15.04 & 25.05 & 12.51 & 28.66\\ 
Deep-LSTM \cite{zheng2017long} & 16.14 & 24.49 & 16.18 & 28.17\\
Random Forest \cite{zhang2017multiobjective} & 20.23 & 30.01 & 22.34 & 29.62\\  \midrule
Random Forest & 19.65 $\pm$ 0.80 & 29.43 $\pm$ 0.24 & 22.40 $\pm$ 0.52 & 29.95 $\pm$ 0.43 \\
\bottomrule
\end{tabular}
\vspace{-2mm}
\end{center}
\caption{Performance comparison (RMSE) of studies from literature on the CMAPSS dataset}
\label{tab:SOA_RMSE_tab}
\end{table*}

\begin{table*}
\begin{center} 
\begin{tabular}{c*{7}{c}r}
\toprule
Label   & Source & Target &        OC            & Fault      & Domain                              & Task                               & Scenario                     \\ \midrule
A1      & FD001  & FD001  &      1 $\to$ 1         &  1 $\to$ 1   & \multirow{4}{*}{$D_S = D_T$}        & \multirow{4}{*}{$T_S = T_T$}       & \multirow{4}{*}{Same Population} \\
A2      & FD002  & FD002  &      6 $\to$ 6         &  1 $\to$ 1   &                                     &                                    &                               \\
A3      & FD003  & FD003  &      1 $\to$ 1         &  2 $\to$ 2   &                                     &                                    &                               \\
A4      & FD004  & FD004  &      6 $\to$ 6         &  2 $\to$ 2   &                                     &                                    &                               \\ \midrule
B1      & FD001  & FD003  &      1 $\to$ 1         &  1 $\to$ 2   & \multirow{2}{*}{$D_S = D_T$}        & \multirow{2}{*}{$T_S \neq T_T$} & New Fault under 1 OC  \\
B2      & FD002  & FD004  &      6 $\to$ 6         &  1 $\to$ 2   &                                     &                                    & New Fault under 6 OCs  \\ \midrule
C1      & FD001  & FD002  &      1 $\to$ 6         &  1 $\to$ 1   & \multirow{2}{*}{$D_S \neq D_T$}  & \multirow{2}{*}{$T_S = T_T$}       & New OCs under 1 Fault \\
C2      & FD003  & FD004  &      1 $\to$ 6         &  2 $\to$ 2   &                                     &                                    & New OCs under 2 Faults \\ \midrule
D       & FD001  & FD004  &      1 $\to$ 6         &  1 $\to$ 2   & $D_S \neq D_T$                   & $T_S \neq T_T$                  & New Fault and New OCs  \\  \midrule
E1      & FD003  & FD001  &      1 $\to$ 1         &  2 $\to$ 1   & \multirow{2}{*}{$D_S = D_T$}        & \multirow{2}{*}{$T_S \neq T_T$} & Fewer Fault under 1 OC \\
E2      & FD004  & FD002  &      6 $\to$ 6         &  2 $\to$ 1   &                                     &                                    & Fewer Fault under 6 OCs \\   \midrule
F1      & FD002  & FD001  &      6 $\to$ 1         &  1 $\to$ 1   & \multirow{2}{*}{$D_S \neq D_T$}  & \multirow{2}{*}{$T_S = T_T$}       & Fewer OC under 1 Fault \\
F2      & FD004  & FD003  &      6 $\to$ 1         &  2 $\to$ 2   &                                     &                                    & Fewer OC under 2 Faults \\   \midrule
G1      & FD002  & FD003  &      6 $\to$ 1         &  1 $\to$ 2   & $D_S \neq D_T$                   & $T_S \neq T_T$                  & New Fault and Fewer OC\\
G2      & FD002  & FD001  &      1 $\to$ 6         &  2 $\to$ 1   & $D_S \neq D_T$                   & $T_S \neq T_T$                  & Fewer Fault and New OCs \\   \midrule
H       & FD004  & FD001  &      6 $\to$ 1         &  2 $\to$ 1   & $D_S \neq D_T$                   & $T_S \neq T_T$                  & Fewer Fault and Fewer OC \\
\bottomrule
\end{tabular}\label{tab:experiments_all_scenarios}
\caption{Experiment settings: $a$) B1, B2, C1, C2, and D correspond to learning scenarios with new fault and/or new operating conditions (OCs) present in the target domain; $b$) A1, A2, A3, and A4 correspond to traditional learning scenarios where the training and the testing data coming from the same population; $c$) E1, E2, F1, F2, and H correspond to scenarios where fault and operating condition in the target domain are a subset of the source domain; $d$) G1 and G2 correspond to scenarios that the source and the target domain are different.}
\label{tab:transfer_learning_scenarios_detailed} 
\end{center}
\end{table*}



Table~\ref{tab:illustration_cosmo_feature} illustrated sensor data (the $7$-th feature) and the proposed COSMO feature (m-{\it k}NN) of unit 49 from $X^1$ and unit 20 from $X^2$. As is shown in figures on the first two columns, the difference between the source data $x_S$ and $x_T$ is larger than the difference between COSMO feature $\theta_S$ and $\theta_T$. In the COSMO feature space, the difference between the source and the target domain, due to novel operating conditions, is reduced compared to the sensor data. Consequently, illustrated in figures shown on the third column, the RUL prediction $\hat{y}_T$ with COSMO feature as input to the RF regressor is a more accurate, compared to sensor data, under scenario C1 where $D_S \neq D_T$. Moreover, as is shown in figures on the last column, RUL prediction $\hat{y}_T$ based on COSMO feature behave similarly compared to the one based on the sensor data, under scenario A1 where $D_S = D_T$.

\newcommand{\tableSubplotSizeB}{0.155}
\begin{table*}
\begin{center}
\begin{tabular}{l*{5}{c}r}
\toprule
&  & \multicolumn{2}{c}{Scenario C1} & \multicolumn{2}{c}{Scenario A1} \\
& Source Domain $D_S$ & \multicolumn{2}{c}{Target Domain $D_T \neq D_S$} & \multicolumn{2}{c}{Target Domain $D_T = D_S$} \\ 
& Unit 49 from $X^{1}$ & Unit 49 from $X^{2}$ & RUL Prediction & Unit 20 from $X^{1}$ & RUL Prediction \\ \midrule
& $x_S \subset X^{1}$ & $x_T \subset X^{2}$ & $\hat{y}_T=f_S(x_T)$ & $x_T \subset X^{1}$ & $\hat{y}_T=f_S(x_T)$ \\ 
\shortstack[l]{{\scriptsize Approach} \\ {\scriptsize with Sensor} \\ {\scriptsize data $x$}} &
\begin{minipage}{\tableSubplotSizeB\textwidth}
      \includegraphics[width=\textwidth]{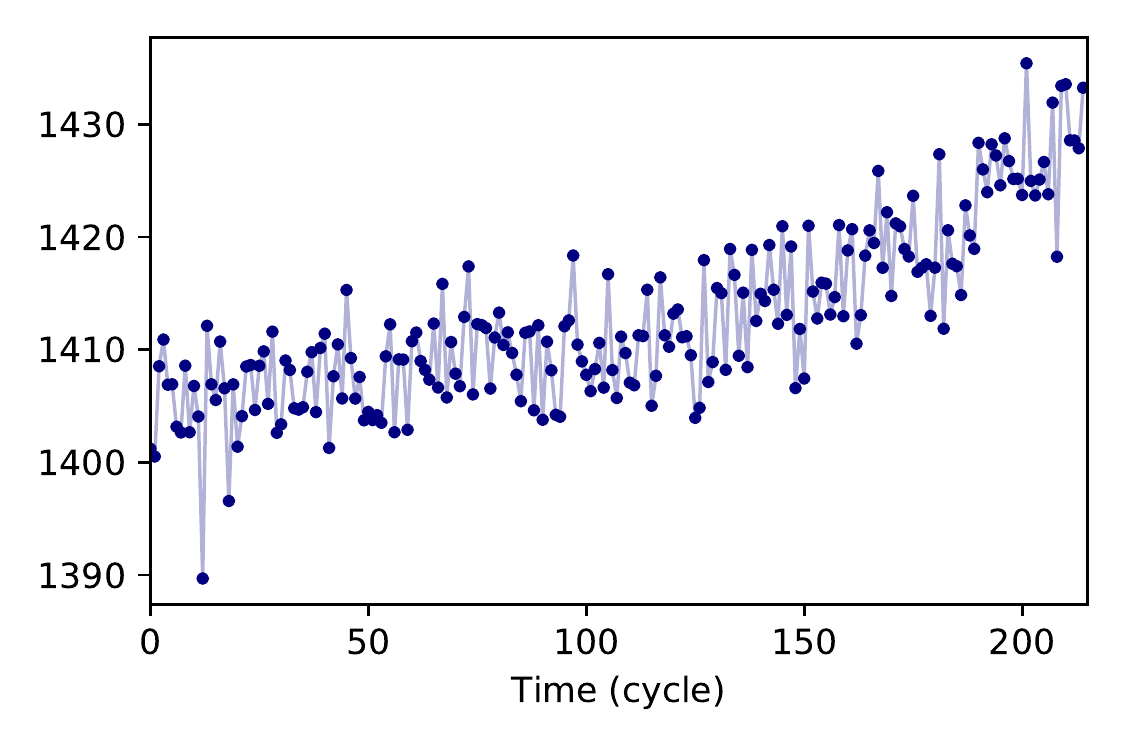}
    \end{minipage}& 
\begin{minipage}{\tableSubplotSizeB\textwidth}
      \includegraphics[width=\textwidth]{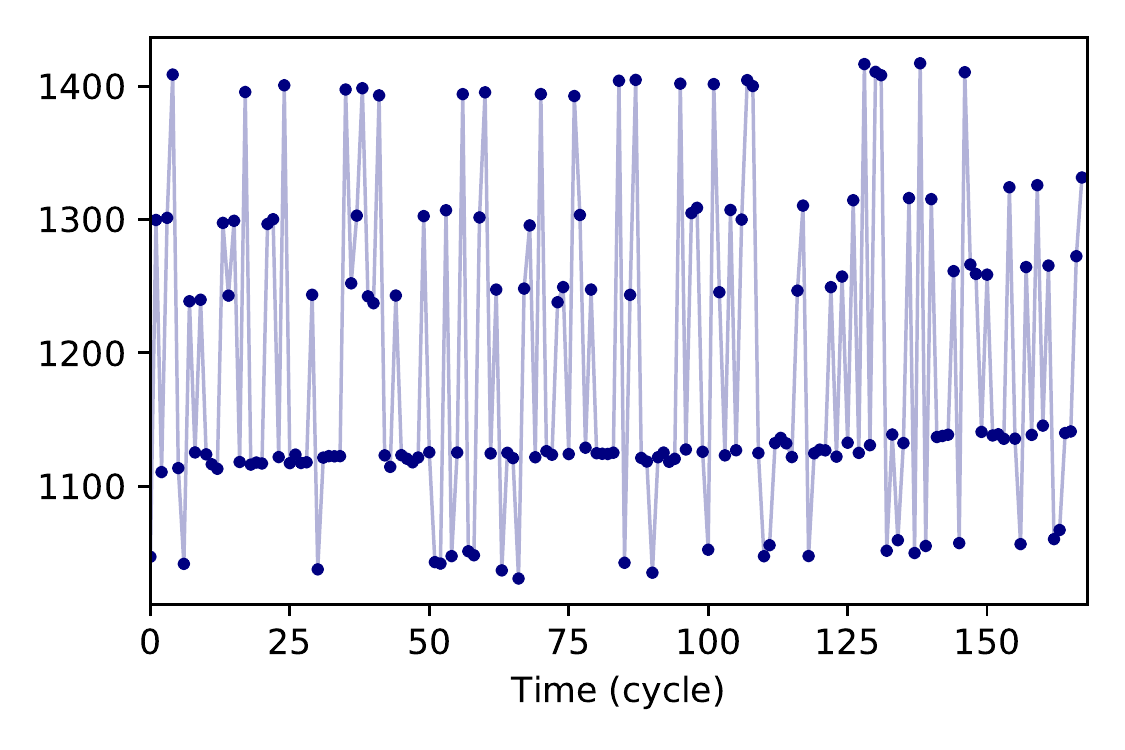}
    \end{minipage}& 
\begin{minipage}{\tableSubplotSizeB\textwidth}
      \includegraphics[width=\textwidth]{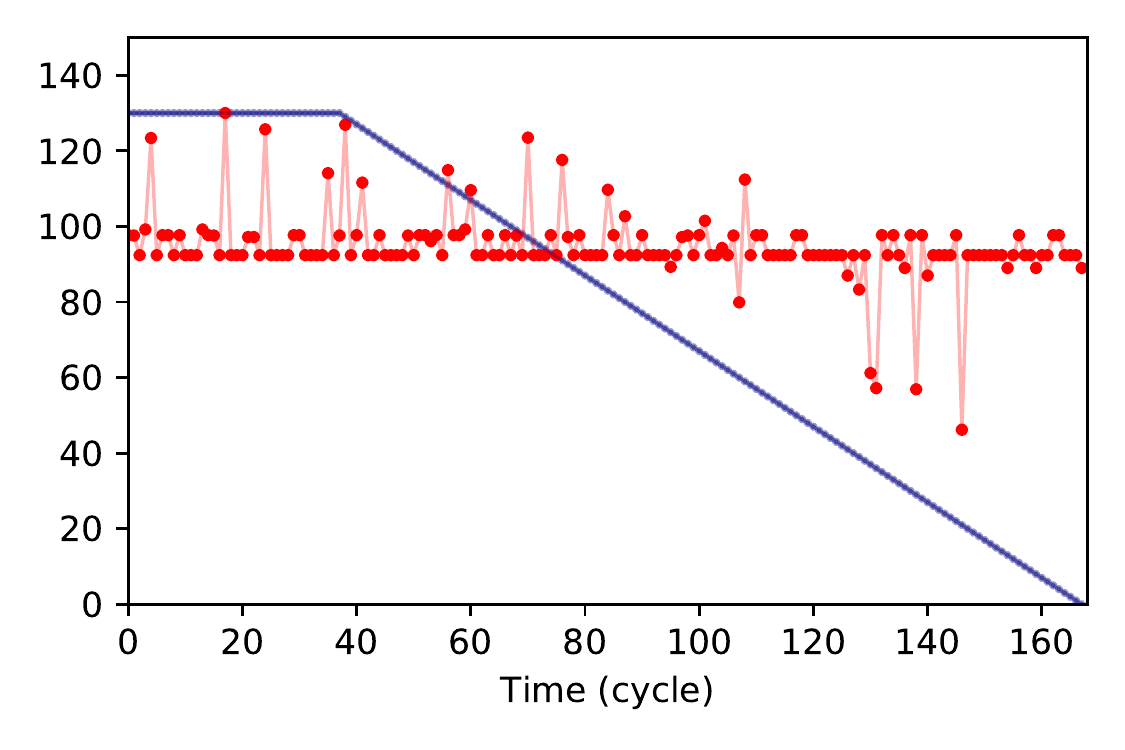}
    \end{minipage}& 
\begin{minipage}{\tableSubplotSizeB\textwidth}
      \includegraphics[width=\textwidth]{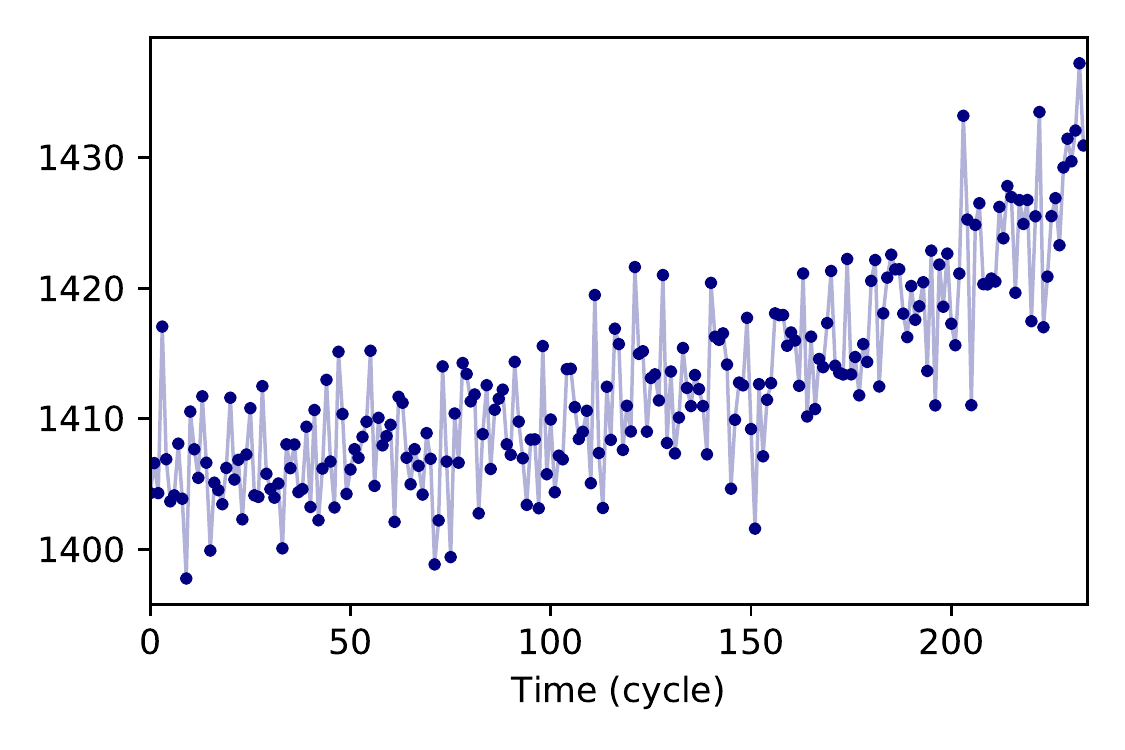}
    \end{minipage}& 
\begin{minipage}{\tableSubplotSizeB\textwidth}
      \includegraphics[width=\textwidth]{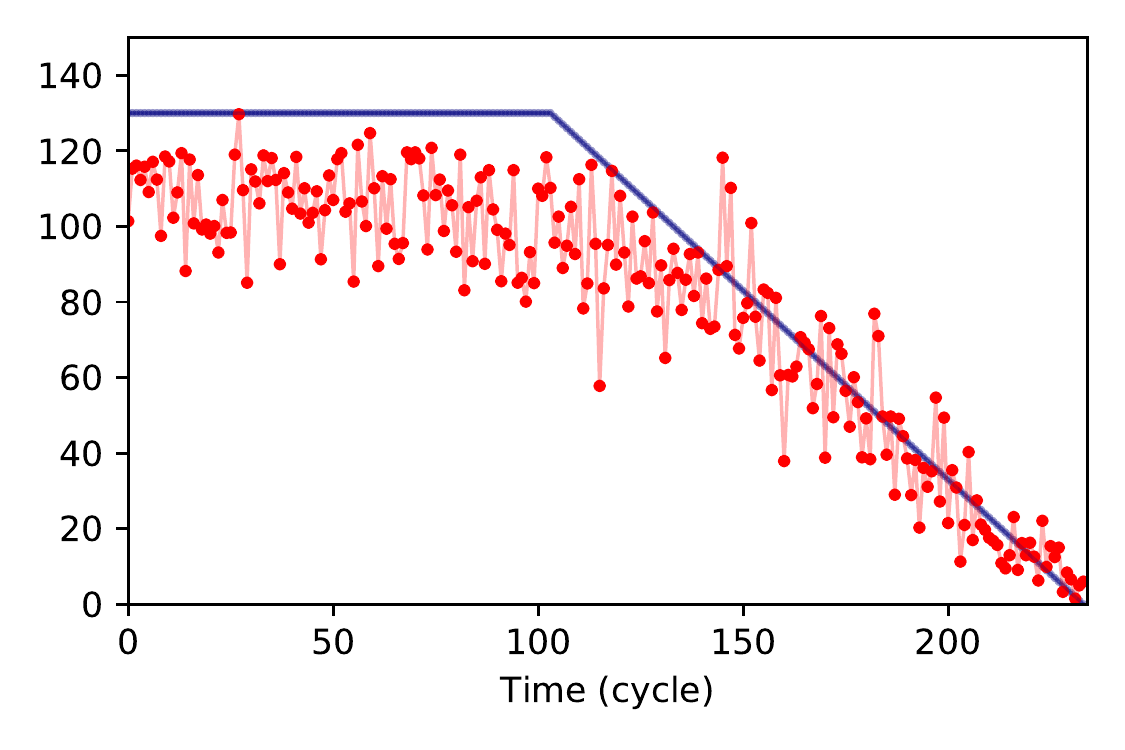}
    \end{minipage}& \\ \midrule
& $\theta_S \subset X^{1}$ & $\theta_T \subset X^{2}$ & $\hat{y}_T=f_S(\theta_T)$ & $\theta_T \subset X^{1}$ & $\hat{y}_T=f_S(\theta_T)$ \\ 
\shortstack[l]{{\scriptsize Approach} \\ {\scriptsize with cosmo} \\ {\scriptsize feature $\theta$}} &
\begin{minipage}{\tableSubplotSizeB\textwidth}
      \includegraphics[width=\textwidth]{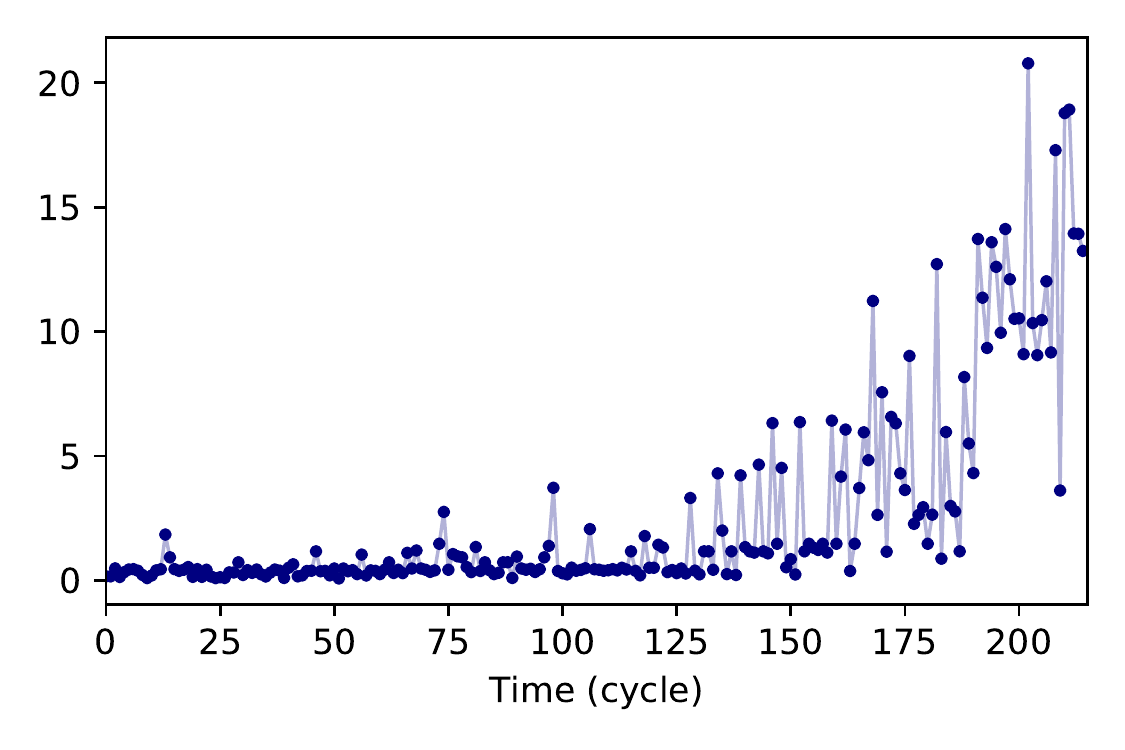}
    \end{minipage}& 
\begin{minipage}{\tableSubplotSizeB\textwidth}
      \includegraphics[width=\textwidth]{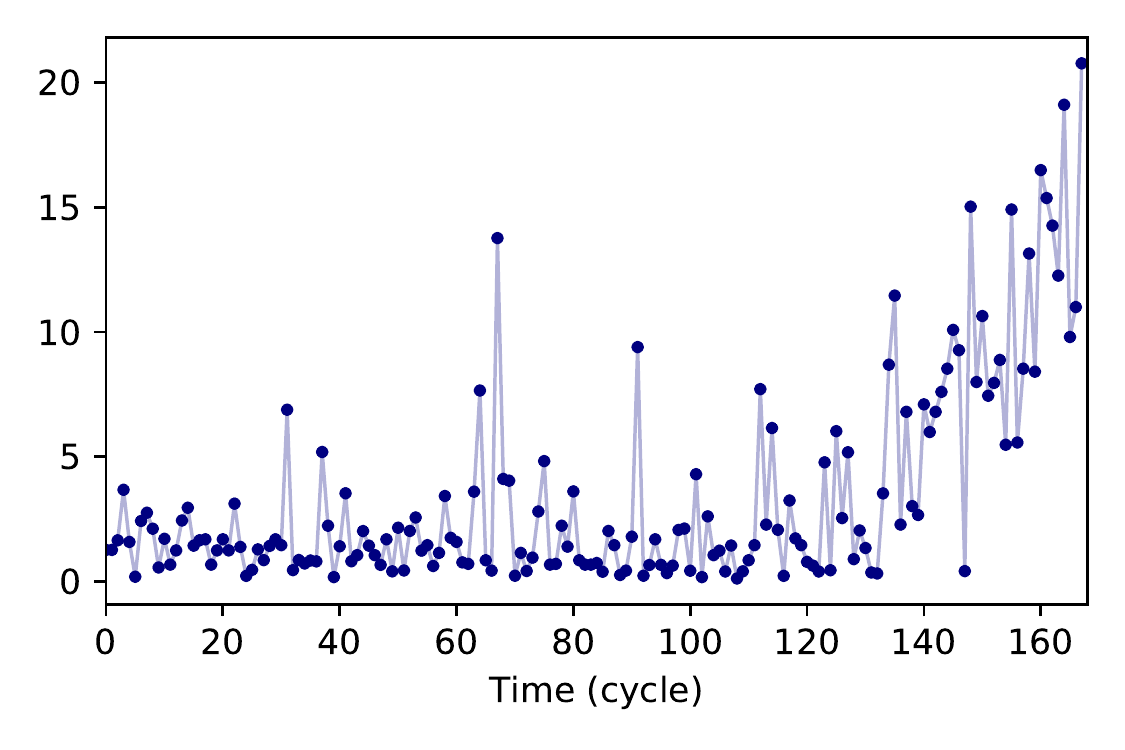}
    \end{minipage}& 
\begin{minipage}{\tableSubplotSizeB\textwidth}
      \includegraphics[width=\textwidth]{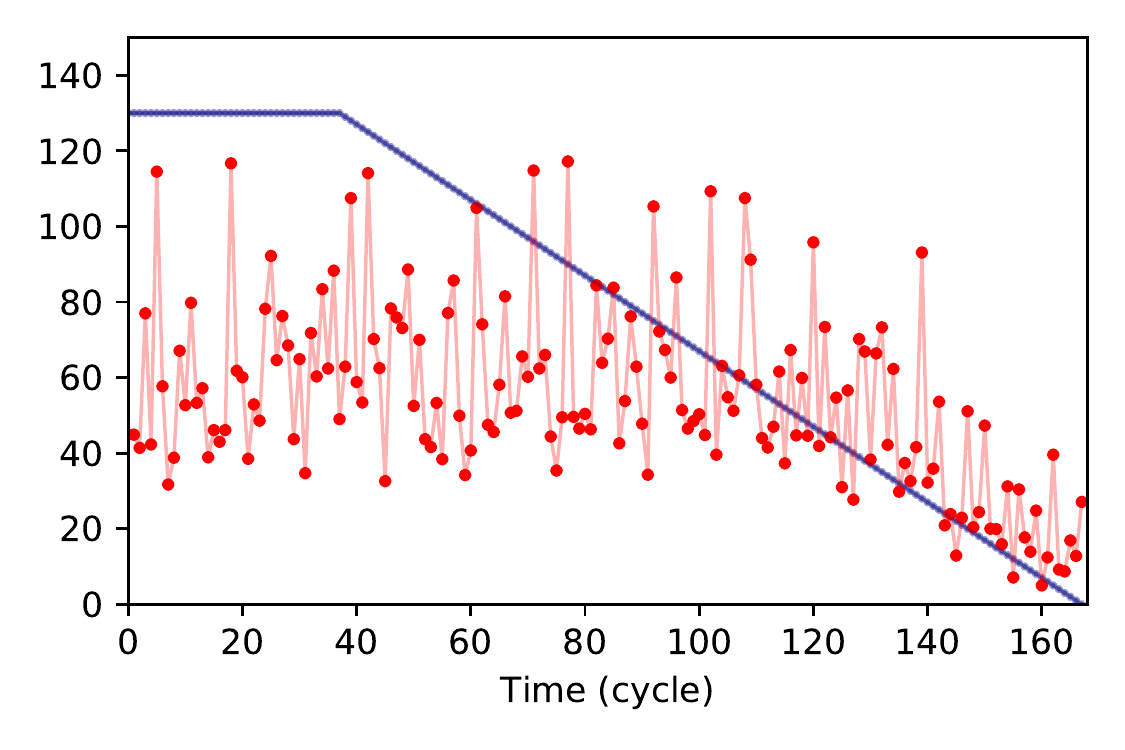}
    \end{minipage}& 
\begin{minipage}{\tableSubplotSizeB\textwidth}
      \includegraphics[width=\textwidth]{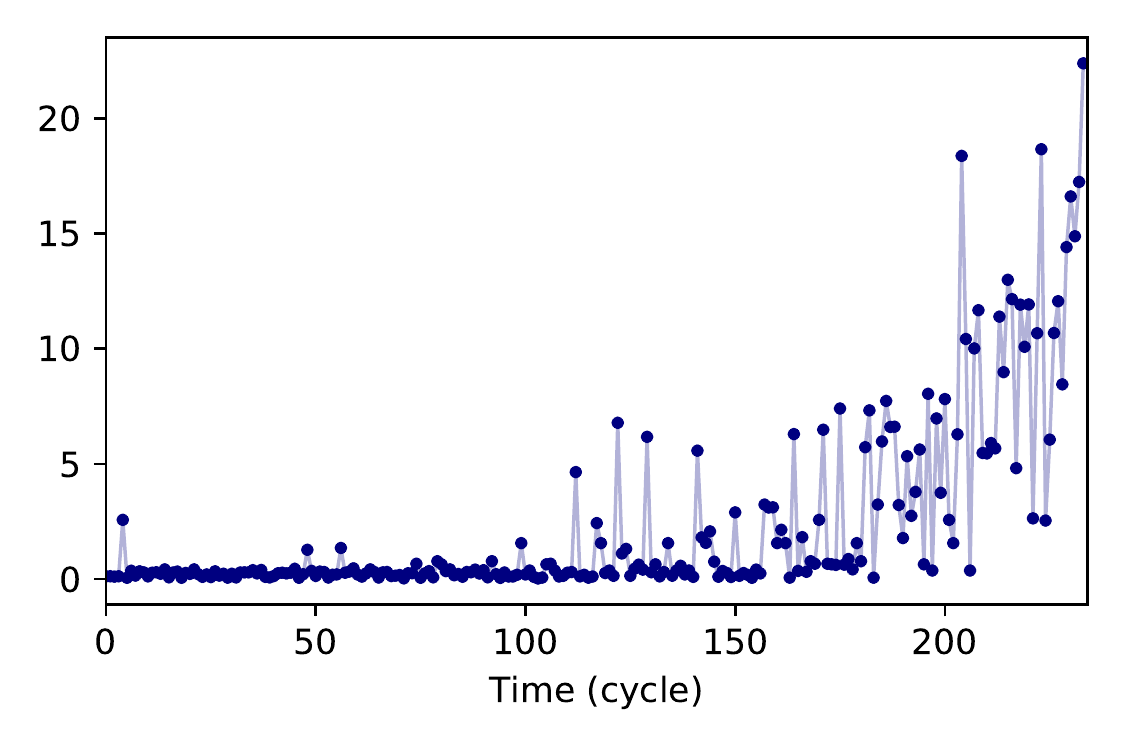}
    \end{minipage}& 
\begin{minipage}{\tableSubplotSizeB\textwidth}
      \includegraphics[width=\textwidth]{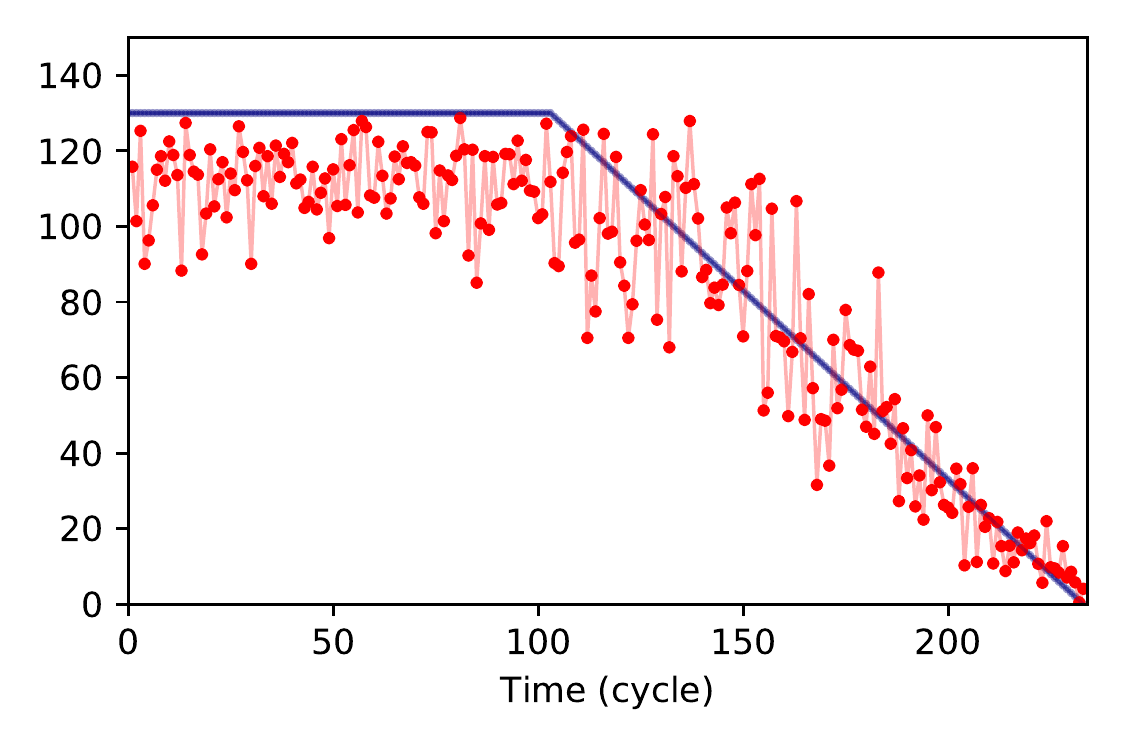}
    \end{minipage}& \\ 
    \bottomrule
\end{tabular}\caption[l]{Illustration of sensor data (the $7$-th feature) and COSMO feature of unit 49 from $X^1$ and unit 20 from $X^2$. Red dots on the third column correspond to RUL prediction $\hat{y}_T$ shows that using COSMO feature $\theta$ is more accurate compared to using sensor data under scenario C1 where $D_S \neq D_T$. The $\hat{y}_T$ on the fifth column shows that the two approaches result in similar performance under scenario A1 where $D_S = D_T$.} 
\label{tab:illustration_cosmo_feature}
\end{center}
\end{table*}

\begin{figure*}
\centering
\subfigure[$\alpha$: scenario A1-A4]{%
\includegraphics[width=0.24\textwidth]{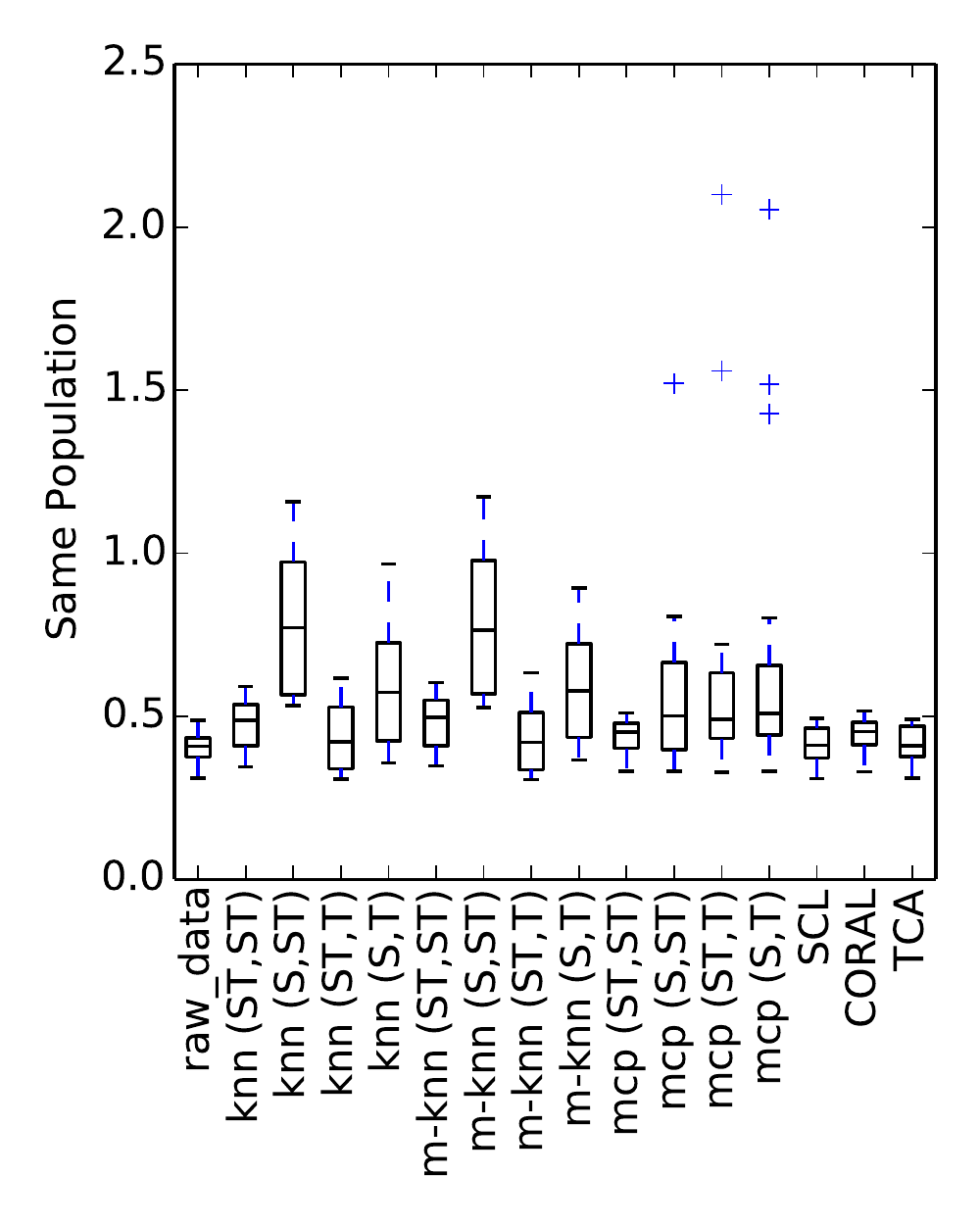}%
\label{fig:box_de_a}%
}
\subfigure[$\beta$: scenario A1-A4]{%
\includegraphics[width=0.24\textwidth]{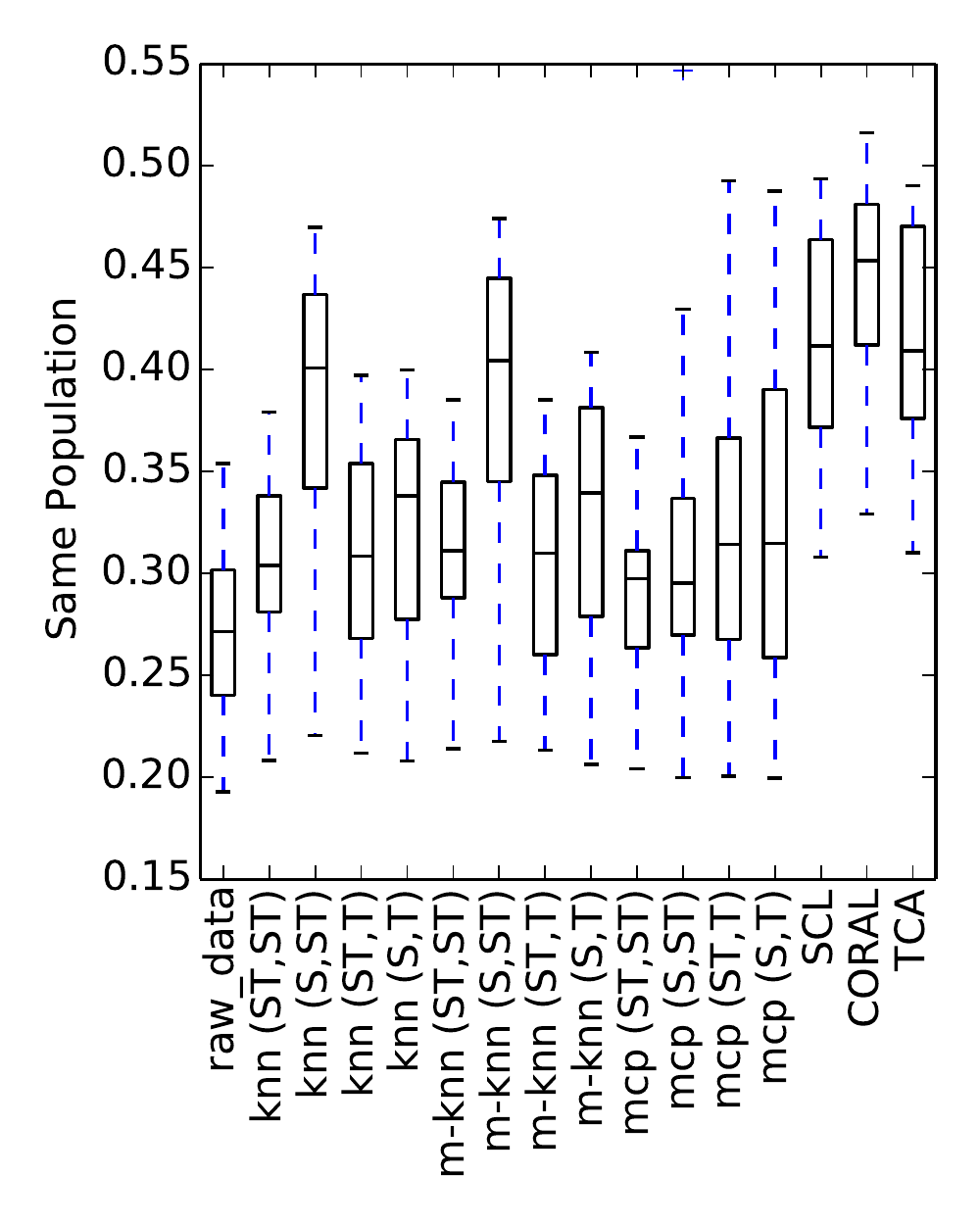}%
\label{fig:box_de_b}%
}
\subfigure[$\alpha$: scenario D]{%
\includegraphics[width=0.24\textwidth]{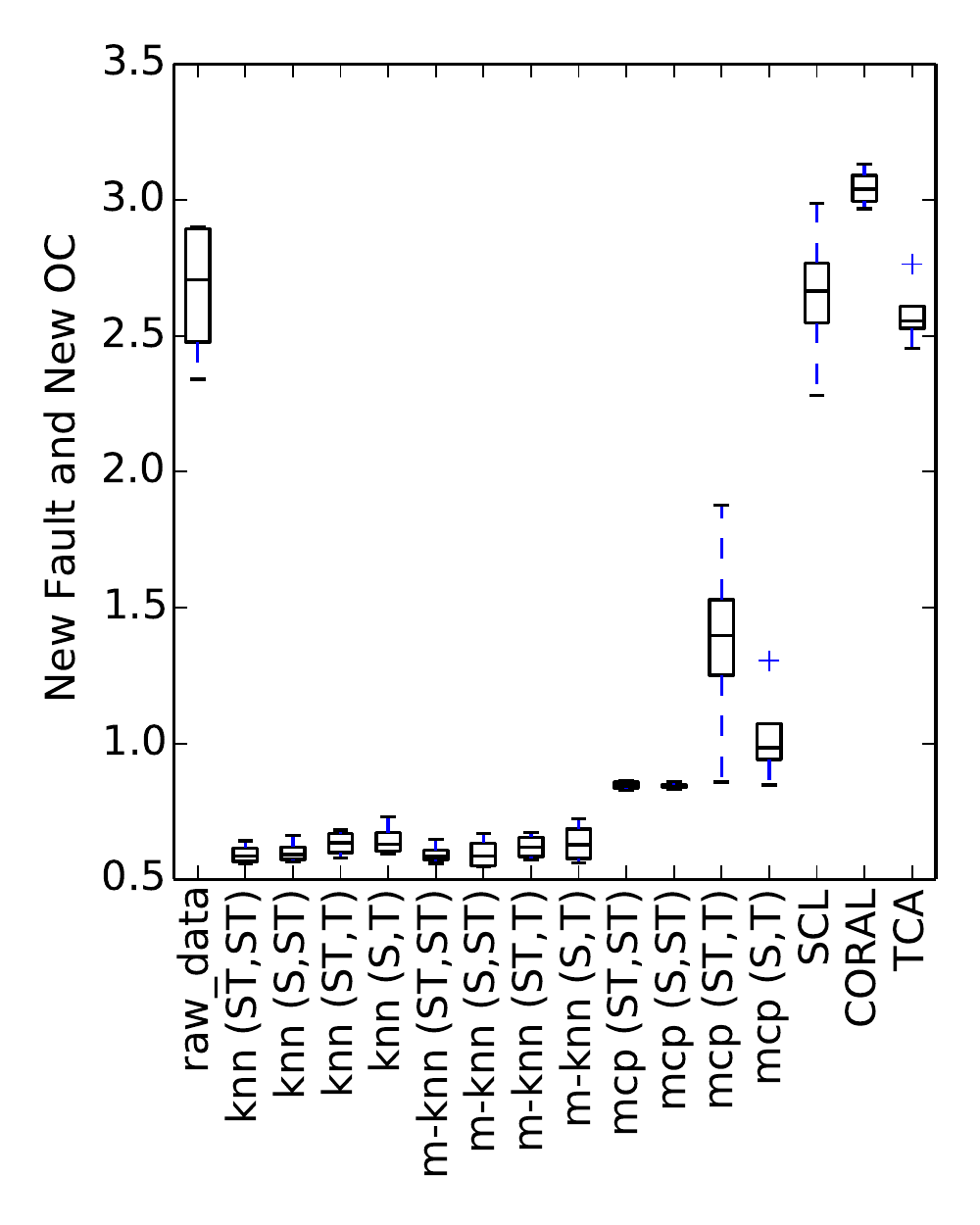}%
\label{fig:box_bo_a}%
}
\subfigure[$\beta$: scenario D]{%
\includegraphics[width=0.24\textwidth]{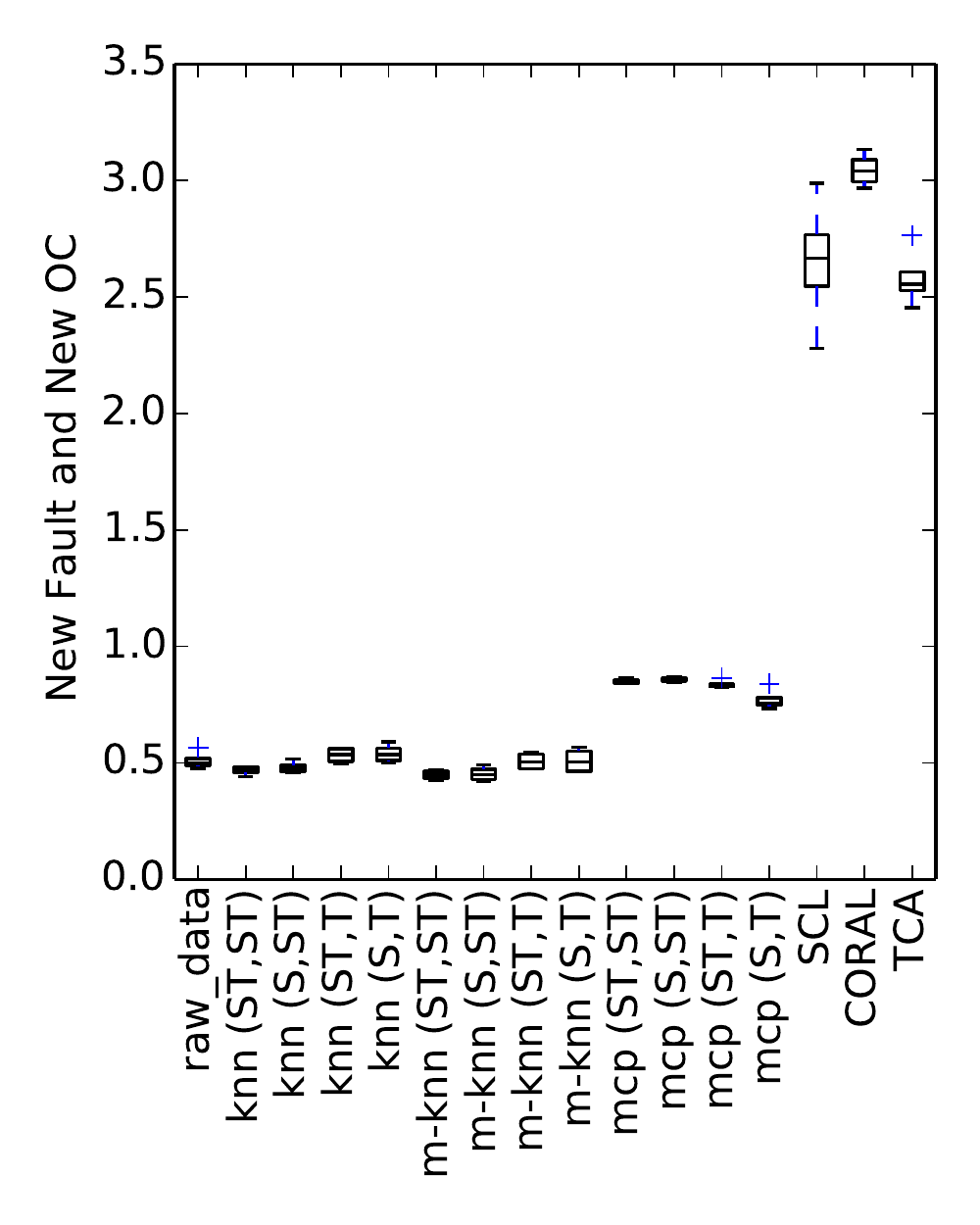}%
\label{fig:box_bo_b}%
}
\subfigure[$\alpha$, scenario B1]{%
\includegraphics[width=0.24\textwidth]{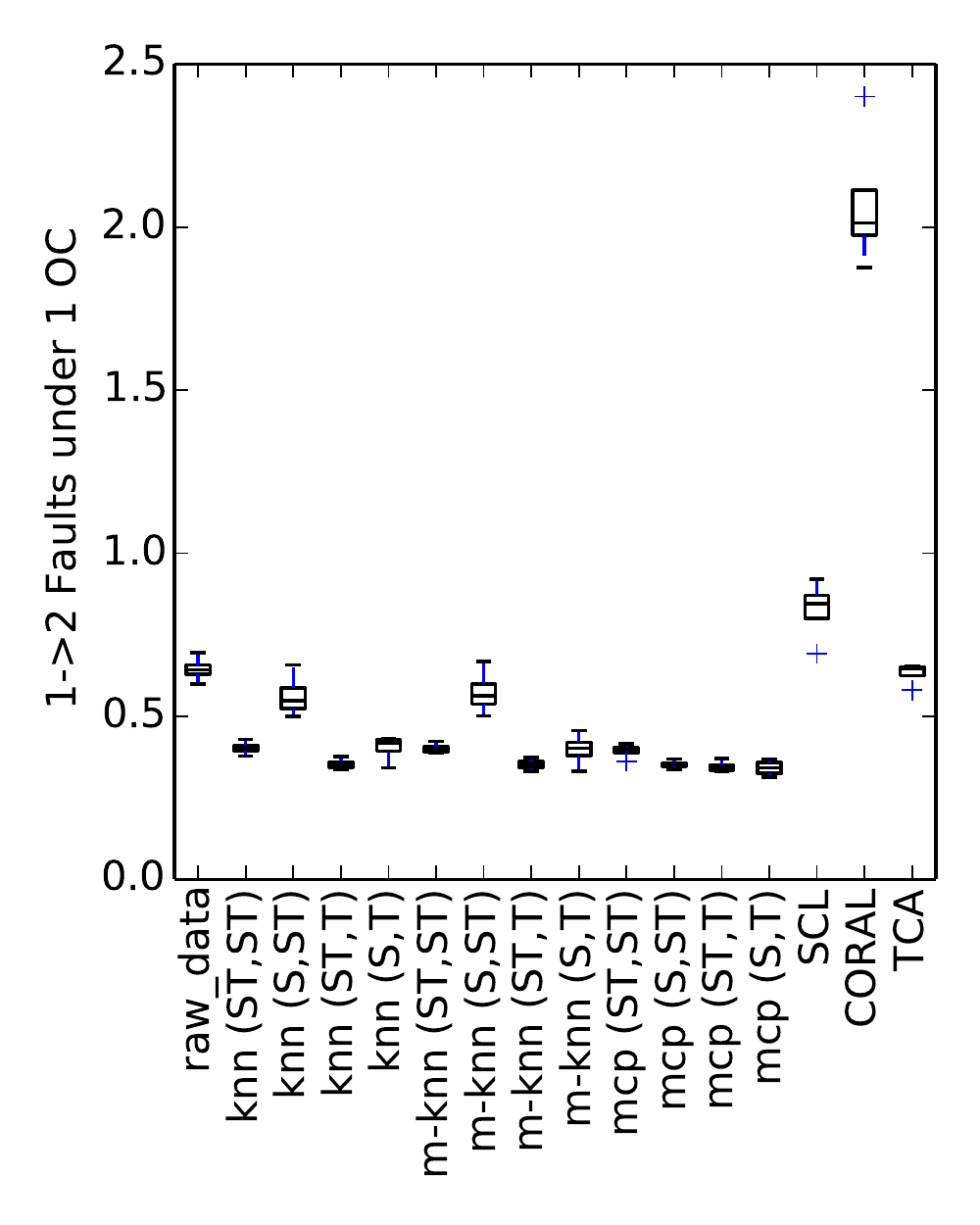}%
\label{fig:box_nf_a1}%
}
\subfigure[$\alpha$, scenario B2]{%
\includegraphics[width=0.24\textwidth]{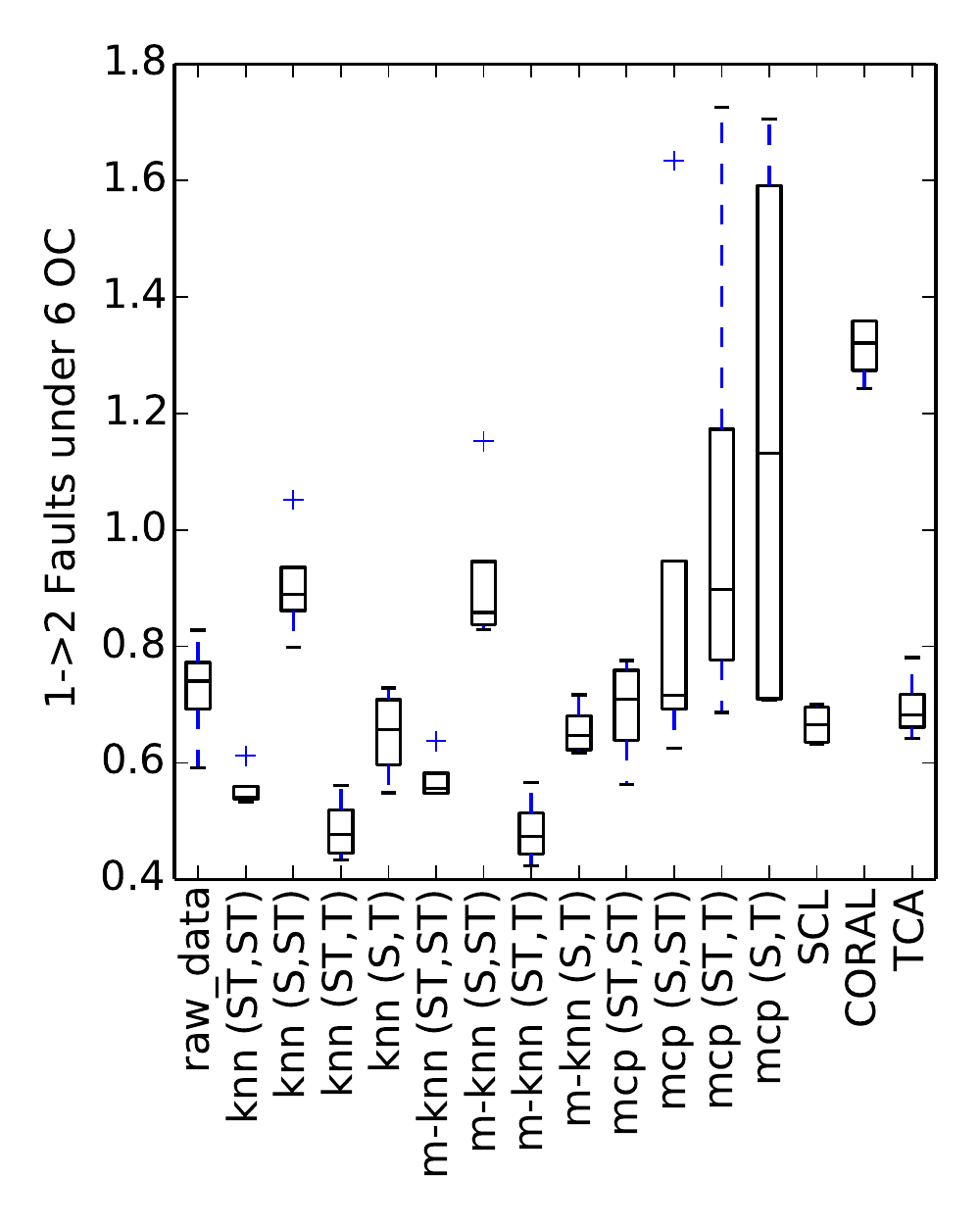}%
\label{fig:box_nf_a6}%
}
\subfigure[$\beta$, scenario B1]{%
\includegraphics[width=0.24\textwidth]{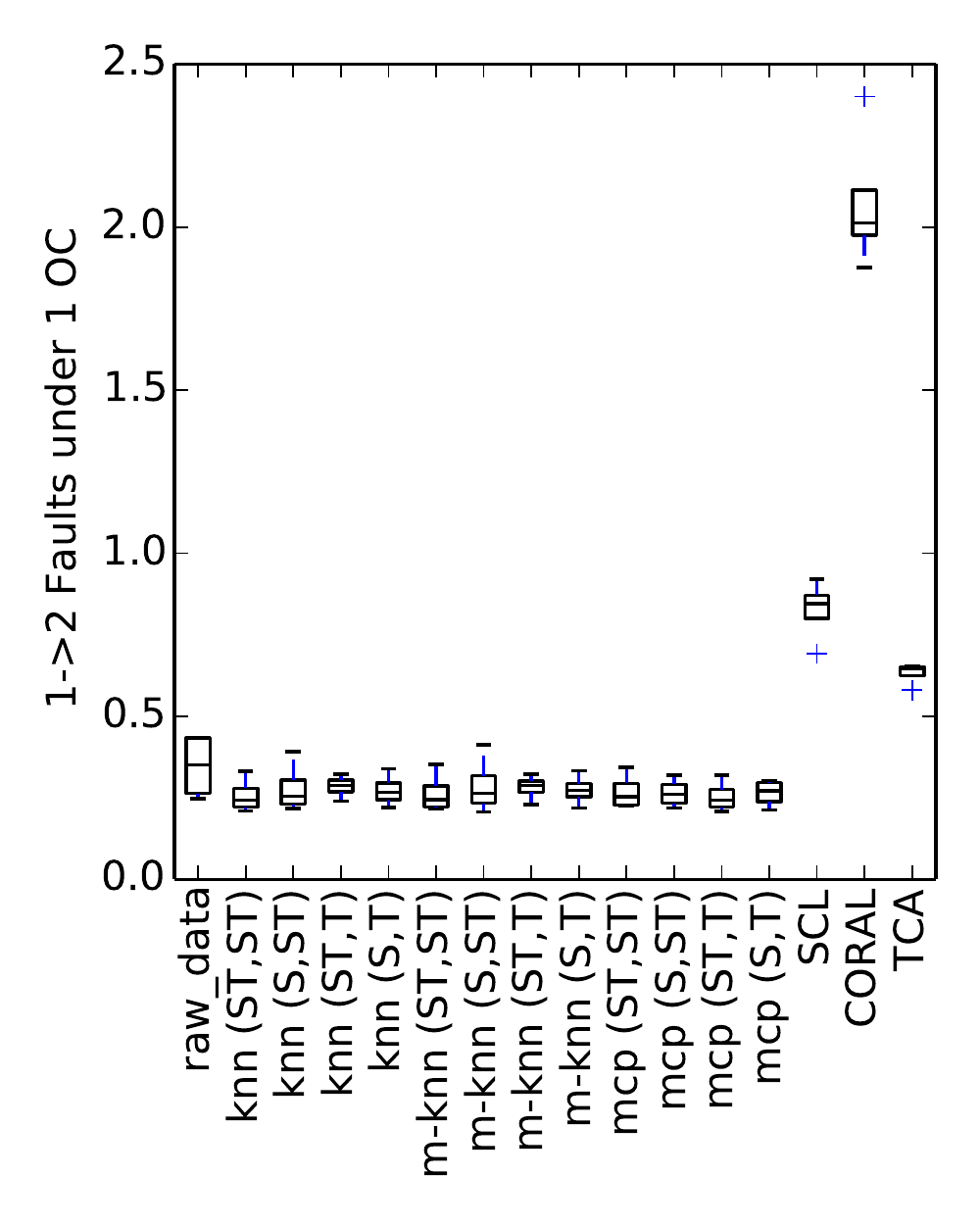}%
\label{fig:box_nf_b1}%
}
\subfigure[$\beta$, scenario B2]{%
\includegraphics[width=0.24\textwidth]{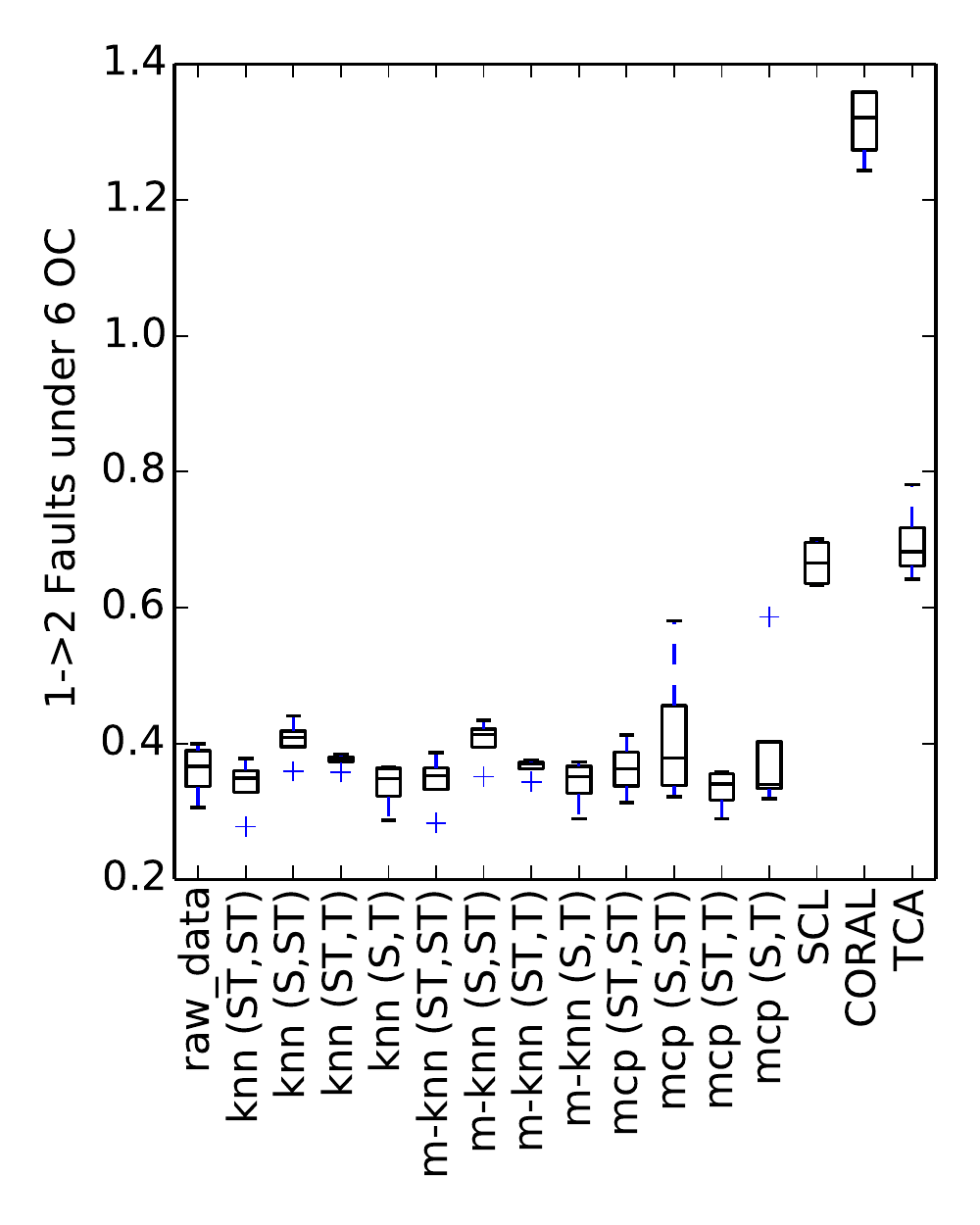}%
\label{fig:box_nf_b6}%
}\vspace{-2mm}
\subfigure[$\alpha$, scenario C1]{%
\includegraphics[width=0.24\textwidth]{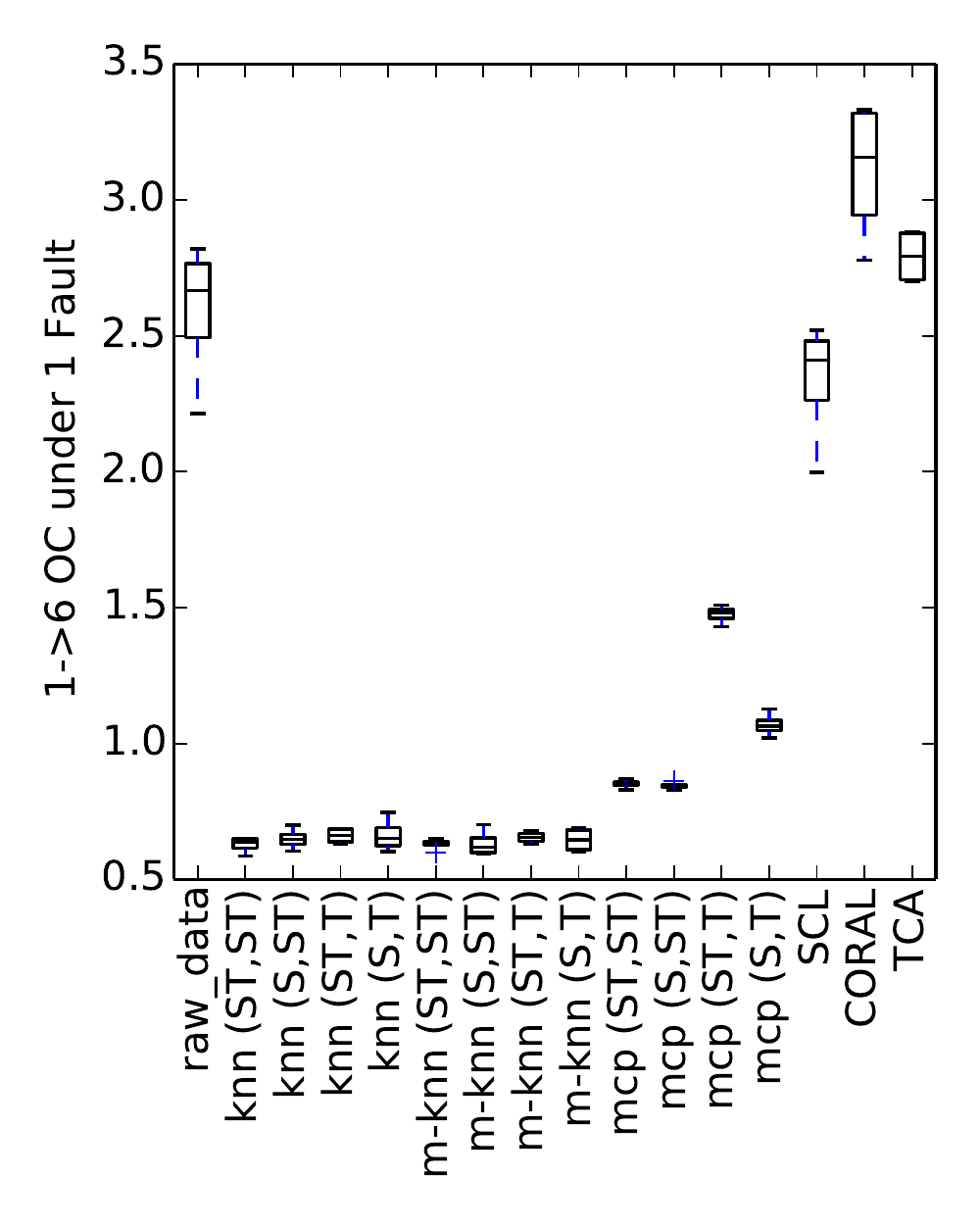}%
\label{fig:box_no_a1}%
}
\subfigure[$\alpha$, scenario C2]{%
\includegraphics[width=0.24\textwidth]{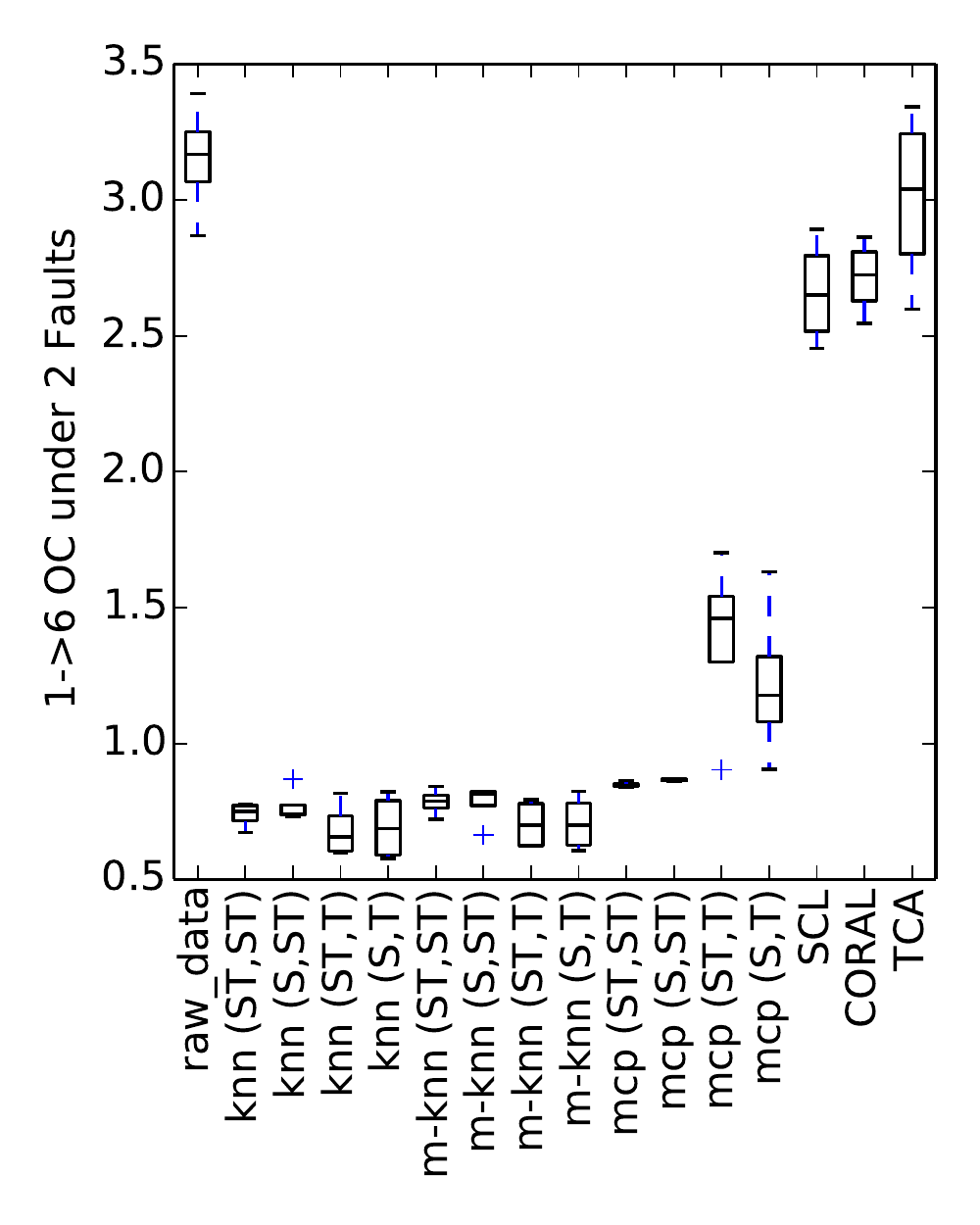}%
\label{fig:box_no_a2}%
}
\subfigure[$\beta$, scenario C1]{%
\includegraphics[width=0.24\textwidth]{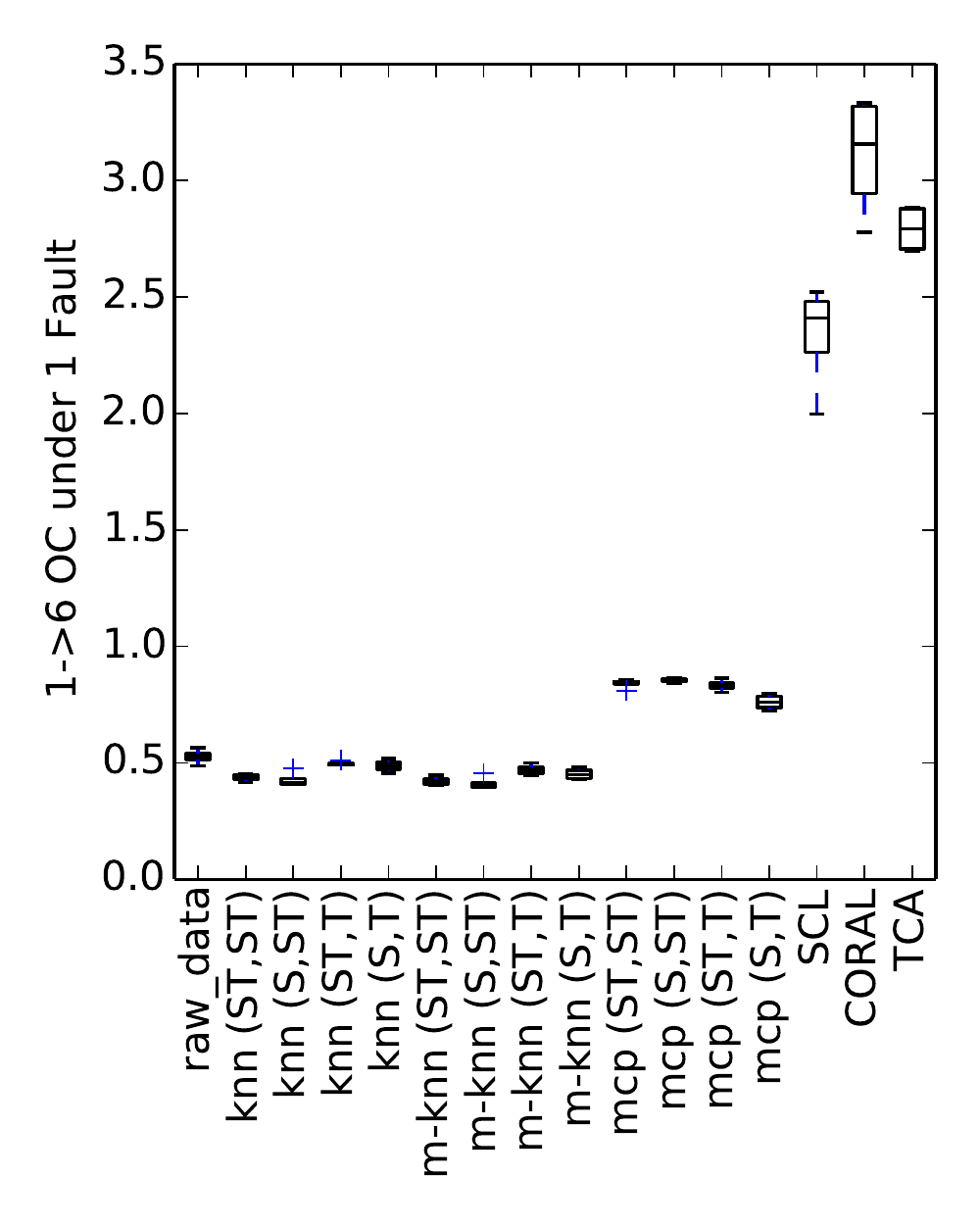}%
\label{fig:box_no_b1}%
}
\subfigure[$\beta$, scenario C2]{%
\includegraphics[width=0.24\textwidth]{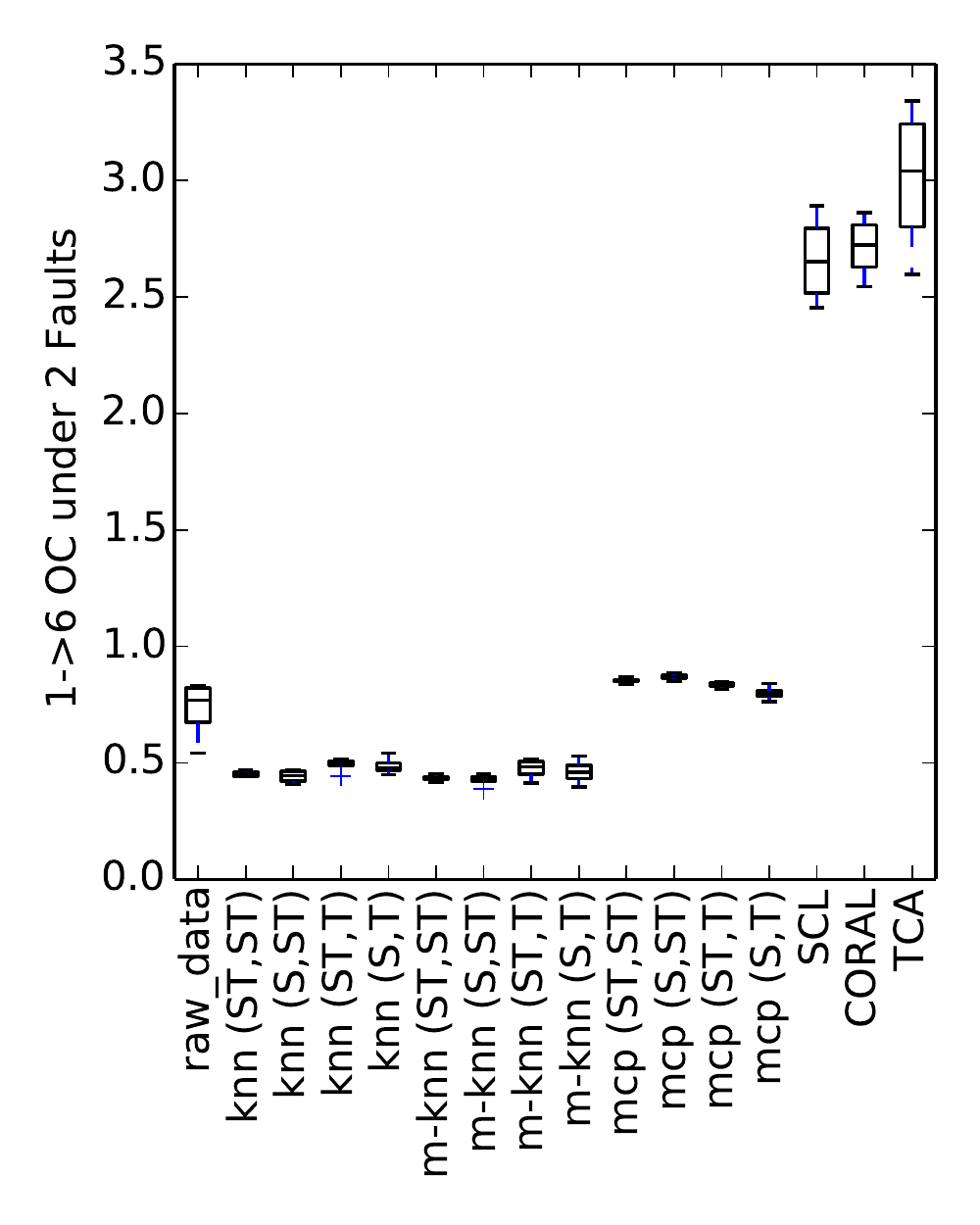}%
\label{fig:box_no_b2}%
}\vspace{-2mm}
\caption{Performance comparison with MAPE on scenario A to D under mode $\alpha$ and $\beta$.}
\label{fig:box_ALL}
\end{figure*}

\end{document}